\documentclass{article}

\usepackage[english]{babel}

\usepackage[letterpaper,top=2cm,bottom=2cm,left=3cm,right=3cm,marginparwidth=1.75cm]{geometry}

\usepackage{amsmath}
\usepackage{graphicx}
\usepackage[table]{xcolor}
\usepackage[colorlinks=true, allcolors=blue]{hyperref}
\usepackage{cleveref}
\usepackage{makecell} 
\usepackage{subcaption}
\usepackage{multirow}
\usepackage{array}
\usepackage{makecell}
\usepackage{wrapfig}
\usepackage{booktabs}

\usepackage{rcw_macros}

\usepackage{colortbl}
\usepackage{rotating}
\usepackage{ifthen}

\usepackage[backend=biber,style=alphabetic]{biblatex}
\usepackage[T1]{fontenc}
\usepackage[utf8]{inputenc} 
\DeclareUnicodeCharacter{200B}{{\hskip 0pt}}

\title{A Roadmap for Climate-Relevant Robotics Research}
\date{}

\usepackage{tabularx}
\usepackage[affil-it]{authblk}  

\author{
Alan Papalia\textsuperscript{1,2,*},
Charles Dawson\textsuperscript{3,*},
Laurențiu L. Anton\textsuperscript{4,$\dagger$},
Norhan Magdy Bayomi\textsuperscript{4,$\dagger$},
Bianca Champenois\textsuperscript{4,$\dagger$},
Jung-Hoon Cho\textsuperscript{4,$\dagger$},
Levi Cai\textsuperscript{5,$\dagger$},
Joseph DelPreto\textsuperscript{4,$\dagger$},
Kristen Edwards\textsuperscript{4,$\dagger$},
Bilha-Catherine Githinji\textsuperscript{4,$\dagger$},
Cameron Hickert\textsuperscript{4,$\dagger$},
Vindula Jayawardana\textsuperscript{4,$\dagger$},
Matthew Kramer\textsuperscript{6,$\dagger$},
Shreyaa Raghavan\textsuperscript{4,$\dagger$},
David Russell\textsuperscript{7,$\dagger$},
Shide Salimi\textsuperscript{8,$\dagger$},
Jingnan Shi\textsuperscript{4,$\dagger$},
Soumya Sudhakar\textsuperscript{4,$\dagger$},
Yanwei Wang\textsuperscript{4,$\dagger$},
Shouyi Wang\textsuperscript{9,$\dagger$},\linebreak
Luca Carlone\textsuperscript{4},
Vijay Kumar\textsuperscript{10},
Daniela Rus\textsuperscript{4},
John E. Fernandez\textsuperscript{4},
Cathy Wu\textsuperscript{4},
George Kantor\textsuperscript{11},
Derek Young\textsuperscript{7},
Hanumant Singh\textsuperscript{1}
}

\begin{document}
\maketitle

\begingroup
\renewcommand\thefootnote{}
\footnotetext{
    \textsuperscript{*}These authors contributed equally to this work.
    \textsuperscript{$\dagger$}Authors listed in alphabetical order.
    \textsuperscript{1}Northeastern University,
    \textsuperscript{2}University of Michigan,
    \textsuperscript{3}Massachusetts Department of Energy Resources,
    \textsuperscript{4}Massachusetts Institute of Technology (MIT),
    \textsuperscript{5}MIT-WHOI Joint Program,
    \textsuperscript{6}Tufts University,
    \textsuperscript{7}University of California Davis,
    \textsuperscript{8}Harvard University,
    \textsuperscript{9}Woods Hole Oceanographic Institution,
    \textsuperscript{10}University of Pennsylvania,
    \textsuperscript{11}Carnegie Mellon University
}
\endgroup

%
%

\begin{figure}[!h]
    \centering
    \includegraphics[width=\textwidth]{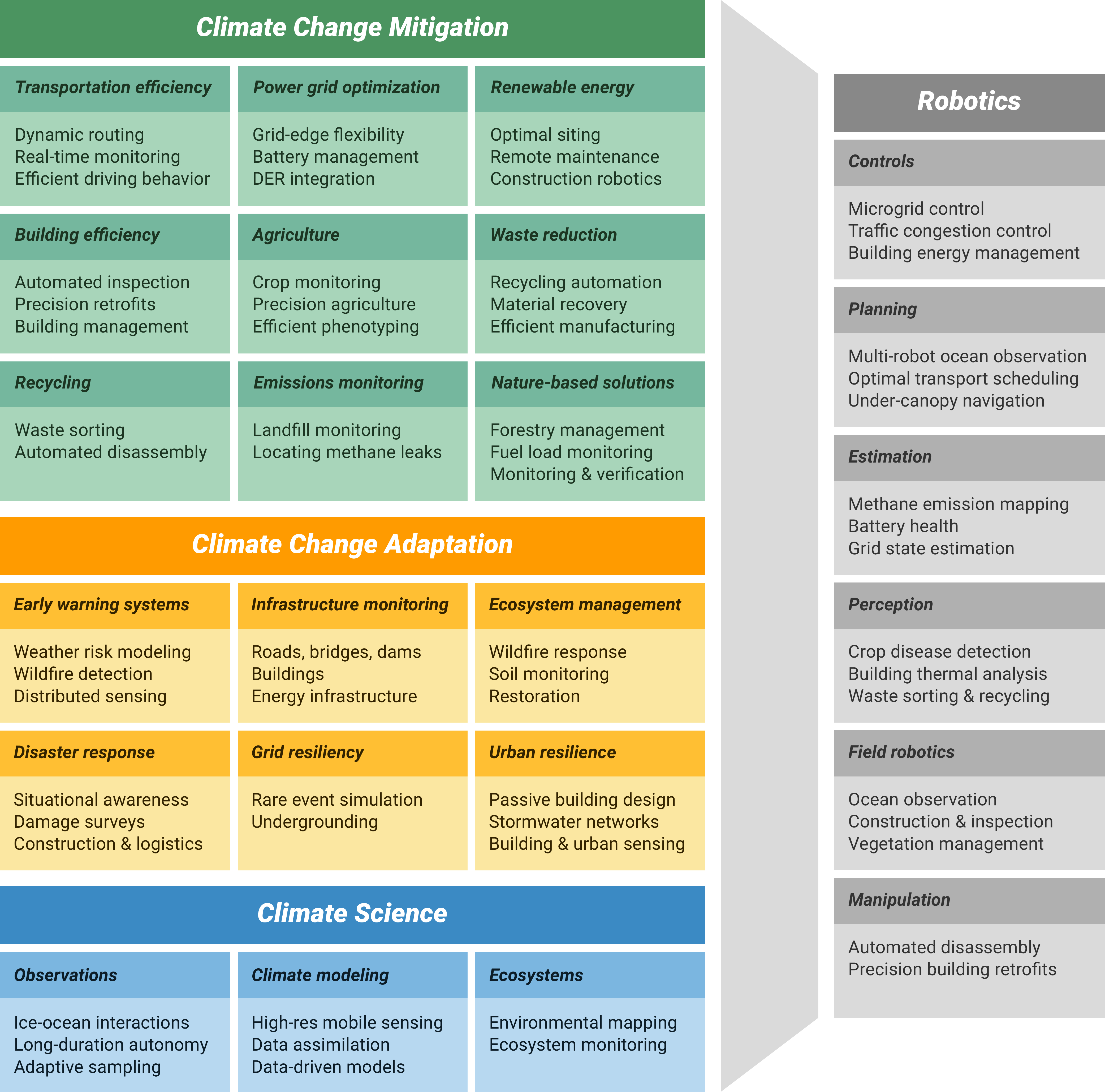}
    \caption{An overview of climate change challenges along
        with selected opportunities for high-impact research in the robotics community.}
\end{figure}


\begin{abstract}
    Climate change is one of the defining challenges of the
    21\textsuperscript{st} century, and
    many in the robotics community are looking for ways to contribute.
    This paper presents a roadmap for climate-relevant robotics research,
    identifying high-impact opportunities for collaboration between roboticists and experts across climate domains such as energy, the built environment, transportation, industry, land use, and Earth sciences.
    These applications include problems such as energy systems optimization, construction, precision agriculture,
    building envelope retrofits, autonomous trucking, and
    large-scale environmental monitoring.
    Critically, we include opportunities to apply not only physical robots but also the broader robotics
    toolkit --- including planning, perception, control, and
    estimation algorithms --- to climate-relevant problems.
    A central goal of this roadmap is to inspire new research directions and
    collaboration by highlighting specific, actionable problems at the
    intersection of robotics and climate.
    This work represents a collaboration between robotics
    researchers and domain experts in various climate disciplines, and it
    serves as an invitation to the robotics community to
    bring their expertise to bear on urgent climate priorities.
\end{abstract}

\section{Introduction}
\label{sec:intro}

Mitigating and adapting to the worst effects of climate change will
require coordinated scientific, engineering, and policy effort across domains
including energy, the built environment, transportation, land use, and earth sciences.
In many of these domains, robotics research has the potential to contribute to climate solutions.
This includes opportunities to
deploy not only physical robots (e.g., drones, autonomous underwater vehicles,
etc.) but also the core computational and theoretical tools of robotics
researchers (e.g., control theory, optimization, perception, estimation, etc.).
In short: it is not only robots but also \textit{roboticists} who can
contribute. Yet many ``climate-curious'' roboticists struggle to
connect their work to climate-relevant problems or know where to begin.
\textbf{This paper aims to bridge the gap between the robotics community and
    climate-relevant research problems.}
\def\applicationInspiredFootnote{\footnote{We note that
        \emph{application-inspired} research includes
        both applied and theoretical research; the common theme is that the research
        is inspired by real-world problems~\cite{stokes2011pasteur}.}~}
Specifically, we aim to support
\emph{application-inspired}\applicationInspiredFootnote
robotics research by identifying concrete opportunities
for robotics to address important problems in climate change mitigation,
adaptation, and science. Our goal is to encourage interdisciplinary collaboration and to
provide roboticists with a structured introduction to applying their expertise
to different climate-relevant problems.

This paper is a collaboration between roboticists and climate domain experts,
presenting the results of in-depth expert discussions and literature reviews.
We intend for this paper to provide a starting point for roboticists to identify
opportunities for impact, encourage further exploration of various climate domains, and inspire
more collaboration between the robotics and climate communities.

\titledsubsection{Motivation and the knowledge gap}
\label{sec:intro:motivation}

This paper was inspired by the common challenges many of us faced in identifying impactful climate-relevant directions for our own research.
Identifying clearly defined, high-impact problems that
the robotics community is well-situated to address is difficult because of two
major factors:
\begin{enumerate}
    \item Identifying problems with meaningful climate impact requires a
          deep understanding of the challenges and nuances of different climate domains, and
    \item Once a problem is identified, it may not be clear whether the robotics
          community can contribute to solving the problem
          (e.g., if policy is the limiting factor, further technical work may have limited impact).
\end{enumerate}
These two factors characterize the knowledge gap that this paper fills:
we provide a structured overview of the climate domain and highlight specific
problems that the robotics community is well-situated to address, emphasizing
the individual robotics subfields that can contribute to each problem.

\titledsubsection{Scope}
\label{sec:intro:structure-and-scope}

Throughout this paper, we take a deliberately broad view of robotics, including not
only physical robots but also core computational and theoretical tools like
optimization, decision making, and estimation theory. For example, state
estimation can help autonomous underwater vehicles gather data in challenging
ocean conditions, but fundamental research on state estimation can also directly
support better climate modeling tools or help integrate distributed energy resources
like solar panels into the electric grid.

Due to the broad scope of both climate change and robotics, we are unable to
provide comprehensive coverage of each climate domain. To focus our discussion,
we prioritize high-impact problems in each climate domain where there is a clear
role for robotics to contribute. We do not cover all
problems matching these criteria, but we are hopeful that this paper will
provide the starting point for a broader discussion and collaboration between
the robotics and climate communities.

We emphasize that our primary goal is not to provide a literature review of
existing robotics research in climate-relevant domains.  Rather, it is to
provide a \emph{roadmap} for future research, identifying areas where robotics
can meaningfully contribute to open problems in each climate domain. We
reference existing work where it is relevant to understand the state of the art
in each area, but we do not aim to provide complete coverage of all works in a
given area, instead highlighting high-impact open problems for future research.
%

At the same time, we deliberately choose not to rank research directions by importance, feasibility, or potential impact
%
Ranking disparate challenges --- e.g., better understanding Arctic ice melt versus improving energy
grid resilience --- is subjective and likely misleading.
%
Instead, our contribution is to curate a selection of high-impact intersections
of climate needs and robotics capabilities. These directions were selected
based on recurring themes in expert interviews and the literature, filtering for importance without claiming exhaustiveness. We hope that this approach
highlights actionable opportunities while maintaining a broad tent that is open
to many forms of contributions and encourages creativity and cross-disciplinary
exploration.

\titledsubsection{Structure}
\label{sec:intro:structure}

This paper presents an interdisciplinary roadmap that identifies promising
research opportunities at the intersection of robotics and climate change.
The roadmap is organized around \textbf{six major climate domains}: energy, the
built environment, transportation, industry, land use, and Earth systems.

Each domain is covered in a dedicated section that follows a consistent
structure:

\begin{itemize}
    \item Each section begins with an \textbf{executive summary} that provides a
          high-level overview of the domain, its climate relevance, and key
          robotics opportunities.
    \item We introduce core background and context for the climate domain (e.g.,
          an overview of electrical energy systems or building and urban systems).
    \item We identify specific climate challenges (e.g., wildfire management,
          methane leakage) that are both scientifically urgent and potentially
          addressable through advances in robotics.
    \item For each challenge, we highlight how different robotics subfields
          (listed below) can contribute through both theoretical
          innovation and field deployment.
    \item Throughout each robotics subsection, we include \textbf{outlook boxes}
          that propose concrete \emph{future directions} for researchers to
          pursue (e.g., “Future Directions: Robot Design and Field Robotics for
          Land Use”).
\end{itemize}

To connect these domains to concrete directions for robotics research, we consider \textbf{six core robotics
    subfields}: perception, planning, control, estimation, manipulation, and field
robotics.  These groupings were chosen to provide broad umbrella categories that
capture both the core competencies of the robotics community and the major
avenues for impact identified in the paper.
These subfields are not mutually exclusive; for example, many tasks
fall at the intersection of perception and estimation or planning and control.
Additionally, these are not the only robotics communities discussed in this
paper. We discuss other subfields --- e.g., design and human-robot interaction --- when relevant, but for consistency we include them under one of these six core subfields.
\Cref{tab:intro:domain-matrix} summarizes links between climate
domains and relevant robotics subfields, with references to where each connection
is discussed in the paper.

\newcommand{\iconcell}[2][0.8cm]{%
  \IfFileExists{#2}{%
    \raisebox{0pt}[\dimexpr#1+0.3ex\relax][0pt]{\includegraphics[height=#1]{#2}}%
    \rule{0pt}{#1} 
  }{[missing icon]}
}

\definecolor{lightgray}{gray}{0.93}
\definecolor{lightgray}{gray}{0.96}

\definecolor{lightblue}{rgb}{0.85, 0.92, 0.98}
\definecolor{skyblue}{rgb}{0.76, 0.88, 0.97}
\definecolor{powderblue}{rgb}{0.90, 0.95, 1.00}

\definecolor{lightgreen}{rgb}{0.88, 0.95, 0.88}
\definecolor{mintgreen}{rgb}{0.84, 0.96, 0.87}
\definecolor{seagreen}{rgb}{0.87, 0.96, 0.92}

\definecolor{lightlavender}{rgb}{0.94, 0.90, 0.98}
\definecolor{lightpeach}{rgb}{1.00, 0.95, 0.88}
\definecolor{sand}{rgb}{0.96, 0.94, 0.88}

\newcommand{\altcolcolors}[9]{%
  & \cellcolor{white}{#1} &%
  \cellcolor{#2}{#4} &%
  \cellcolor{#3}{#5} &%
  \cellcolor{#2}{#6} &%
  \cellcolor{#3}{#7} &%
  \cellcolor{#2}{#8} &%
  \cellcolor{#3}{#9} \\
  \cline{2-8}
}

\def\methodRotate{45}
\def\domainRotate{0}
\renewcommand{\arraystretch}{2.1}

\begin{table}[h]
  \centering

  \begin{tabular}{
    >{\centering\arraybackslash}m{0.75cm}  
    >{\centering\arraybackslash}m{3.5cm} |
    >{\centering\arraybackslash}m{1.3cm}
    >{\centering\arraybackslash}m{1.3cm}
    >{\centering\arraybackslash}m{1.3cm}
    >{\centering\arraybackslash}m{1.3cm}
    >{\centering\arraybackslash}m{1.3cm}
    >{\centering\arraybackslash}m{1.3cm}
    }

    \multicolumn{2}{c}{} &
    \multicolumn{6}{c}{\Large\textbf{Robotics Discipline}}                     \\

                         &
                         & \rotatebox{\methodRotate}{\textbf{Controls}}
                         & \rotatebox{\methodRotate}{\textbf{Planning}}
                         & \rotatebox{\methodRotate}{\textbf{Estimation}}
                         & \rotatebox{\methodRotate}{\textbf{Perception}}
                         & \rotatebox{\methodRotate}{\textbf{Field Robotics}}
                         & \rotatebox{\methodRotate}{\textbf{Manipulation}}   \\
    \cline{2-8}

    \multirow{6}{*}{
      \rotatebox{90}{
        \parbox[c]{8cm}{\Large\centering\textbf{Climate Domains}}
      }
    }

    \altcolcolors
    {\shortstack{\iconcell{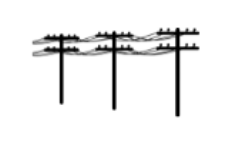}                          \\\rotatebox{\domainRotate}{\textbf{Energy}}}}
    {white} {lightgray}
    {Sec. \ref{energy:autonomy:controls}}
    {Sec. \ref{energy:autonomy:controls}}
    {Sec. \ref{energy:autonomy:estimation}}
    {Sec. \ref{energy:autonomy:perception}}
    {Sec. \ref{energy:autonomy:fieldrobotics}}
    {Sec. \ref{energy:autonomy:fieldrobotics}}

    \altcolcolors
    {\shortstack{\iconcell{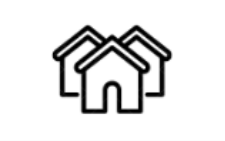}               \\\rotatebox{\domainRotate}{\textbf{Built Environment}}}}
    {white} {lightgray}
    {Sec. \ref{sec:buildings:controls}}
    {}
    {Sec. \ref{sec:buildings:estimation}}
    {Sec. \ref{sec:buildings:perception}}
    {Sec. \ref{sec:buildings:fieldrobotics-and-manipulation}}
    {Sec. \ref{sec:buildings:fieldrobotics-and-manipulation}}

    \altcolcolors
    {\shortstack{\iconcell{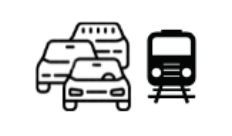}                       \\\rotatebox{\domainRotate}{\textbf{Transportation}}}}
    {white} {lightgray}
    {Sec. \ref{sec:transport:controls-and-planning}}
    {Sec. \ref{sec:transport:controls-and-planning}}
    {Sec. \ref{sec:transport:state-estimation}}
    {Sec. \ref{sec:transport:perception}}
    {Sec. \ref{sec:transport:field-robotics}}
    {}

    \altcolcolors
    {\shortstack{\iconcell{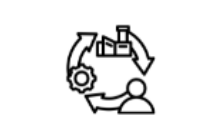}                        \\\rotatebox{\domainRotate}{\textbf{Industry}}}}
    {white} {lightgray}
    {} 
    {Sec. \ref{sec:industry:autonomy:planning-and-manipulation}} 
    {} 
    {Sec. \ref{sec:industry:autonomy:percep}} 
    {} 
    {Sec. \ref{sec:industry:autonomy:planning-and-manipulation}} 

    \altcolcolors
    {\shortstack{\iconcell{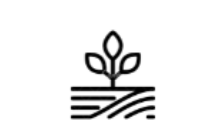}                        \\\rotatebox{\domainRotate}{\textbf{Land Use}}}}
    {white} {lightgray}
    {Sec. \ref{sec:land_use:controls-and-planning}} 
    {Sec. \ref{sec:land_use:controls-and-planning}} 
    {Sec. \ref{sec:land_use:estimation-and-perception}} 
    {Sec. \ref{sec:land_use:estimation-and-perception}} 
    {Sec. \ref{sec:land_use:field-robotics}} 
    {Sec. \ref{sec:land_use:manipulation}} 

    \altcolcolors
    {\shortstack{\iconcell{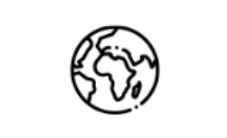}                   \\\rotatebox{\domainRotate}{\textbf{Earth Systems}}}}
    {white} {lightgray}
    {Sec. \ref{sec:earthsciences:controls-planning}}
    {Sec. \ref{sec:earthsciences:controls-planning}}
    {Sec. \ref{sec:earthsciences:state-estimation-and-perception}}
    {Sec. \ref{sec:earthsciences:state-estimation-and-perception}}
    {Sec. \ref{sec:earthsciences:field-robotics-design}}
    {Sec. \ref{sec:earthsciences:manipulation}}
  \end{tabular}
  \caption{
    A visual map of the intersections between climate domains and robotics
    disciplines that are discussed in this paper.  We note that this paper does
    not try to list exhaustively list all possible intersections; see
    \Cref{sec:intro:structure-and-scope,sec:intro:motivation} for discussion on
    how we chose what topics to discuss.
    %
    %
    %
  }
  \label{tab:intro:domain-matrix}

\end{table}

\vspace{-2em}

\titledsubsection{Similar efforts in other communities}

This is not the first work to address interdisciplinary climate-motivated research.
Several other communities have published similar
efforts to identify high-impact research directions in climate, notably machine
learning~\cite{rolnick2022tackling} and control systems~\cite{annaswamy2024control,khargonekar2024climate}.
There is some overlap between these works and ours, particularly where robotics
intersects with machine learning and control systems. These previous efforts
provide a strong foundation and inspiration for our own; our work complements
them by (a) framing the challenges in the context of the broader robotics
community, (b) providing more extensive background on the relevant climate
domains (to help roboticists understand the context for each problem), and
(c) diving more deeply into the technical challenges and specific ways robotics technologies can contribute.


In addition to these broad surveys, there have been several domain-focused surveys covering robotics applications in geosciences~\cite{gil2018intelligent}, environmental monitoring~\cite{dunbabinRobotsEnvironmentalMonitoring2012}, agriculture~\cite{vougioukasAgriculturalRobotics2019}, and forestry~\cite{oliveiraAdvancesForestRobotics2021}, among others. This paper complements these domain-specific works by highlighting areas where core robotics capabilities can contribute across different domains (e.g. automated inspection systems with applications in energy, the built environment, and industrial sectors). We also note that while many researchers focus specifically on various application area (e.g. ocean or agricultural roboticists), many researchers see themselves primarily as experts in core robotics disciplines like perception, controls, or planning rather than focusing on any one application. We also believe that the unique structure of this paper (shown in Table~\ref{tab:intro:domain-matrix}) will help roboticists connect their work in each of these core robotics disciplines to impactful climate-relevant problems.





\newpage
\tableofcontents
\pagebreak

%
%
%
%
%
%
%

\titleformat{\paragraph}[block] 
{\normalfont\normalsize\bfseries} 
{} 
{0em} 
{} 

\autsection{Energy}{Charles Dawson and Lauren\c{t}iu L. Anton}
\label{sec:energy}

\begin{boxMarginLeft}
      Access to clean, affordable, and reliable energy is fundamental to modern society. Primary energy is consumed mainly for electric power, direct heating, and transportation, each requiring deep decarbonization for lasting climate stability. Electrification is central to this transition, driving increased demand for electrical power infrastructure. At the same time, improvements in geothermal and district heating systems are driving new infrastructure development for decarbonized heating needs. This section introduces key stakeholders in electric power and thermal energy systems, challenges for decarbonization, and opportunities for robotics and autonomy to support climate mitigation and adaptation, including:
      \begin{itemize}
            \item \emph{Controls \& Planning}: Safely integrating distributed energy resources (DERs) like solar, and storage into the grid while optimizing grid balancing, load management, and thermal networks.
            \item \emph{Estimation}: Improving visibility into electrical and thermal networks through enhanced state estimation and fault detection.
            \item \emph{Perception}: Enabling autonomous inspection of substations, power lines, and pipelines, with mapping and sensor fusion for real-time monitoring of underground infrastructure.
            \item \emph{Field Robotics \& Manipulation}: Streamlining infrastructure construction and maintenance, including solar installations, geothermal inspections, underground cables, and district energy systems.
      \end{itemize}
\end{boxMarginLeft}

\begin{figure}[b!]
      \centering
      \includegraphics[width=\linewidth]{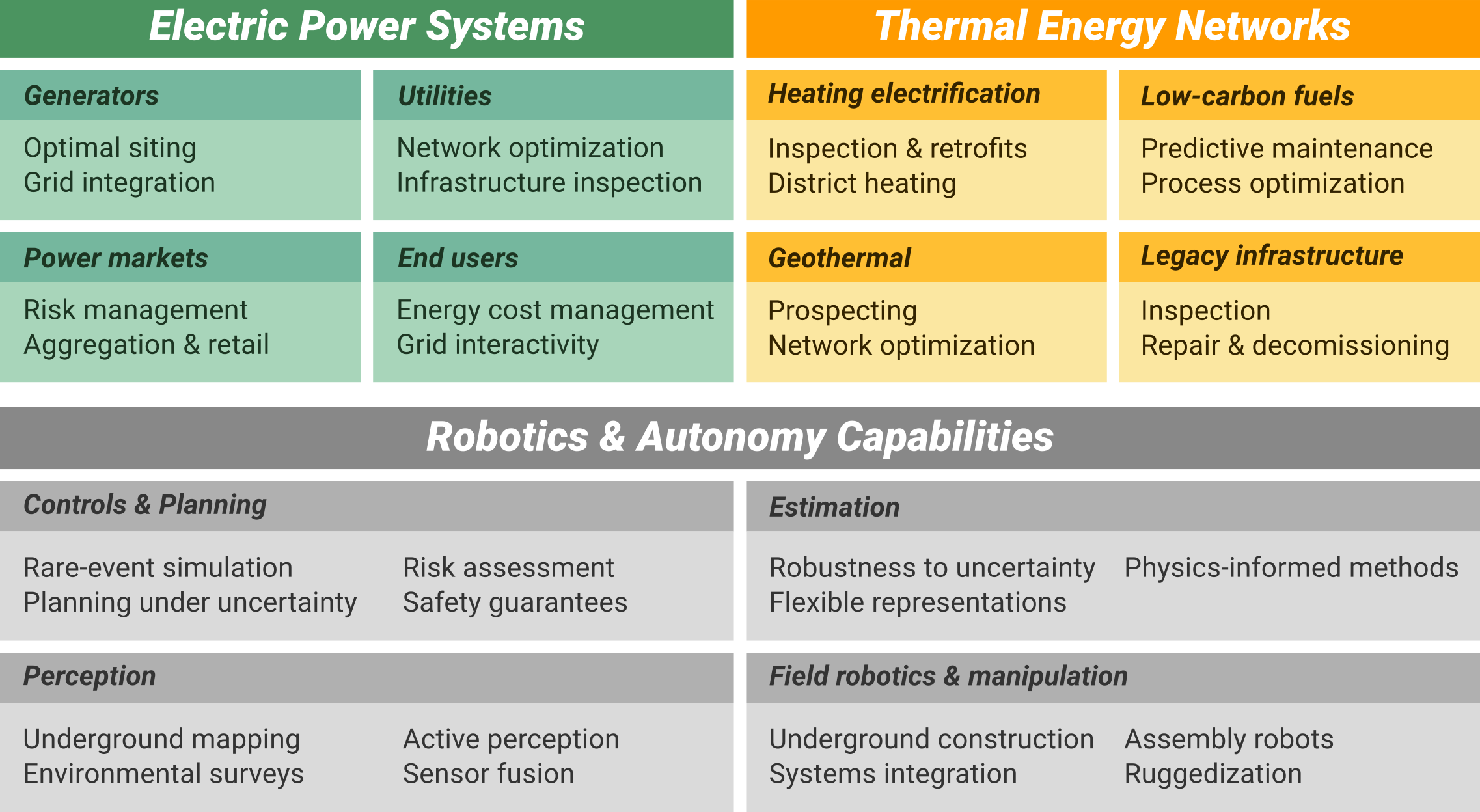}
      \caption{This section summarizes the challenges of energy sector decarbonization, covering challenges in both electric power and thermal energy systems, and provides a roadmap for autonomy and robotics researchers to contribute.}
      \label{energy:fig:headline}
\end{figure}


\newpage

\begin{figure}[tb!]
      \centering
      \includegraphics[width=\linewidth]{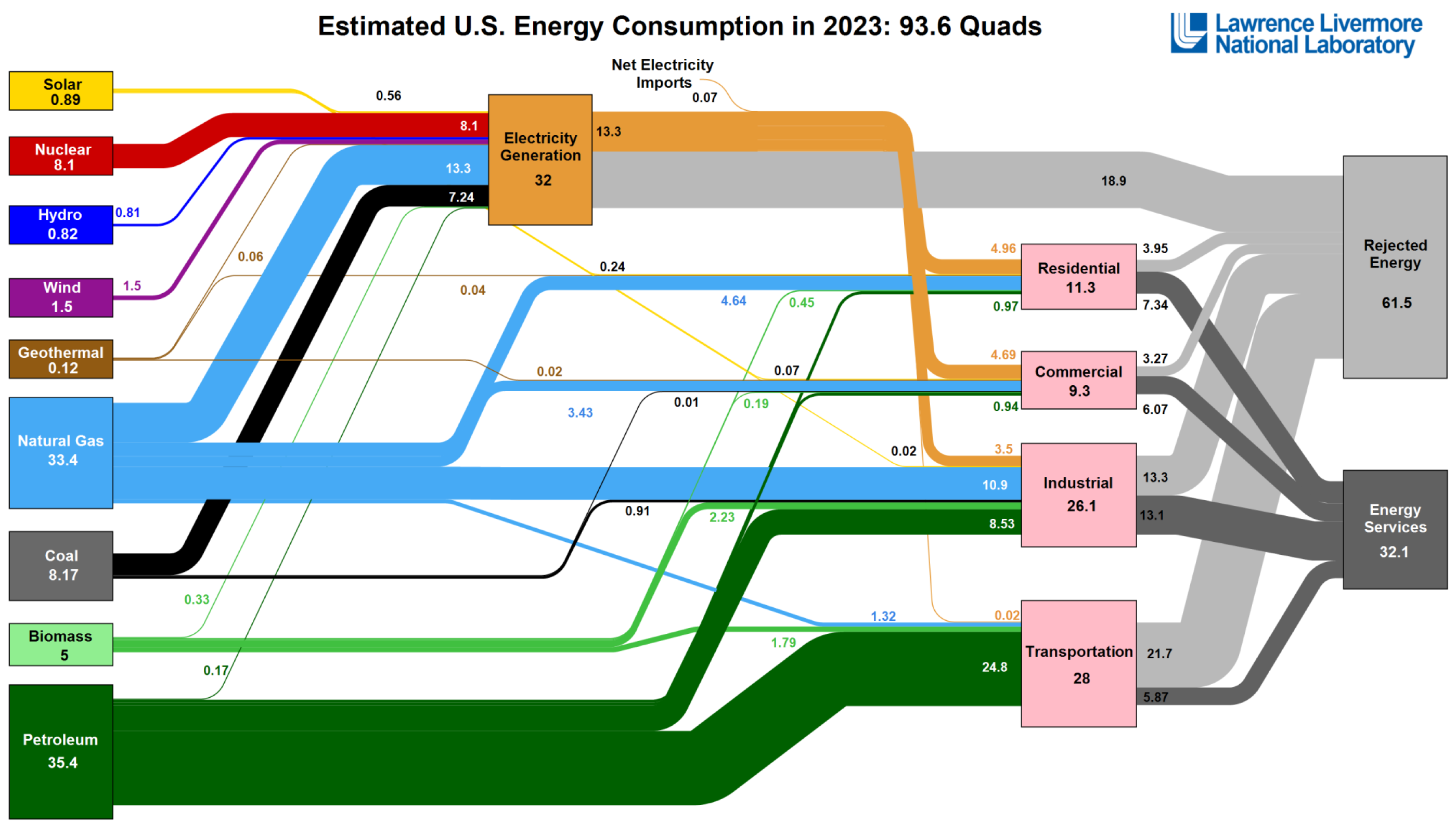}
      \caption{
            Energy flow diagram from primary energy source to end use~\cite{lawrencelivermorenationallaboratoryEstimatedUSEnergy}. Heating and cooling represents the bulk of residential energy use. Major commercial and industrial uses also include refrigeration and industrial process loads.
      }
      \label{energy:fig:us-energy-2023}
\end{figure}

Modern life is incredibly energy-intensive, with most of this energy currently coming from fossil fuels~\cite{smilEnergyCivilizationHistory2017}. While fossil fuels have enabled unprecedented economic growth, it has also driven anthropogenic climate change~\cite{IPCC_2022_WGIII_SPM}. Addressing climate change requires a global transition to low-carbon energy sources while simultaneously meeting rising energy demands in developing nations~\cite{owid-energy}. To give a sense of scale, Figure \ref{energy:fig:us-energy-2023} illustrates the flow of energy in the U.S. economy. Most energy is either converted to electricity or consumed directly for transportation or as heat in industrial processes and buildings. Decarbonizing both electric power and thermal energy systems is thus crucial for reducing the global economy's dependence on fossil fuels.\footnote{Transportation is covered separately in \Cref{sec:transport}.}

\emph{Electric power systems} currently account for 20\% of global energy consumption~\cite{owid-energy}. Decarbonizing this sector requires both ``electrifying everything'' (shifting end uses like heating and transportation from fossil fuels to electricity) and replacing emitting power plants with clean alternatives. These changes will require substantial grid expansion to serve newly-electrified uses as well as real-time optimization and resilient infrastructure to integrate variable renewable energy sources with the grid. At the same time, the power sector must also adapt to a changing climate, including extreme heat that strains grid capacity and severe weather that threatens infrastructure, while needing to modernize aging transmission and distribution networks.

\emph{Thermal energy systems} currently serve more than twice the demand of electricity, representing roughly 50\% of global final energy use and 38\% of energy-related CO$_2$ emissions~\cite{iea2024renewables}. Decarbonizing thermal energy involves three strategies: electrifying where feasible, adopting low-carbon fuels where needed (for hard-to-electrify cases), and expanding geothermal systems. In most cases, low- and medium-temperature heat can be electrified using heat pumps and electric boilers, but many high-temperature industrial processes (such as cement, steel, and glass) are difficult to electrify. Instead, these use cases may require zero-carbon fuels like hydrogen, biomass, and synthetic fuels. Geothermal and district energy systems can complement both categories: reducing strain on the electric grid for low- and medium- temperature applications and providing high-temperature heat for industrial applications. With many technology applications emerging, innovation to scale-up and drive down costs is needed.

Sections~\ref{energy:subsec:electric_power_systems} and~\ref{energy:subsec:thermal-energy-networks} give an overview of electric and thermal energy systems. Within each sector, we identify key stakeholders, discuss challenges for decarbonization, and identify opportunities for high-impact research. After this overview, Sections~\ref{energy:autonomy:controls}--\ref{energy:autonomy:fieldrobotics} provide an in-depth roadmap for future research in the robotics and autonomy community, identifying opportunities for research in controls, planning, estimation, perception, and field robotics to support decarbonization and adaptation in the energy sector.

\titledsubsection{Overview of Electric Power Systems}
\label{energy:subsec:electric_power_systems}


Economy-wide decarbonization requires both electrification (shifting end uses from fossil fuels to electricity wherever possible) and replacing emitting power plants with clean energy sources. This transition places substantial demands on the electric power sector, requiring not only deep decarbonization but also substantial growth to meet demand from newly-electrified heating, transportation, and industrial processes. At the same time, the structure of the industry is changing---moving from large, centralized fossil-fueled power plants to decentralized networks of smaller distributed energy resources (DERs) like solar photovoltaics (PV), wind, battery storage, and flexible demand. This trend towards decentralization presents a challenge for grid planning, operations, and resilience.

\begin{figure}[tb]
      \centering
      \includegraphics[width=\linewidth]{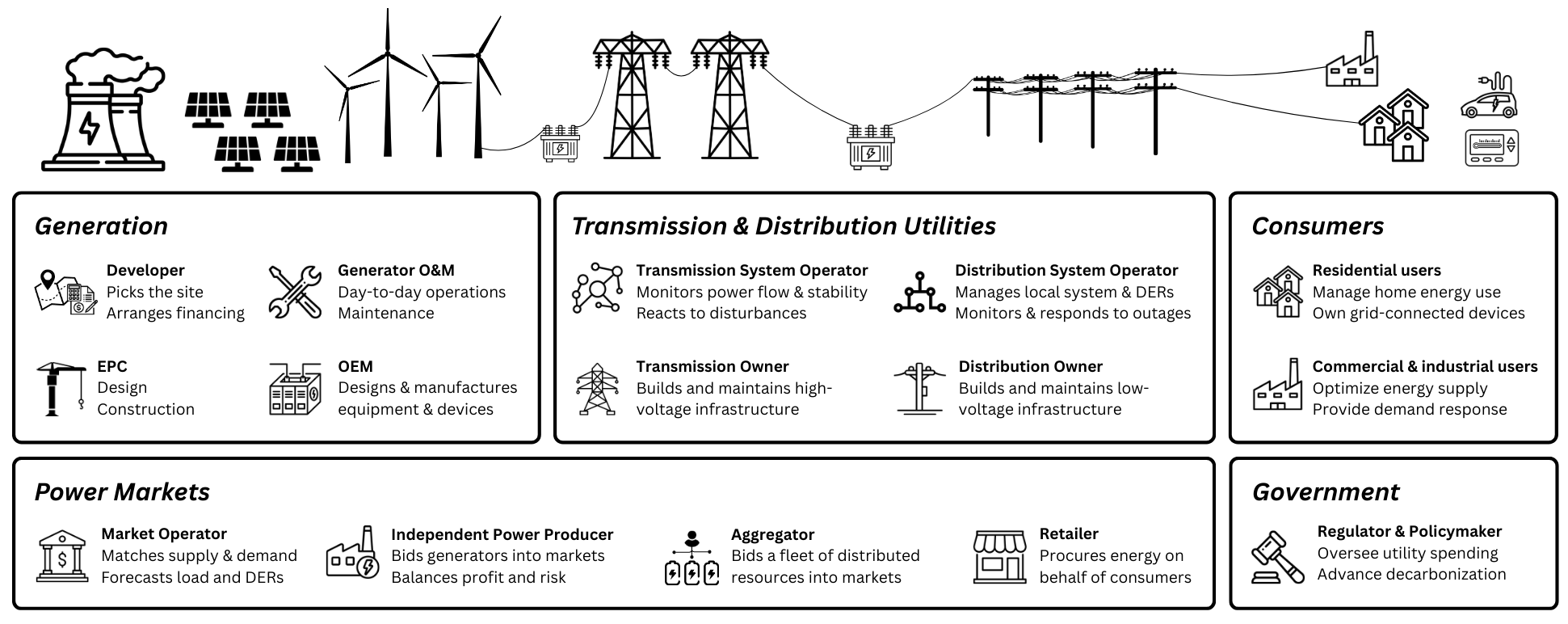}
      \caption{A diagram of the electric power sector, showing both physical infrastructure (generation, transmission, distribution, and end uses) and key stakeholders. Understanding the roles and motivations of these different stakeholders will help increase the impact of research in this field.}
      \label{energy:fig:stakeholder_map}
\end{figure}

\begin{wrapfigure}{r}{0.45\textwidth}
      \includegraphics[width=\linewidth]{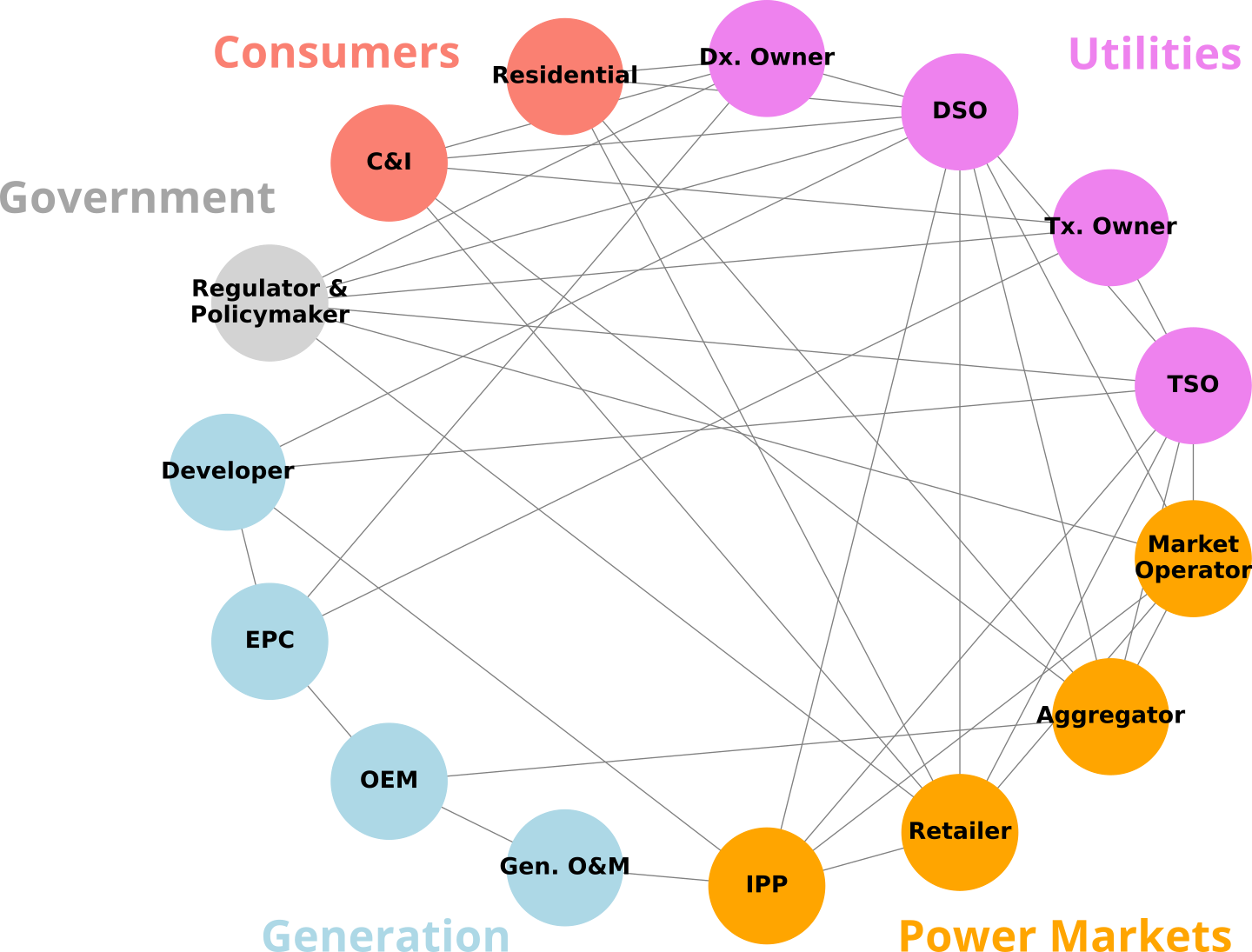}
      \caption{An illustration of the complex interactions between power systems
            stakeholders.}
      \label{energy:fig:stakeholder_network}
\end{wrapfigure}

The electric power sector is notoriously complex, with many interacting stakeholders and a mix of market-based mechanisms and regulatory oversight that varies by region. These stakeholders include independent system operators (ISOs) that manage power on the high-voltage transmission system, utilities that own and operate grid infrastructure, aggregators that coordinate DERs, and regulatory bodies that enforce market rules and emissions targets. Figure~\ref{energy:fig:stakeholder_map} provides a visual summary these stakeholders, and Figure~\ref{energy:fig:stakeholder_network} summarizes the interactions between them.

Understanding the structure of the electric power sector will help researchers maximize the impact of their work. There are opportunities for robotics and autonomy to contribute throughout the electric power sector, but different applications will require working with different stakeholders (for example, power plant developers for construction robots and grid operators for real-time network optimization).

To highlight where robotics and autonomy research can drive innovation, this
section introduces five core functional groups of the electric power sector:
\emph{Generation} (planning, building, and operating power plants),
\emph{Grid Operations} (maintaining power balance and reliability),
\emph{Power Markets} (trading electricity in real-time, day-ahead, and
long-term markets), \emph{Consumers} (including residential, commercial, and
industrial), and \emph{Regulation \& Policy} (overseeing energy markets and
utility investment decisions).

\titledsubsubsection{Generation}
\label{energy:subsubsec:generation}

In the power sector, climate change mitigation starts with decarbonizing generation: replacing power plants that emit greenhouse gases with non-emitting sources like wind, solar, hydropower, geothermal, and nuclear~\cite{ipcc-spm}. Replacing fossil fuels while simultaneously increasing global capacity to serve electrified end uses and improve the quality of life in developing nations will place a substantial burden on the generation sector. Electricity generation will need to quadruple to support economy-wide decarbonization in the United States alone~\cite{larson2021netzero}. While necessary, this shift introduces new technical, logistical, and economic challenges~\cite{iea2024renewables}.
By supporting the development and operation of new power plants, autonomy and controls researchers can accelerate and reduce the cost of this transition.

\paragraph{Key Stakeholders}

The lifecycle of a power plant involves a range of specialized stakeholders.
\emph{Developers} initiate projects by finding promising sites and arranging
financing~\cite{nilsonSurveyUtilityScaleWind2024}. \emph{Engineering,
procurement, and construction} (EPC) firms manage physical design, construction,
and commissioning. \emph{Operations \& maintenance} (O\&M) teams are responsible
for day-to-day inspections and maintenance to keep the plant running at peak
performance~\cite{renOffshoreWindTurbine2021}. \emph{Original equipment
manufacturers} (OEMs) design and manufacture major components like wind
turbines, inverters, and battery packs; provide field support; and make sure
that their products meet performance and grid compatibility standards.
Generation owners are also responsible for selling electricity in power markets;
in this context, they are known as independent power producers (IPPs), discussed
further in Section~\ref{energy:subsubsec:power-markets}.

\paragraph{Challenges \& Opportunities}

Developers typically manage a portfolio of potential projects, of which only a subset will typically be completed (e.g., due to permitting challenges or a lack of available grid capacity~\cite{josephrandQueued2024Edition2024,SpeedingSolarProjects}). Developers seek to reduce risk for new projects, and must sift through large amounts of data from land records, utility data, and environmental reviews to find suitable prospects. Robotics and autonomy can help by enabling automated environmental surveys and optimizing site selection.

EPC firms manage the later stages of a project, and are concerned with executing on-time and on-budget, which can be challenging on large or remote construction sites. EPCs can benefit from design automation and robotic inspection to save time and keep projects running smoothly. They can also benefit from construction and assembly robots to improve efficiency and automate dull, dirty, and dangerous tasks like ``racking'' solar panels (lifting and installing heavy panels~\cite{builtroboticsAutonomousSolarPiling, aesAESLaunchesFirst}). We note, however, that grid integration, not construction, is currently the bottleneck for deploying renewable generation at scale~\cite{doeofficeofpolicyQueuedNeedTransmission2022}.

Once a facility is operational, O\&M teams are responsible for maximizing power output and avoiding outages (often working closely with OEMs). O\&M teams would benefit from research in predictive maintenance and remotely-operated or autonomous inspection and maintenance robots~\cite{renOffshoreWindTurbine2021}.

OEMs play a role throughout the project lifecycle, from technology development to manufacturing and field maintenance. They must bring new technologies to market while ensuring their products meet safety and performance standards~\cite{ieee1547-2018,ul1741-2021}. Research on safety certification, predictive maintenance, and use cases like inverter control and battery management can help OEMs deliver innovative, reliable products.



\titledsubsubsection{Grid Operations}
Delivering electricity from generators to consumers requires careful planning and operation of grid infrastructure. Responsibility for grid operations can vary widely between jurisdictions. For example, some regions rely on vertically integrated utilities to manage the entire system (including owning and operating generation), while others split responsibility between the utility that owns electrical infrastructure and an independent grid operator. Operators in all jurisdictions face challenges in managing aging infrastructure and increasing deployment of DERs. Robotics and autonomy can help address these challenges by making electric networks more observable, adaptable, and resilient.

\paragraph{Key Stakeholders}
The electric grid is divided into high-voltage transmission and low-voltage
distribution networks, reflecting both functional and organizational boundaries.
\emph{Transmission System Operators} (TSOs) must ensure that generation and
demand are balanced in each moment and that the system can recover from
unexpected faults (e.g., through an ``N-1'' analysis that considers all possible
single failures). \emph{Distribution System Operators} (DSOs) are responsible
for regulating the voltage on distribution circuits, coordinating with DERs like
rooftop solar, batteries, and electric vehicles (EVs), and responding to storms
and other outages. In some jurisdictions, the transmission operators are
separate (``unbundled'') from \emph{Transmission Owners} (TOs) who maintain
high-voltage assets. In contrast, distribution systems are almost always owned
and operated by a single entity.

In power systems with a competitive market for electricity, market coordination may occur within the TSO or it may fall to dedicated entities. We discuss these financial roles in detail in Section~\ref{energy:subsubsec:power-markets}.

\paragraph{Challenges \& Opportunities}
While institutional structures vary across jurisdictions, the physical challenges of operating and maintaining the grid are broadly the same.
%
Transmission systems face challenges in both operations and planning. In operations, TSOs must keep the system balanced in the face of highly variable energy production from wind and solar. Increased uncertainty means that TSOs will need improved tools for scheduling generators, managing power flows, and responding to faults, drawing on research in decentralized control, optimization, and reinforcement learning~\cite{molzahnSurveyDistributedOptimization2017,chenReinforcementLearningSelective2022}.
On the planning side, TSOs are under pressure to quickly connect new generation to the grid, but doing this safely requires labor-intensive engineering studies. Research in optimization under uncertainty, rare-event simulation, and safety certification can help grid operators safely interconnect new generators.

Distribution systems present a distinct set of challenges, especially for observability and DER integration. Unlike TSOs, most DSOs operate with limited sensor coverage and incomplete knowledge of the state of their networks~\cite{dehghanpourSurveyStateEstimation2019}. As a result, graph-based and physics-informed inference research could help improve visibility in distribution networks. Additionally, while DSOs have historically been fairly passive, increased deployment of rooftop solar, batteries, and EVs will require DSOs to take a more active role in managing voltage and power flow. Tools for real-time DER coordination, such as decentralized control using inverters, are promising but have not yet been validated at scale~\cite{molzahnSurveyDistributedOptimization2017}. Another challenge for DSOs is fault detection and restoration, which is often still a manual process. Autonomy research can support automated fault detection, isolation, and restoration (FDIR) by coordinating field sensors and relays~\cite{zidanFaultDetectionIsolation2017}. Controls and planning research can support real-time reconfiguration by finding optimal switching plans in complex networks, reducing outage duration.

Both transmission and distribution operators face the burden of maintaining aging infrastructure. Many utilities manage assets installed decades ago, often with incomplete digital documentation. Robotics can help automate inspections (particularly for dangerous, energized equipment) and vegetation clearance, as well as assisting in recovery after storms and fires~\cite{xcelenergyHearingExhibit1022020,pacificgas&electric2024RDStrategy}. Subsurface mapping technologies and special-purpose robots such as pipe crawlers can also help inspect underground infrastructure~\cite{pacificgas&electric2024RDStrategy}. Over time, these tools could evolve into autonomous maintenance platforms capable of both monitoring and repairing faulty equipment.

\titledsubsubsection{Power Markets}
\label{energy:subsubsec:power-markets}

Historically, the grid was operated by vertically integrated utilities that owned everything from power plants to the meters on customers' houses.
Today, many regions have introduced markets in which independent generators compete to produce electricity at the lowest cost. No two electricity markets are identical, and their structures reflect local policy choices, regulatory frameworks, and legacy infrastructure. For example, many regions in the US combine the TSO and market operator in a single entity (an ISO), while many European markets maintain separation between markets and transmission operation.
%
Markets typically include day-ahead and real-time energy markets, capacity markets (which pay generators for availability during peak periods), and ancillary services markets (for frequency regulation, reserves, and voltage support). Electricity can also be sold outside of wholesale markets via long-term bilateral contracts known as power purchase agreements (PPAs).

As power markets grow more complex, driven by increasing uncertainty from renewables and the emergence of DERs, market participants will need tools that support faster, more informed decision-making. Autonomy research can help provide these tools, enhancing performance, improving scalability, and reducing volatility in power markets.
These capabilities will not replace the need for regulatory changes, but can extend the capacity of existing market structures to handle greater complexity and support the integration of low-carbon and flexible energy resources at scale.

\paragraph{Key Stakeholders}
\emph{Market operators} or \emph{power exchanges} (PXs) forecast demand,
administer auctions, and enforce rules to ensure reliability and fair
competition. \emph{Independent Power Producers} (IPPs) participate by bidding
generation assets into these markets (optimally balancing revenues from selling
energy, capacity, and ancillary services) and may also sell electricity via
long-term PPAs.
\emph{Retailers} act as intermediaries, purchasing energy from wholesale markets
and offering contracts to customers that are typically more price stable than
market clearing prices. \emph{Aggregators} control fleets of up to thousands of
small DERs like batteries, solar PV, and EVs, bidding these resources in markets
as if they were a single, large entity (known as a virtual power plant or VPP).

\paragraph{Challenges \& Opportunities}

Unlike most commodity markets, power markets are very tightly coupled to the underlying physics of the grid. Market operators must respond to dynamic system conditions and manage bids across multiple timescales to reliably balance supply and demand~\cite{cainHistoryOptimalPower2012}.
As variable renewable energy becomes more common, traditional forecasting and optimization tools will need to improve~\cite{hongProbabilisticElectricLoad2016}.
While physical robots are less relevant for power markets, the computational tools of autonomy research can help enhance market workflows through improved forecasting and system identification, rare-event prediction (e.g., for periods of extreme demand), and anomaly detection to identify market manipulation. Intelligent market-clearing engines that adapt to changing participant behavior and resource constraints could reduce errors and increase robustness under uncertainty.

IPPs seek to maximize profits and face the challenge of managing physical constraints while bidding into multiple markets (day-ahead energy markets, real-time balancing, long-term PPAs and capacity markets, and ancillary services). Market volatility, regulatory changes, and unexpected outages introduce further risk. Autonomy-enabled tools can support decision-making by simulating market outcomes under uncertainty and optimizing bids based on forecast price and availability scenarios; for example, using reinforcement learning, stochastic programming, or probabilistic revenue modeling~\cite{caoDeepReinforcementLearningBased2020}.
Aggregators play a similar role to IPPs, but rather than managing a small number of large power plants, they manage up to thousands of distributed resources~\cite{jenniferdowningPathwaysCommercialLiftoff2023}. Aggregators need to estimate the state of each resource under their control, forecast aggregate behavior, and design optimal control strategies. They must also balance their bidding strategy with consumer preferences (e.g., shifting when an EV charges to reduce load on the grid, but making sure it is fully charged each morning). Robotics and autonomy can support distributed estimation and control algorithms, particularly learning-based methods that can scale to thousands of assets and align market participation with constraints from user preferences.

Retailers often operate with thin profit margins and must hedge against wholesale price volatility while offering attractive prices to customers. In addition to sharing some of the uncertainty-modeling needs of IPPs, retailers could also benefit from research on modeling customers' energy behavior (for instance, by estimating cooling needs for a house) and optimizing flexible loads~\cite{zhangOptimalLearningBasedDemand2016}.

\titledsubsubsection{Consumers}
Consumers of electricity have historically been passive, with little participation in the electric power system beyond paying a monthly bill. Today, that model is changing. Flexible loads, electric vehicles, and behind-the-meter generation are giving customers new choices and responsibilities in how they manage energy. Many consumers now receive time-varying pricing that reflects the cost of generating and transmitting energy at different times of day, participate in demand response programs to reduce consumption during peak hours to support grid reliability, and install backup generation and storage for resiliency. Autonomy researchers can help consumers take on these new roles by simplifying and automating energy management, enabling homes and businesses to act as self-optimizing, resilient participants in decarbonized energy systems.


\paragraph{Key Stakeholders}
Electricity consumers differ widely in how they use energy and interact with the grid. Residential, commercial, and industrial customers each have unique challenges and opportunities for autonomy and robotics research.
\emph{Residential customers} connect at low voltage and rely on utilities or
retailers to supply their electricity. Their energy use is driven by comfort,
convenience, and cost, and increasingly includes flexible loads such as HVAC
systems, EVs, smart appliances, rooftop solar, and battery storage.
%
\emph{Commercial customers} include offices, retail buildings, campuses,
hospitals, and other facilities with moderate power demand. These customers are
sensitive to energy costs and often receive more complex billing (e.g. peak
demand charges). Their decision-making is guided by facility managers or energy
service companies who seek to maximize cost savings without disrupting business
operations.
%
\emph{Industrial customers} operate energy-intensive facilities such as
factories, warehouses, and data centers. Their loads include process equipment,
motors, refrigeration, and fleet-scale EV charging for both passenger vehicles
and freight. These customers often have dedicated energy management teams, take
part in wholesale markets and long-term PPAs, and may operate behind-the-meter
microgrids.

\paragraph{Challenges \& Opportunities}

Residential customers typically lack the time, expertise, or interest to manage their energy use manually. Coordinating HVAC, EV charging, and appliance use under time-varying rates or demand response programs requires real-time optimization and personalized adaptation~\cite{jenniferdowningPathwaysCommercialLiftoff2023}. Autonomy research can enable home energy management systems (HEMS) that learn user preferences, predict occupancy, and dynamically coordinate device loads~\cite{beaudin2015hemsreview,shareef2018reviewhems}. Opportunities include adaptive control of thermostats and water heaters, automated EV charging schedules, and energy monitoring with fault detection. Systems must be low-maintenance, interpretable, and robust to network failures and user overrides.

Commercial customers must balance business needs and cost, often across multiple buildings in a portfolio. Automation can support predictive control of HVAC and lighting systems, as well as real-time fault detection~\cite{Shaikh2014review}. Robotic and autonomy tools can also assist with retrofitting legacy equipment or integrating solar PV and batteries into building energy management systems. Large-scale EV charging infrastructure in parking lots presents an emerging area where coordinated scheduling and dynamic pricing can benefit from autonomy-driven dispatch and queuing algorithms~\cite{huElectricVehicleFleet2016}.

Like commercial customers, industrial users must optimize their business' energy use, but the need to integrate with high-power equipment, unique business and operational constraints, and participation in wholesale markets complicates this task. Autonomy researchers can support multi-timescale optimization for load scheduling, predictive maintenance, and demand management for electrified process loads. Distributed control and reinforcement learning can support large-scale EV fleets and microgrids~\cite{huElectricVehicleFleet2016}.

Across all segments, retrofitting is a practical barrier to adoption. Robotic solutions can help with installing and maintaining meters, smart devices, and other infrastructure. By focusing on customer-centered automation, researchers can unlock a major source of system flexibility while improving reliability and usability for end users.

\titledsubsubsection{Regulation \& Policy}
It is impossible to separate the electric power system from its regulatory context. Regulators determine how markets are structured and how utilities serve the public. As the electric power sector changes in response to new technologies and decarbonization goals, regulators and policymakers must evolve by adapting new rules, standards, and incentives. While robotics and autonomy are not themselves regulatory tools, they can play a critical supporting role in helping institutions govern more effectively and in designing technologies that are easier to regulate.


\paragraph{Key Stakeholders}
Electric power regulation spans multiple layers of governance. Institutional
roles may differ across jurisdictions, but they often follow common functional
categories. For example, in the United States, \emph{federal entities} like
the Federal Energy Regulatory Commission (FERC) oversee wholesale markets,
transmission planning, and interconnection rules, while state-level
\emph{public utility commissions} (PUCs) regulate utilities, distribution
systems, and retail pricing. In Canada, responsibilities are primarily
provincial, with agencies like the Ontario Energy Board or the Alberta Utilities
Commission fulfilling both retail and wholesale regulatory functions. Across the
European Union, regulatory oversight is shaped by \emph{national energy
regulators} under the coordination of ACER (Agency for the Cooperation of Energy
Regulators), with TSOs governed by EU-wide codes and policies such as the Clean
Energy Package. In all three regions, additional roles are played by
\emph{legislative bodies} (setting mandates), \emph{system planners}
(conducting long-term assessments), and \emph{environmental or permitting
agencies}. Despite structural differences, regulators everywhere share the task
of enabling secure, affordable, and decarbonized electric power systems.


\paragraph{Challenges \& Opportunities}
Regulators must make policy decisions under deep uncertainty, often without clear market signals or data from emerging technologies~\cite{fremethInformationAsymmetriesRegulatory2012}. Evaluating the impacts of DER integration and flexible loads relies on scenario-based planning tools that have limited ability to model consumer behavior~\cite{massachusetts_cecp_2022}. While utilities, not regulators, are responsible for grid planning and interconnection, there is a lack of transparent, verifiable, and accessible models for regulators who seek to understand and audit utilities' planning decisions.

Robotics and autonomy research can help bridge these gaps by expanding regulatory visibility and enhancing institutional capacity. Utilities can enhance auditing and compliance functions with autonomous drones and sensor systems that verify vegetation management, equipment installation, or storm recovery performance in the field. Interpretable prediction and optimization systems could support regulators in assessing DER control strategies, market participation logic, or grid reliability impacts~\cite{forresterStateRegulatoryOpportunities2025}. Simulation tools that incorporate autonomous DER behavior and uncertainty can help agencies explore tradeoffs in rate design, resilience targets, market reform (e.g., resource adequacy), or decarbonization pathways~\cite{massachusetts_cecp_2022,knightMassachusettsTechnicalPotential2023,mettetalChargingForwardEnergy2023}. For example, emissions-aware dispatch algorithms or equitable load-shedding schemes could be assessed using autonomy-enabled models before adoption into code or standards.

Regulatory approval hinges on systems being not only safe and effective, but also understandable. While approval processes vary --- from technology certifications in the EU to rate case reviews in North America --- there is universal need for automated systems to be explainable and auditable. Frameworks that embed transparency, auditability, and alignment with public policy goals will be better positioned for adoption and scale.



\titledsubsubsection{Electric power systems summary}
Decarbonizing the electric power system requires re-imagining how electricity is generated, delivered, consumed, and governed. This transition will introduce significant technical and institutional complexities, with challenges emerging throughout the sector. Generators must be sited, built, and maintained at unprecedented scale, often in remote locations or with limited workforce capacity. Transmission and distribution networks must accommodate dynamic power flows, integrate millions of distributed devices, and remain resilient to climate-driven disruptions. Markets must evolve to reflect increasingly granular and stochastic behavior, while enabling fair and efficient participation from both large producers and distributed energy resources. Customers, once passive recipients, are now central actors who can use automation to manage flexible loads, onsite generation, and energy resilience. Meanwhile, regulators face growing pressure to assess new technologies, evaluate systemic risks, and design incentives that align private innovation with public goals.

Robotics and autonomy research can support this transformation by addressing core needs across stakeholder groups. Field robotics can accelerate construction, inspection, and maintenance across infrastructure classes. Perception and estimation algorithms can improve grid monitoring, customer device coordination, and predictive asset management. Intelligent control can optimize load patterns, DER coordination, and power system operations in real time. Autonomous simulation and planning tools can support both market participants and policymakers in navigating uncertain futures. Together, these tools can help power systems operate more flexibly, more safely, and at lower cost---expanding the frontier of what is technically and institutionally feasible in the clean energy transition. Table~\ref{energy:tab:stakeholder_research_map} summarizes these research opportunities, which we explore in greater depth in Sections~\ref{energy:autonomy:controls}--\ref{energy:autonomy:fieldrobotics}.

\titledsubsection{Overview of thermal energy networks}
\label{energy:subsec:thermal-energy-networks}

Nearly 50\% of all global energy is used as heat~\cite{iea2024renewables}, either at low temperatures for heating buildings (so called ``low grade'' heat), medium temperatures for certain industrial processes like food processing (``medium grade''), and high temperatures for applications like steel, glass, and cement (``high grade''). Today, this heat is delivered by thermal energy systems that exist in parallel to (and sometimes overlapping with) the electric power system. Thermal energy can be delivered by a variety of different carriers, each with its own network (e.g., fuels like natural gas and hydrogen in pipelines, electricity in the grid, steam and hot water in geothermal wells and district heating loops). Decarbonizing these thermal networks is essential for climate change mitigation but faces unique challenges. This section provides an introduction to these challenges and opportunities for robotics and autonomy to contribute across three types of thermal networks: electric thermal, fuel, and geothermal systems.

\titledsubsubsection{Electric thermal systems}\label{energy:thermal:electric}

Electrification is the first choice for decarbonizing many low- and medium-temperature applications. In buildings, electric heat pumps can provide both heating and cooling, and electric boilers can provide medium-temperature steam for many industrial processes. Even some high-temperature uses may be electrified; for example using electric arc furnaces or kilns for making steel and glass, respectively. Where feasible, electrifying these applications allows us to piggy-back on efforts to decarbonize the electric power system as a whole, dramatically reducing greenhouse gas emissions from heat~\cite{larson2021netzero}. In residential applications, replacing combustion with electric heating can reduce indoor air pollution and improve quality of life. By helping to reduce installation costs and improve efficiency, robotics and autonomy can contribute to faster adoption, making electric heating and cooling more accessible and cost effective.

\paragraph{Key Stakeholders}
Electric thermal systems involve multiple stakeholders, from equipment manufacturers to building operators.
\emph{HVAC manufacturers} produce electric heat pumps, electric boilers, and
thermal storage technologies for residential and commercial buildings.
\emph{Industrial equipment providers} develop high-temperature electric
furnaces, electric kilns, and process heaters for industrial applications.
\emph{Building managers} and \emph{energy service companies} (ESCOs) are
responsible for deploying these technologies  in customer facilities to reduce
energy costs and emissions. In larger installations such as industrial parks,
campuses, and some cities, \emph{district heating operators} are emerging to
centralize electric heating equipment to serve multiple buildings.

\paragraph{Challenges \& Opportunities}

Installation complexity remains a barrier for many electric heating technologies, particularly for industrial-scale electric boilers and heat pumps that require precise alignment of high-pressure piping, thermal insulation, and valve systems~\cite{williamgoetzlerGuidanceDocumentSpace2024}. Autonomous pipe-fitting and welding robots can ensure millimeter precision during installation, reducing leaks. Thermal lining applicators with robotic arms can uniformly coat surfaces with high-temperature insulation, reducing thermal losses.

Maintenance and real-time monitoring can prolong the lifespan and performance of online equipment. UAV-based thermal imaging can be used as a non-invasive method to quickly scan for thermal leaks, enabling targeted retrofits (discussed further in \Cref{sec:built-environment} on the built environment). Autonomous sensor networks can monitor the health of electric heating equipment, improving the implementation of protective maintenance, while reducing downtime~\cite{zhangDataDrivenMethodsPredictive2019}. Likewise, robotic maintenance crawlers with sensors can autonomously navigate underground and difficult to reach equipment to identify equipment conditions.

In the case of larger district energy networks that use electric boilers and heat pumps, self-healing control systems can be implemented to dynamically reroute energy flows in the event of a blockage or leak, minimizing losses and service disruptions~\cite{kuntuarovaDesignSimulationDistrict2024}. HVAC systems of all sizes can also benefit from controllers that adapt to the thermal properties of a building to improve efficiency and reduce demand on the electric grid.

\titledsubsubsection{Fuel systems}

While electrification is advancing rapidly, fuels (i.e. chemical energy carriers) continue to play an important role in decarbonization. Fuels will be particularly important in uses where electrification is not currently economically or technically feasible, such as high temperature heat, long-haul aviation and shipping, and long-duration energy storage~\cite{iea2023efuels}. While nearly all fuels used today are fossil fuels, future low- or zero-emission fuels include biogas, hydrogen, and ammonia. In the long term, there are opportunities for robotics and autonomy to help reduce costs for novel low-emissions fuels, supporting decarbonization in hard-to-electrify applications.
In the short term, there are benefits from using robotics and autonomy to maintain legacy natural gas pipes. In certain settings, it can be better to extend the life of aging pipes by five to ten years (until more customers have electrified and the pipes can be removed for good) rather than replace them with entirely new investments in fossil fuel infrastructure with a 30 year design life.

\paragraph{Key Stakeholders}
Fuel-based energy systems involve a range of actors across the value chain including
\emph{producers} of biogas, hydrogen, and other fuels, \emph{pipeline operators}
and \emph{utilities} that oversee fuel distribution, and \emph{offtakers},
including both industrial users like cement and chemical manufacturers that rely
on combustion for difficult-to-electrify high-temperature processes and
operators of peaking power plants. As in electric power systems, there is also
an industry of \emph{EPC firms} and \emph{OEMs} who support these primary
players, and government entities who make policies around fuel use and
infrastructure siting.

\paragraph{Challenges \& Opportunities}
In certain applications like high-temperature industrial heat, long-distance transportation by plane or ship, and long-duration seasonal energy storage, fuel remains more practical than electrification. However, to mitigate the climate impact of these fuels, not only must the industry transition to low-emission fuels, but large amounts of fuel infrastructure must be modernized or replaced.

The biggest challenges for this transition is economical production of low-emission alternatives to fossil fuels, despite the recent history of subsidies in some countries~\cite{bernhardlorentzLowcarbonFuelsLast2024}. This is primarily a chemical and industrial engineering problem, rather than an autonomy and robotics problem, but there are ways that roboticists can play a supporting role. Because these producers must keep large, complex facilities running with minimal downtime~\cite{iea2023efuels}, predictive maintenance and autonomous inspection robots could help save costs and improve efficiency~\cite{zhangDataDrivenMethodsPredictive2019}. If production of these fuels takes off, robotics can also assist with constructing and operating the pipeline infrastructure to carry these fuels (in many cases, existing natural gas infrastructure cannot be easily repurposed due to material incompatibility, embrittlement, and risk of leaks~\cite{telessyRepurposingNaturalGas2024}).

Decarbonization also poses challenges for pipeline operators and natural gas utilities. In many regions, utilities must spend large sums to replace leaky or unsafe natural gas pipe. However, the transition from fossil gas to electrification means that these utilities also face the prospect of steeply declining future sales, so any money spent replacing pipe risks becoming a ``stranded asset'' that burdens utility customers for decades to come~\cite{ago2025gsep}. Robotic systems for inspecting and repairing legacy natural gas networks can help utilities avoid spending on entirely new gas infrastructure and bridge the gap to electrification~\cite{ismailDevelopmentInpipeInspection2012,jangReviewTechnologicalTrends2022}. For example, it may be better to repair leaky gas pipe in a residential neighborhood to last five to ten more years while the neighborhood electrifies, rather than replace it with new infrastructure with a 20--30 year service life.

\titledsubsubsection{Geothermal Systems}
Geothermal systems extract thermal energy from beneath the Earth's surface to provide electricity, heating, or cooling. These systems can be categorized into three levels: high-temperature systems for high-grade industrial heat, medium-temperature systems for district heating, and low-temperature systems for direct heating and cooling using heat pumps. High- and medium-temperature systems can also be used to generate electricity. While geothermal is less mature than fuel or electric energy systems (particularly for medium- and low-temperature systems), it is a promising low-carbon solution that can operate continuously and independently of weather, providing a useful counterpoint to renewable energy like wind and solar. However, high capital costs, slow deployment, and site-specific challenges limit adoption~\cite{iea2024geothermal}. Robotics and autonomy research can help reduce installation costs, accelerate subsurface exploration, improve reliability, and optimize heat production across all types of geothermal systems.

\paragraph{Key Stakeholders}
Geothermal projects involve a broad range of stakeholders depending on system
scale. High- and medium-temperature projects typically involve \emph{developers}
who identify promising sites and secure financing, \emph{drilling service
providers} who drill geothermal wells, and supporting \emph{EPC firms} (for
design and construction) and \emph{OEMs} (for turbines and generators).
In low-temperature applications, stakeholders include many of the same
stakeholders involved in heating electrification in
Section~\ref{energy:thermal:electric}, with the addition of \emph{shallow
drilling contractors}. Increasingly, \emph{electric utilities} are interested in
geothermal for electricity generation, while regulations in some jurisdictions
are prompting \emph{natural gas utilities} to apply their workforce and
construction expertise to district heating and networked geothermal systems
(where low-temperature geothermal equipment is shared between buildings in a
neighborhood).

\paragraph{Challenges \& Opportunities}
High-temperature geothermal systems have historically been limited to regions with volcanic or tectonic activity and require deep, high-temperature drilling. These projects face high capital costs and exploratory risk~\cite{osti_1983898}. Robotics can address several pain points: autonomous inspection robots can assess the integrity of deep geothermal wells and detect issues like scaling or corrosion, automated brine handling systems can optimize the flow of fluid through the system, and remote maintenance systems can service high-temperature components like turbines, separators, and valves with improved safety and efficiency~\cite{freireAdvancedEISBasedSensor2024,zhangRealtimeTemperatureMonitoring2025}. As demand for deep geothermal grows, semi-autonomous drilling control systems may accelerate deployment in new regions and reduce non-productive time~\cite{mihaiDemonstrationAutonomousDrilling2022,sugiuraOilGasDrilling2021,liuReviewClassificationStructural2019}.

Medium-temperature systems are better suited for widespread deployment, since they do not need to be sited in regions with tectonic activity. These systems can support both electrification and district heating networks. The primary challenges are site characterization, cost-effective drilling, and long-term reservoir management~\cite{osti_1983898,iea2024geothermal}. Robotics and autonomy can contribute by automating geophysical surveys (e.g., seismic or resistivity mapping robots, or active sampling or optimization algorithms to search for promising sites), robotic well completion systems, and sensor networks for monitoring thermal drawdown in subsurface reservoirs~\cite{wangDeepLearningBased2023,yuDeepLearningGeophysics2021}.

Low-temperature geothermal systems, especially ground-source heat pumps (GSHPs), are widely applicable and highly scalable. However, they face deployment challenges due to labor-intensive drilling, site-specific customization, and limited visibility into long-term performance~\cite{Varela2024GEN,Liu2023GridCostGHP}. Robotic solutions can support automated shallow drilling and pipe installation, grouting systems, and borehole placement guided by ground penetrating radar. Once operational, thermal sensor networks and autonomous control systems can optimize heat pump efficiency and provide fault detection~\cite{manservigiDiagnosticApproachFault2022,vandrevenIntelligentApproachesFault2023}. In campus-scale or neighborhood-scale deployments, these systems can be integrated into smart energy management systems to dynamically manage heating and cooling demand based on weather, occupancy, and grid signals~\cite{sandouPredictiveControlComplex2005,klemmModelingOptimizationMultienergy2021}.

\titledsubsubsection{Thermal Energy Networks Summary}
Decarbonizing thermal energy systems requires a portfolio of solutions to meet the diverse demands of heating, cooling, and industrial processes. Across electrification, fuel infrastructure, and geothermal systems, common challenges include high installation costs, site-specific design constraints, limited visibility into long-term system performance, and the need for seamless integration with building infrastructure and broader energy networks. Robotics and autonomy research can  accelerate deployment, reduce operational risk, and unlock decarbonization at scale by supporting subsurface exploration and installation, automated retrofitting, and data-driven control. Table~\ref{energy:tab:stakeholder_research_map} summarizes these research opportunities, which we explore in greater depth in Sections~\ref{energy:autonomy:controls}--\ref{energy:autonomy:fieldrobotics}.

\begingroup

\begin{table}[h]
    \centering
    \small
    \caption{Mapping of stakeholders to research areas}
    \label{energy:tab:stakeholder_research_map}
    \setlength{\tabcolsep}{4pt}
    \renewcommand\arraystretch{1.0}
    \begin{tabular}{|l|cccc|cccc|cc|cc|ccc|ccc|}
        \hline
                                                                                                   & \multicolumn{12}{c|}{\textbf{Electric Power}} & \multicolumn{6}{c|}{\textbf{Thermal Energy}}                                                                                                                                                                                                                                                                                                                                                                                                                                                                                                         \\
        \cline{2-19}
                                                                                                   & \multicolumn{4}{c|}{\textbf{Generation}}      & \multicolumn{4}{c|}{\textbf{Markets}}        & \multicolumn{2}{c|}{\textbf{Grid}} & \multicolumn{2}{c|}{\textbf{Users}} & \multicolumn{3}{c|}{\textbf{Fuels}} & \multicolumn{3}{c|}{\textbf{Geothermal}}                                                                                                                                                                                                                                                                                                                                             \\
        \cline{2-19}
                                                                                                   & \rotatebox{90}{Developers}                    & \rotatebox{90}{EPC}                          & \rotatebox{90}{O\&M}               & \rotatebox{90}{OEMs}                & \rotatebox{90}{Exchanges}           & \rotatebox{90}{IPPs}                     & \rotatebox{90}{Retailers} & \rotatebox{90}{Aggregators} & \rotatebox{90}{TSO} & \rotatebox{90}{DSO} & \rotatebox{90}{Residential} & \rotatebox{90}{C\&I} & \rotatebox{90}{Producers} & \rotatebox{90}{Pipelines} & \rotatebox{90}{Utilities} & \rotatebox{90}{Developers} & \rotatebox{90}{Drillers} & \rotatebox{90}{District heating\ } \\ \hline

        \multicolumn{19}{|l|}{\textbf{Control \& Planning}}                                                                                                                                                                                                                                                                                                                                                                                                                                                                                                                                                                                                                                               \\ \hline
        \hyperref[energy:autonomy:control:planning_design]{Site selection}                         & $\bullet$                                     &                                              &                                    &                                     &                                     &                                          &                           &                             &                     &                     &                             &                      &                           &                           &                           & $\bullet$                  & $\bullet$                &                                    \\
        \hyperref[energy:autonomy:control:planning_design]{Portfolio optimization}                 & $\bullet$                                     &                                              &                                    &                                     &                                     & $\bullet$                                & $\bullet$                 & $\bullet$                   &                     &                     &                             &                      &                           &                           &                           &                            &                          &                                    \\
        \hyperref[energy:autonomy:control:planning_design]{Design automation}                      &                                               & $\bullet$                                    &                                    &                                     &                                     &                                          &                           &                             &                     &                     &                             &                      &                           &                           &                           &                            &                          &                                    \\
        \hyperref[energy:autonomy:control:safety]{Safe DER \& inverter control}                    &                                               &                                              &                                    & $\bullet$                           &                                     &                                          &                           & $\bullet$                   &                     & $\bullet$           &                             &                      &                           &                           &                           &                            &                          &                                    \\
        \hyperref[energy:autonomy:control:risk_security]{Risk management}                          &                                               &                                              &                                    &                                     & $\bullet$                           & $\bullet$                                & $\bullet$                 & $\bullet$                   &                     &                     &                             &                      &                           &                           &                           &                            &                          &                                    \\
        \hyperref[energy:autonomy:control:risk_security]{Security analysis}                        &                                               &                                              &                                    &                                     &                                     &                                          &                           &                             & $\bullet$           &                     &                             &                      &                           &                           &                           &                            &                          &                                    \\
        \hyperref[energy:autonomy:control:grid_edge]{Network optimization}                         &                                               &                                              &                                    &                                     &                                     &                                          &                           &                             & $\bullet$           & $\bullet$           &                             &                      &                           &                           &                           &                            &                          & $\bullet$                          \\
        \hyperref[energy:autonomy:control:risk_security]{Storm \& wildfire preparedness}           &                                               &                                              &                                    &                                     &                                     &                                          &                           &                             &                     & $\bullet$           &                             &                      &                           &                           &                           &                            &                          &                                    \\
        \hyperref[energy:autonomy:control:grid_edge]{Home energy management}                       &                                               &                                              &                                    &                                     &                                     &                                          &                           &                             &                     &                     & $\bullet$                   &                      &                           &                           &                           &                            &                          &                                    \\
        \hyperref[energy:autonomy:control:grid_edge]{EV fleet management}                          &                                               &                                              &                                    &                                     &                                     &                                          &                           &                             &                     &                     &                             & $\bullet$            &                           &                           &                           &                            &                          &                                    \\
        \hyperref[energy:autonomy:control:grid_edge]{HVAC control}$^*$                             &                                               &                                              &                                    &                                     &                                     &                                          &                           &                             &                     &                     & $\bullet$                   & $\bullet$            &                           &                           &                           &                            &                          &                                    \\
        \hline
        \multicolumn{19}{|l|}{\textbf{Estimation}}                                                                                                                                                                                                                                                                                                                                                                                                                                                                                                                                                                                                                                                        \\ \hline
        \hyperref[energy:autonomy:estimation:batteries]{Battery management}                        &                                               &                                              &                                    & $\bullet$                           &                                     &                                          &                           &                             &                     &                     &                             &                      &                           &                           &                           &                            &                          &                                    \\
        \hyperref[energy:autonomy:estimation:network_state]{Network state estimation}              &                                               &                                              &                                    &                                     &                                     &                                          &                           &                             & $\bullet$           & $\bullet$           &                             &                      &                           &                           &                           &                            &                          &                                    \\
        \hyperref[energy:autonomy:estimation:fdir]{Fault detection}                                &                                               &                                              &                                    &                                     &                                     &                                          &                           &                             & $\bullet$           & $\bullet$           &                             &                      &                           &                           & $\bullet$                 &                            &                          & $\bullet$                          \\
        \hyperref[energy:autonomy:estimation:customer_der]{DER forecasting}                        &                                               &                                              &                                    &                                     &                                     &                                          &                           & $\bullet$                   &                     &                     &                             &                      &                           &                           &                           &                            &                          &                                    \\
        \hyperref[energy:autonomy:estimation:customer_der]{Load disaggregation}                    &                                               &                                              &                                    &                                     &                                     &                                          &                           &                             &                     & $\bullet$           &                             &                      &                           &                           &                           &                            &                          &                                    \\
        \hyperref[energy:autonomy:estimation:customer_der]{Customer segmentation}                  &                                               &                                              &                                    &                                     &                                     &                                          & $\bullet$                 &                             &                     &                     &                             &                      &                           &                           &                           &                            &                          &                                    \\
        Energy efficiency audits$^*$                                                               &                                               &                                              &                                    &                                     &                                     &                                          &                           &                             &                     &                     & $\bullet$                   &                      &                           &                           &                           &                            &                          &                                    \\
        \hline
        \multicolumn{19}{|l|}{\textbf{Perception}}                                                                                                                                                                                                                                                                                                                                                                                                                                                                                                                                                                                                                                                        \\ \hline
        \hyperref[energy:autonomy:perception:inspection]{Environmental surveys}                    & $\bullet$                                     &                                              &                                    &                                     &                                     &                                          &                           &                             &                     &                     &                             &                      &                           &                           &                           &                            &                          &                                    \\
        \hyperref[energy:autonomy:perception:inspection]{Inspection}                               &                                               & $\bullet$                                    & $\bullet$                          &                                     &                                     &                                          &                           &                             & $\bullet$           & $\bullet$           &                             &                      & $\bullet$                 & $\bullet$                 & $\bullet$                 &                            &                          & $\bullet$                          \\
        \hyperref[energy:autonomy:perception:inspection]{Vegetation management}                    &                                               &                                              &                                    &                                     &                                     &                                          &                           &                             & $\bullet$           & $\bullet$           &                             &                      &                           &                           &                           &                            &                          &                                    \\
        \hyperref[energy:autonomy:perception:underground]{Underground mapping}                     &                                               &                                              &                                    &                                     &                                     &                                          &                           &                             &                     & $\bullet$           &                             &                      & $\bullet$                 & $\bullet$                 & $\bullet$                 &                            & $\bullet$                & $\bullet$                          \\
        \hline
        \multicolumn{19}{|l|}{\textbf{Field Robotics \& Manipulation}}                                                                                                                                                                                                                                                                                                                                                                                                                                                                                                                                                                                                                                    \\ \hline
        \hyperref[energy:autonomy:fieldrobotics:inspection_maintenance]{Construction \& assembly}  &                                               & $\bullet$                                    &                                    &                                     &                                     &                                          &                           &                             & $\bullet$           & $\bullet$           &                             &                      &                           &                           &                           &                            &                          &                                    \\
        \hyperref[energy:autonomy:fieldrobotics:inspection_maintenance]{Maintenance \& inspection} &                                               &                                              & $\bullet$                          &                                     &                                     &                                          &                           &                             & $\bullet$           & $\bullet$           &                             &                      & $\bullet$                 & $\bullet$                 & $\bullet$                 &                            &                          & $\bullet$                          \\
        \hyperref[energy:autonomy:fieldrobotics:inspection_maintenance]{Underground construction}  &                                               &                                              &                                    &                                     &                                     &                                          &                           &                             &                     & $\bullet$           &                             &                      &                           & $\bullet$                 & $\bullet$                 &                            & $\bullet$                & $\bullet$                          \\
        \hline
        \multicolumn{19}{r}{\footnotesize{$^*$also covered in Section~\ref{sec:built-environment} on the built environment.}}
    \end{tabular}
\end{table}

\endgroup

\newpage

\titledsubsection{Controls \& planning}\label{energy:autonomy:controls}

\titledsubsubsection{Network \& grid-edge control}\label{energy:autonomy:control:grid_edge}

As the electric grid transitions from fossil-fueled power plants to renewable energy, there is a corresponding shift from large, centralized power plants to modular distributed energy resources (DERs). This shift creates a natural need for improved control strategies.
Better controllers can support not only climate change mitigation, by enabling grid operators to manage thousands of distributed devices, but also adaptation, by making the grid more resilient to extreme events.

There is a long history of collaboration between the controls and power systems communities~\cite{annaswamyControlSocietalScaleChallenges2024}, with applications to both grid operations~\cite{molzahnSurveyDistributedOptimization2017,chenReinforcementLearningSelective2022,dontiMachineLearningSustainable2021,espinaDistributedControlStrategies2020} and the optimal control of individual generators~\cite{pereraMachineLearningTechniques2014,cetinayOptimalSitingSizing2017,fernandez-blancoOptimalEnergyStorage2017,wongReviewOptimalPlacement2019,corsoSequentiallyOptimizedData2024,caoDeepReinforcementLearningBased2020,cutlerREoptPlatformEnergy2017,haoOptimalCoordinationBuilding2018,shapiroTurbulenceControlWind2022}. Many control paradigms common in the robotics community are already commonplace in power systems. For example, hierarchical control schemes with droop control have been used for decades to separate concerns for frequency regulation and power balance between different regions and timescales, model predictive control (MPC) has seen widespread adoption for evaluation and operation of energy storage systems~\cite{cutlerREoptPlatformEnergy2017,haoOptimalCoordinationBuilding2018}, and there has been intense research interest in reinforcement learning for power systems in recent years (albeit with fewer field deployments to date~\cite{chenReinforcementLearningSelective2022,caoDeepReinforcementLearningBased2020,duDeepReinforcementLearning2022,yangReinforcementLearningSustainable2020}). Rather than reviewing the entire history of this intersection, this section briefly highlights three promising application areas for controls research, with reference to relevant prior surveys, before diving into specific opportunities for high-impact research in the following sections.

The first use case is for grid operators aiming to get more performance out of the existing grid using network control and optimization. Building new energy infrastructure is costly and time consuming, but advanced controls and optimization can help grid operators optimize existing networks. At the transmission level, there is a need for optimal switching and power flow control to minimize congestion~\cite{pillayCongestionManagementPower2015}. In the low-voltage distribution system, which has historically been a more passive system, grid operators must now actively manage thousands of distributed energy resources like solar, storage, and EVs connected to the low-voltage grid~\cite{forresterStateRegulatoryOpportunities2025}. As a result, there is a need for scalable algorithms for distributed optimization and control for energy networks~\cite{molzahnSurveyDistributedOptimization2017,chenReinforcementLearningSelective2022,zidanFaultDetectionIsolation2017}.

While the underlying physics of electrical and thermal power networks are different, similar network optimization and control problems also arise in district heating and low-temperature geothermal networks~\cite{kuntuarovaDesignSimulationDistrict2024,buffaAdvancedControlFault2021,caiAgentbasedDistributedDemand2020,sandouPredictiveControlComplex2005}. Rather than balancing voltage and the flow of electrical power, thermal networks must balance injections and withdrawals of heat at different nodes and can improve efficiency by matching customers with different heating needs; for example, by simultaneously heating households and cooling a data center or food service facility~\cite{wahlroosUtilizingDataCenter2017}. Unlike electricity networks, thermal networks are inherently able to store energy (e.g. in hot water). This creates opportunities for advanced controllers to coordinate between electrical and thermal networks~\cite{vandermeulenControllingDistrictHeating2018}.

The second use case involves controlling individual energy assets to safely connect new resources to the grid and maximize performance. The large spinning turbines in traditional generators provide inertia that allows those power plants to use relatively simple controllers to regulate voltage and frequency~\cite{milanoFoundationsChallengesLowinertia2018,chowPowerSystemModeling2020}. In contrast, most solar panels, wind turbines, and batteries connected to the grid utilize ``grid-following'' inverters which convert direct current (DC) to alternating current (AC) that matches the voltage and frequency of the rest of the grid~\cite{rathnayakeGridFormingInverter2021}. If not precisely tuned, the control loops governing these inverters can couple with the internal dynamics of the grid, causing oscillations and instability for the TSO~\cite{milanoFoundationsChallengesLowinertia2018}. As a result, there is an increasing need for equipment manufacturers to ensure that their inverters can safely integrate with a range of grid conditions (including potential adverse interactions with other nearby inverters), provide ``grid-forming'' capabilities to regulate voltage and frequency at the transmission level, and support emerging use cases at the distribution level like microgrids capable of disconnecting from the main grid during emergencies~\cite{simpson-porcoSecondaryFrequencyVoltage2015}.
We discuss opportunities to enable ``plug-and-play'' safety guarantees further in Section~\ref{energy:autonomy:control:safety}. In addition to the challenges of safely connecting new resources to the grid, there is also active research in optimally controlling individual assets, particularly batteries and EVs, to maximize financial performance~\cite{yangModellingOptimalEnergy2022,huElectricVehicleFleet2016}.

The third use case targets the increasing numbers of so-called ``grid-edge'' devices like electric vehicles, smart thermostats, and flexible commercial loads like refrigeration and data centers. These devices can provide valuable flexibility for grid operators, since utilities and aggregators can use these devices to reduce demand during peak hours. However, realizing this flexibility requires additional care: while a battery can be charged and discharged at will, smart thermostats need to balance grid signals with occupant comfort and the unique thermal properties of each building. Similarly, while EVs can provide flexibility by shifting their charging patterns, they must make sure that drivers have a sufficiently charged vehicle each morning. Solving these control problems will require tighter integration of control and modeling to represent customer preferences and the diverse dynamics of grid-edge devices, as discussed further in Section~\ref{energy:autonomy:estimation:customer_der}

\titledsubsubsection{Safe control \& certification}\label{energy:autonomy:control:safety}

A persistent challenge in power systems is the difficulty of deploying advanced control strategies in real grids. While power producers have an incentive to adopt new controllers if they provide better economic performance~\cite{subramanyaExploitingBatteryStorages2022}, grid operators have a mandate to ensure reliability, leading them to take a conservative, worst-case approach to integrating new DERs into the power system~\cite{dahyeonyuFlexibleEVConnections}. As OEMs develop new control systems to improve DER performance, it may not be feasible for grid operators to individually verify each new controller, and OEMs may be unwilling to share the details of proprietary control algorithms~\cite{moringNodalOperatingEnvelopes2024}. As a result, there is a need to efficiently verify that new controllers will be able to ``plug and play'' safely with a variety of grid conditions; for example, to check that inverter controllers meet requirements for riding through disturbances or supporting the frequency of the grid~\cite{hopeReachabilityBasedMethodLargeSignal2011,villegaspicoVoltageRideThroughCapability2014}. The need for efficient safety verification is most acute at the distribution level, since individually analyzing thousands of small, diverse DERs is an emerging challenge for DSOs. However, TSOs may also be able to leverage work in this area to streamline interconnection analysis for large resources~\cite{davidrosnerCommissionerRosnersLetters}.

Research in safe control can help address this challenge. Safety filters like control barrier functions (CBFs,~\cite{amesControlBarrierFunction2017}) and Hamilton-Jacobi reachability~\cite{bansalHamiltonJacobiOverview2017} can filter potentially-unsafe controllers to obtain control signals that are guaranteed to preserve safety. These techniques have a long history in robotics research, with applications to drones~\cite{xuSafeTeleoperationDynamic2018,zhangGCBFNeuralGraph2024} and self-driving cars~\cite{amesControlBarrierFunction2017,singletaryOnboardSafetyGuarantees2022,xiaoBarrierNetDifferentiableControl2023}. In the most extreme examples, these safety filters have been demonstrated to prevent drones and autonomous vehicles from crashing despite unsafe actions by human operators~\cite{xuSafeTeleoperationDynamic2018,singletaryOnboardSafetyGuarantees2022,alanControlBarrierFunctions2023,wabersichDataDrivenSafetyFilters2023}. These techniques use knowledge of system dynamics and constraints to derive state-dependent constraints on an admissible set of safe controls, with accompanying proofs on safety and stability.

Similar methods have been adapted for power systems, where they are often referred to as dynamic injection limits or operational envelopes for DERs~\cite{rossStrategiesNetworkSafeLoad2021,rossManagingVoltageExcursions2019,jangProbabilisticConstraintConstruction2024,nazirConvexInnerApproximation2019,nazirGridawareAggregationRealtime2021,nguyenConstructingConvexInner2019,petrouOperatingEnvelopesProsumers2020,yiFairOperatingEnvelopes2022,russellStochasticShapingAggregator2022}. Nevertheless, a number of important open questions remain. A key challenge is the interaction between the DSO, the DER aggregator, and the OEM that provides the controller. While the DSO wishes to impose safeguards, the aggregator and OEM may be unwilling to share proprietary information about its control strategy with the DSO. Some existing works assume that the DSO is able to simulate the aggregator's behavior~\cite{rossManagingVoltageExcursions2019,rossStrategiesNetworkSafeLoad2021}, but this framework may not satisfy the aggregator and OEM's need for privacy. Other works do not require the DSO to model the precise control strategy but still require the DSO to model the dynamics of customer-sited DERs like smart thermostats and HVAC systems~\cite{jangProbabilisticConstraintConstruction2024}, but grid operators may be unwilling to assume this extra modeling responsibility. Other works use optimization to derive constraints on aggregator behavior that do not depend on the details of the aggregated DER. \cite{nazirConvexInnerApproximation2019,nazirGridawareAggregationRealtime2021,nguyenConstructingConvexInner2019} compute inner approximations of the set of safe aggregator behaviors. DSOs would then rely on aggregators to internally manage the dynamics of their aggregated DERs to stay within these safe sets rather than modeling the aggregation themselves. \cite{yiFairOperatingEnvelopes2022} adds additional fairness constraints in the case when safe sets are constructed for multiple aggregators simultaneously.

Open problems remain on designing safety filters that minimize (or establish lower bounds on) information exchange between the DSO and aggregator. In addition, there is a lack of research on how online filters can integrate with existing power systems and markets. Many power system operators, including in the United States, United Kingdom, and Australia, are actively considering how these safeguarding methods can be used to safely interconnect a growing fleet of renewable generators and DERs~\cite{dahyeonyuFlexibleEVConnections}, but there is a gap in understanding how safety filters may limit the ability of resources to provide various grid services (e.g., whether safety filters may affect a DER's ability to provide frequency support). In the context of these policy discussions, there is a need for both foundational research into new safety filter methods and applied research on the use of these algorithms in real-world power markets.

\titledsubsubsection{Risk management \& security analysis}\label{energy:autonomy:control:risk_security}

While energy flows through the grid in every hour of the year, the handful of peak hours in each year are the most important for planning the power system and for the finances of power market participants (many of whom are compensated primarily for their performance during these hours). As a result, the ability to efficiently simulate the power system during extreme conditions and verify its performance is a foundational capability for many power systems stakeholders, particularly TSOs (who are responsible for ensuring that enough generation and transmission is available to meet demand during extreme conditions). Traditionally, grid operators and producers have focused on extreme events driven by high demand for electricity, which historically are hot summer days with high demand for air conditioning, but there is an emerging need to model other types of extreme events, like severe storms and wildfires. As climate change increases the frequency and severity of both storms and fire-prone weather, these use cases will become increasingly important for climate adaptation.
Moreover, many TSOs are actively considering reforms to how they model and analyze extreme events~\cite{schlag2020capacity,iso_ne_2023_extreme_weather}, so timely research in this area can support active policy and market design debates in the power sector.

This need aligns with ongoing research on rare event simulation and safety verification in both the autonomy and power systems communities. An important challenge in simulating extreme grid conditions is the need to consider uncertainty from black-box sources like weather forecasting models, which presents a challenge for safety assessment methods based on formal methods~\cite{zhangReviewSettheoreticMethods2020,jinReachabilityAnalysisBased2010,althoffFormalCompositionalAnalysis2014,el-guindyCompositionalTransientStability2017,hopeReachabilityBasedMethodLargeSignal2011,villegaspicoVoltageRideThroughCapability2014,bouvierDistributedTransientSafety2022} and motivates the use of simulation-driven safety verification. Earlier works applied standard Monte Carlo methods to probabilistic power flow studies~\cite{wuStochasticSecurityConstrainedUnit2007,hajianProbabilisticPowerFlow2013}, analyzing the availability of energy supplies~\cite{shahzadProbabilisticSecurityAssessment2021,hegazyAdequacyAssessmentDistributed2003,mccalleyProbabilisticSecurityAssessment2004,ortega-vazquezEstimatingSpinningReserve2009,huangModelingDynamicDemand2015}, and predicting the impact of severe weather events~\cite{panteliPowerSystemResilience2017,younesiAssessingResilienceMulti2020,bessaniProbabilisticAssessmentPower2019}. However, a major challenge for Monte Carlo methods is scalability, as the number of samples required to cover the space of possible outcomes increases exponentially with the dimension of the search space. This challenge, which arises in both robotics~\cite{dawsonBayesianApproachBreaking2023,corsoSurveyAlgorithmsBlackBox2021} and power systems~\cite{dawsonAdversarialOptimizationLeads2023} prompts the use of optimization-based algorithms~\cite{molzahnGridAwareGridAgnosticDistribution2019,dontiAdversariallyRobustLearning2021} and accelerated rare-event simulation algorithms like importance sampling~\cite{guoEfficientCascadingOutage2018}, cross-entropy methods~\cite{leitedasilvaRiskAssessmentProbabilistic2019,gonzalez-fernandezCompositeSystemsReliability2013,gonzalez-fernandezReliabilityAssessmentTimeDependent2011}, Markov chain sequential Monte Carlo~\cite{dawsonAdversarialOptimizationLeads2023,wangMarkovChainMonte2018,huaExtractingRareFailure2015,goodridgeRareeventStudyFrequency2021,goodridgeUncoveringLoadAlteringAttacks2024}, and Gaussian process learning~\cite{tanScalableRiskAssessment2024}. An exciting area for future work is the combination of rare-event sampling with learned prior distributions like diffusion models~\cite{hoDenoisingDiffusionProbabilistic2020} and normalizing flows~\cite{rezendeVariationalInferenceNormalizing2015}, inspired by recent work on self-driving cars~\cite{zhongGuidedConditionalDiffusion2023}. These learned prior distributions could enable more accurate risk assessment of extreme weather and behavior of consumer-controlled distributed energy resources. In addition to simulating peak load conditions, another area for future work is extending these methods to manage risk from wildfires~\cite{muhsWildfireRiskMitigation2020,vazquezWildfireMitigationPlans2022,sohrabiWildfireProgressionSimulation2024,nematshahiRiskAssessmentTransmission2025} and severe storms~\cite{younesiTrendsModernPower2022,guikemaPredictingHurricanePower2014,bhusalModelingNaturalDisasters2020,younesiAssessingResilienceMulti2020a}, which will require much higher spatial and temporal granularity and integration of geographic, vegetation, and real-time weather data.

\titledsubsubsection{Optimal planning \& design automation}\label{energy:autonomy:control:planning_design}

To design new energy infrastructure, designers must choose an optimal mix of different technologies to balance competing objectives. For example, policymakers and grid operators define decarbonization pathways by modeling what generation mix (e.g., solar, wind, battery storage, etc.) is needed to achieve greenhouse gas reduction targets at minimum cost~\cite{larson2021netzero}. Commercial and industrial customers designing microgrids for backup power need to optimize the combination of solar, storage, and backup diesel or gas generation to last through expected outages~\cite{rigo-marianiIntegratedOptimalDesign2017,hassanOptimalDesignMicrogrids2011,arefifarOptimumMicrogridDesign2013,zhangOptimalDesignCHPbased2015}.

Similar design challenges arise in the optimal siting of renewable energy resources~\cite{qiuDecarbonizedEnergySystem2024} and planning so-called ``multi-energy'' systems (e.g., simultaneously optimizing electrical and natural gas~\cite{khorramfarCosteffectivePlanningDecarbonized2025} or hydrogen networks~\cite{klemmModelingOptimizationMultienergy2021}. An emerging need for advanced planning tools is for developing high- and medium-temperature geothermal systems where performance depends on difficult-to-predict subsurface features. Several works have explored optimization and active sampling to this problem~\cite{corsoSequentiallyOptimizedData2024,chenEfficientOptimizationWell2015}.

Optimization-based planning tools in power systems are relatively mature~\cite{mitenergyinitiativeGenXConfigurablePower}, but handling uncertainty (e.g., future weather patterns for renewables, or subsurface potential for geothermal, etc.) is a long-standing open problem where interdisciplinary research could help provide scalable, rigorous approaches to power systems optimization under uncertainty~\cite{aienComprehensiveReviewUncertainty2016,gorenstinPowerSystemExpansion1993,merrillRiskUncertaintyPower1991,verasteguiAdaptiveRobustOptimization2019}.

\vspace{1em}
\begin{boxMarginLeft}
      \subsubsection*{Future Directions: Control \& Planning in Energy}
      \begin{itemize}
            \item \emph{Foundations in power systems control.}
                  There is a long history of control theory research in power
                  systems. Successful collaboration between the autonomy and power
                  systems community should recognize and build off of this history.

            \item \emph{High-impact applications for control systems.}
                  Network optimization (in both electric transmission, distribution
                  networks, and certain thermal networks), grid-forming inverters,
                  and grid-edge resources like HVAC and industrial loads are
                  impactful applications for advanced controllers.

            \item \emph{Safety guarantees.}
                  Formal methods like reachability and
                  barrier functions can provide safety guarantees to help integrate clean energy resources into
                  the grid, but research is needed to understand the interaction of
                  these methods with grid dynamics and power markets.

            \item \emph{Rare-event simulation and uncertainty quantification.}
                  Research on rare event simulation, particularly for improving
                  scalability or integrating with complex weather prediction models,
                  can help support climate adaptation to storms and wildfire risk in
                  the power sector.

            \item \emph{Optimal design under uncertainty.}
                  Optimization-based planning tools are already used to help solve
                  optimal design problems in power systems, but interdisciplinary
                  research could support more usable, scalable tools for handling
                  uncertainty in these design problems.
      \end{itemize}
\end{boxMarginLeft}

\titledsubsection{Estimation}\label{energy:autonomy:estimation}

\titledsubsubsection{Grid state estimation}\label{energy:autonomy:estimation:network_state}

Visibility into real-time grid state is a persistent challenge for grid operators, although this problem manifests differently for the transmission and distribution systems. Transmission system operators focus on the flow of power through the network, which requires estimating the frequency and relative phase of alternating current at different locations using dedicated sensors called phasor measurement units~\cite{blumeElectricPowerSystem2016}. Distribution system operators typically care more about maintaining acceptable voltage throughout the distribution system, but the larger number of nodes in the distribution system means that direct sensing is not feasible. Instead, DSOs rely on distribution system state estimation (DSSE) algorithms to infer the voltages at every node using sparse measurements from sensors at key locations~\cite{primadiantoReviewDistributionSystem2017,dehghanpourSurveyStateEstimation2019,wuRobustStateEstimator2013,bloodElectricPowerSystem2008}. While this section focuses on estimation for distribution networks, we do not discount need for estimation tools at the transmission level, where TSOs must be able to detect large-scale oscillations and isolate the source of any instability (often a misconfigured generator or unintentional coupling between equipment at different nodes~\cite{caiWideAreaInterArea2013,liuOscillationMonitoringSystem2007,maslennikovOnlineOscillationsManagement2019}).

Classical DSSE algorithms are designed for static systems with known power flow patterns and easily quantified sources of uncertainty, and they struggle to adapt to the correlated, non-Gaussian behavior of renewables like PV (unlike households, all solar panels in a neighborhood turn on and off at the same time, so the random behavior of different DERs can be highly correlated~\cite{dehghanpourSurveyStateEstimation2019, muscasEffectsMeasurementsPseudomeasurements2014, sanjariProbabilisticForecastPV2017}).
In response, various works have explored using improved uncertainty representations to capture the time-varying, correlated behavior of renewable resources~\cite{zhaoRobustForecastingAided2018, bililMMSEBasedAnalyticalEstimator2018, wengProbabilisticJointState2019, zhangForecastingAidedJointTopology2024,alhalaliStateEstimatorElectrical2015}
%
The rise of connected devices such as smart thermostats, water heaters, and industrial equipment has also increased uncertainty for DSSE. Classical DSSE algorithms rely on the passive, predictable nature of loads to augment sparse sensor measurements with ``pseudo-measurements'' based on historical load profiles~\cite{dehghanpourSurveyStateEstimation2019}. However, active consumer behavior creates a new feedback loop between electricity consumption and system state that makes historical load profiles and passive estimation methods difficult to apply~\cite{liuDistributionSystemState2021}.
Some works respond to increasingly active loads by adding sensors~\cite{sarriStateEstimationActive2012}, but this approach carries substantial cost. Rather than add new sensors, other works have attempted to model the behavior of responsive loads as part of the state estimation process~\cite{liuDistributionSystemState2021} and apply more robust inference algorithms like Markov Chain Monte Carlo~\cite{pegoraroBayesianApproachDistribution2017,massignanBayesianInferenceApproach2022,sanjariProbabilisticForecastPV2017} and particle filters~\cite{zhangForecastingAidedJointTopology2024,alhalaliStateEstimatorElectrical2015,nanchianThreephaseStateEstimation2016}.

Two exciting areas of future work stand out. First, recent work in the robotics community has applied deep generative models as flexible uncertainty representations in a range of domains~\cite{zhongGuidedConditionalDiffusion2023,zhangTrajectoryPredictionAutonomous2022,ivanovicMultimodalDeepGenerative2021,xueSSLSTMHierarchicalLSTM2018,shiSGCNSparseGraph2021,zhaoMultiAgentTensorFusion2019,liGRINGenerativeRelation2021,liEvolveGraphMultiAgentTrajectory2020}, which can be transferred to uncertainty modeling for renewables and energy consumer behavior in DSSE. Furthermore, recent advances in robust inference, for example using graphical models~\cite{dellaertFactorGraphsExploiting2021}, approximate Bayesian methods~\cite{zhangForecastingAidedJointTopology2024,alhalaliStateEstimatorElectrical2015,nanchianThreephaseStateEstimation2016,rezendeVariationalInferenceNormalizing2015,wongVariationalInferenceParameter2020}, and physics-informed machine learning~\cite{ostrometzkyPhysicsInformedDeepNeural2020,caoPhysicsInformedGraphicalLearning2024,zamzamPhysicsAwareNeuralNetworks2020}
may help DSSE systems adapt to increased uncertainty and the coupling introduced by  active consumers.

\titledsubsubsection{Fault detection, isolation, \& restoration}\label{energy:autonomy:estimation:fdir}

A special case of network state estimation is fault detection, isolation, and recovery (FDIR), which aims to identify faults (e.g., short circuits from a tree falling on a wire) and reconfigure the network to route power around the fault~\cite{zidanFaultDetectionIsolation2017}. FDIR is especially important in distribution networks, which are typically operated radially such that only one path of power is active for each customer, so any fault will cause all downstream customers to lose power (in contrast, transmission networks are more often operated as meshed networks with multiple paths for power to flow). Existing methods estimate the location of the fault either by analyzing changes in impedance in distribution circuits~\cite{girgisFaultLocationTechnique1993,mora-florezComparisonImpedanceBased2008,jiaHighFrequencyImpedance2018}, by analyzing time series measurements from the fault~\cite{iurinicDistributionSystemsHighImpedance2016}, or through data-driven methods~\cite{jiangFaultDetectionIdentification2014,rizeakosDeepLearningbasedApplication2023,hossanDataDrivenFaultLocation2019}.

The scale of the distribution network poses a challenge for estimation algorithms in FDIR. Distribution utilities often rely on private communication networks with limited bandwidth, so transmitting high-resolution measurements from thousands of field devices back to a centralized estimation system is not feasible. Prior research has explored decentralized approaches that decompose the FDIR problem for different sections of the distribution grid~\cite{zidanFaultDetectionIsolation2017}. As utilities deploy smart meters there is an opportunity to push computation for fault detection out to these edge devices (along with other use cases, like DER management, power quality measurement, etc.), making use of high-resolution local measurements without using limited bandwidth. As discussed in Section~\ref{energy:autonomy:estimation:network_state}, another challenge for network state estimation algorithms is the increasing variability from intermittent renewable resources, which will require increasingly robust estimation methods for fault detection~\cite{jiaHighFrequencyImpedance2018,zidanFaultDetectionIsolation2017}. In addition, there are opportunities to more tightly integrate decentralized and distributed estimation with control algorithms to optimally reconfigure the distribution system after locating the fault~\cite{nguyenAgentBasedRestoration2012,konarDistributedOptimizationAutonomous2023,arifOptimizingServiceRestoration2018}. Finally, while the physics of electrical and thermal networks (like district heating and geothermal systems) are quite different, there are opportunities to develop physics-based fault detection algorithms that can transfer across domains, especially as electrical and thermal networks become more tightly integrated~\cite{buffaAdvancedControlFault2021,vandrevenIntelligentApproachesFault2023,manservigiDiagnosticApproachFault2022}.

\titledsubsubsection{Customer behavior \& distributed energy resource management}\label{energy:autonomy:estimation:customer_der}

Unlike conventional power plants, which are required to report their status to grid operators in real time, customers have historically been a black box. However, utilities are increasingly interested in understanding customer behavior in order to better plan and operate the distribution system. For example, there is a long history of power systems research on load disaggregation, which seeks to estimate which equipment a customer is running at any given time based on their electricity consumption; for example, to detect when an EV has started charging~\cite{kolter2011redd,jazizadehEMBEDDatasetEnergy2018,volkerFIREDFullylabeledHIghfRequency2020}. This information can help utilities decide when to upgrade grid infrastructure to accommodate new loads like EVs and heat pumps~\cite{antonSmartMeterInsights2024}. The ability to group customers according to load profiles and technologies is also useful for retailers, who can use this information to design products targeted at specific customer classes (although this raises privacy and consumer protection concerns).

When customers are able to shift demand to reduce peak load on the grid, utilities can use this demand response ability to reduce the need for expensive grid upgrades~\cite{jenniferdowningPathwaysCommercialLiftoff2023,antonopoulosArtificialIntelligenceMachine2020}. However, this requires the utility or aggregator to have confidence in how much capacity is available in real time from distributed energy resources like smart thermostats, EVs, and batteries. Prior works have proposed learning dynamic models of building HVAC systems and occupant behavior for demand response~\cite{zhangOptimalLearningBasedDemand2016,songStateSpaceModeling2019,pallonettoDemandResponseAlgorithms2019,tangDatadrivenModelPredictive2022,sarranDatadrivenStudyThermostat2021}. An open question in this area is how to integrate these learned models with optimal control in a way that can scale to aggregations of thousands of diverse loads.

\titledsubsubsection{Battery state estimation}\label{energy:autonomy:estimation:batteries}

Battery energy storage is a critical tool for decarbonizing the power system, providing both a buffer for intermittent renewable generation and resiliency to weather-induced outages. However, realizing this potential requires solving a number of challenging state and parameter estimation problems, as surveyed in \cite{linModelingEstimationAdvanced2019,zhengThermalStateMonitoring2024}. In particular, in order to safely and efficiently operate a battery, we must be able to estimate its state of charge (SoC), state of health (SoH, a measure of battery aging), and state of power (SoP, a measure of how much power can be charged or discharged at any moment). Each of these parameters depends on nonlinear electrochemical dynamics and cannot be reliably measured open-loop~\cite{linModelingEstimationAdvanced2019}. Prior work has explored a range of state estimation and parameter identification algorithms for these problems, including extended and unscented Kalman filters~\cite{plettExtendedKalmanFiltering2004a,plettExtendedKalmanFiltering2004,xiongDatadrivenMultiscaleExtended2014,didomenicoLithiumIonBatteryState2010,wangRevisitingStateChargeEstimation2017,zhangAdaptiveUnscentedKalman2015}, particle filters~\cite{sahaPrognosticsMethodsBattery2009,liuNovelTemperaturecompensatedModel2014,schwunkParticleFilterState2013}, sliding mode observers~\cite{kimNovelStateCharge2006}, partial differential equation (PDE)-based observers~\cite{mouraBatteryStateEstimation2015}, genetic algorithm- or particle swarm-based global optimization~\cite{huElectrothermalBatteryModel2011,huComparativeStudyEquivalent2012,huComparativeStudyControlOriented2019} and neural networks (either with or without physics-based structure or an accompanying model-based observer like a Kalman filter~\cite{mouraBatteryStateEstimation2015,andreComparativeStudyStructured2013,chenStatechargeEstimationLithiumion2019,liPhysicsinformedNeuralNetworks2021,weiMachineLearningbasedHybrid2022}). An important area for future research in this area is tighter integration of planning, control, and estimation; for example, by using battery health models to derive optimal fast-charging schedules~\cite{chungOptimizationElectricVehicle2020}.

\titledsubsubsection{Geothermal resource characterization}\label{energy:autonomy:estimation:geotherm}

Low-temperature geothermal systems use the Earth as a heat sink to make electric heating and cooling more efficient; they do not have strict locational requirements. In contrast, medium- and high-temperature geothermal systems actually extract energy from the Earth's crust. To be viable, these systems must be located in regions with suitable subsurface conditions: a combination of high temperature gradients (which provides energy to extract), high permeability (which allows water to extract that energy), and high recharge rate (allowing the resource to continue to produce energy over time). The long-term commercial success of geothermal power will depend on the ability to consistently identify and develop high-quality geothermal resources~\cite{iea2024geothermal}.

Estimation methods can help in three ways. At the portfolio level, active sampling and Bayesian inversion techniques can help efficiently search for promising places to drill~\cite{chenEfficientOptimizationWell2015,corsoSequentiallyOptimizedData2024,clemensArtificialIntelligenceCentricLowEnthalpy2024,zhangProbabilisticGeothermalResources2023,chenEfficientBayesianInversion2014}.
After drilling a well, inversion tools can also help estimate the subsurface conditions and long-term potential of each well~\cite{osullivan7ReservoirModeling2025,liuInvertingMethodsThermal2018,ciriacoGeothermalResourceReserve2020}. Finally, integrating reservoir estimation techniques with advanced controllers can help boost well production and improve resource economics~\cite{wangDeepLearningBased2023,zhangRealtimeTemperatureMonitoring2025}.


\vspace{1em}
\begin{boxMarginLeft}
      \subsubsection*{Future Directions: Estimation in Energy}
      \begin{itemize}
            \item \emph{Grid state and DER estimation.}
                  There are opportunities for estimation to help solve practical
                  problems throughout the power system. Grid operators can benefit
                  from improved network state estimation and fault detection
                  methods, and aggregators and retailers can benefit from an
                  improved understanding of customer and DER behavior.

            \item \emph{Robustness and scalability.}
                  Estimation algorithms that are robust to uncertainty from
                  intermittent renewable generation and capable of scaling to large
                  numbers of grid nodes and connected devices will have the greatest
                  opportunity for impact.

            \item \emph{Advanced battery management.}
                  Estimation research can support OEMs in developing advanced
                  battery management systems, which can improve the cost and
                  performance of energy storage systems in EVs and grid-scale
                  storage.

            \item \emph{Geothermal prospecting.}
                  Estimation and inversion methods can help improve the economics of
                  medium- and high-temperature geothermal by assisting with
                  prospecting and resource characterization.

            \item \emph{Theoretical and algorithmic foundations.}
                  In several areas, we find that the robotics and power systems
                  communities are wrestling with the same fundamental questions. How
                  much physics-based structure should be incorporated into
                  data-driven estimators? Can learned models approximate higher
                  fidelity model for real-time estimation and control? What
                  optimized control strategies might arise from the tighter
                  integration of estimation, planning, and control? Answering these
                  questions will enable grid operators to get ``more for less'' by
                  optimizing the performance of their existing networks.

      \end{itemize}
\end{boxMarginLeft}

\titledsubsection{Perception}\label{energy:autonomy:perception}


\titledsubsubsection{Underground mapping}\label{energy:autonomy:perception:underground}

Due to concerns over both aesthetics and resiliency, electric utilities are increasingly interested in moving overhead power lines underground~\cite{panossianPowerSystemWildfire2023}. Unfortunately, the cost of underground power lines can be 3--10 times higher than the cost of overhead lines~\cite{warwickElectricityDistributionSystem2016}. The high cost of underground construction also increases the cost of new thermal networks like district heating and geothermal. While much of this cost difference is due to specialized excavation equipment and labor, a substantial cost comes from the risk of encountering hidden obstacles or existing underground utility lines such as power and gas lines (this risk is particularly acute for less-disruptive ``trenchless'' construction methods). Since accidental contact with natural gas lines can lead to extreme safety risks and expensive delays, improved underground mapping and perception methods could reduce the risk and cost of underground construction.

The primary challenge in underground mapping is uncertainty: records from previous construction projects can be unreliable, and even direct sensors like ground penetrating radar (GPR) can be extremely noisy depending on the subsurface composition. By fusing information from these unreliable sources together with known waypoints like manhole covers, it may be possible to create higher-accuracy maps and reduce the risk of accidentally damaging existing utility lines. Some works have explored using GPR measurements to help localize robots in GPS-denied environments~\cite{baikovitzGroundEncodingLearned2021,ortGROUNDEDLocalizingGround2021}, but we expect that further work in simultaneous localization and mapping can help create high-quality maps of the underground environment itself, along the lines of the work in~\cite{kouros3DUndergroundMapping2018,exodigoExodigoSolvingUnderground,wangUndergroundInfrastructureDetection2023,vohraAutomatedUndergroundMapping2024,bilalInferringMostProbable2018,zhouUndergroundPipelineMapping2022}.

\titledsubsubsection{Autonomous inspection \& monitoring}\label{energy:autonomy:perception:inspection}

To maintain reliability and mitigate fire risk, grid operators need to inspect and maintain large fleets of poles and overhead power lines. In the United States alone, there are 6.3 million miles of distribution lines~\cite{warwickElectricityDistributionSystem2016}. Many of these lines run through rural areas and the wildland-urban interface, where the combination of increasingly hot and dry weather and aging power infrastructure creates a substantial wildfire risk~\cite{panossianPowerSystemWildfire2023}. Utilities currently spend large sums on manual vegetation surveys to identify which trees to trim to reduce fire risk, but the time and expense of these surveys means that industry best practice is to check each line only once every few years. Some utilities have explored using satellite imagery to conduct these surveys remotely~\cite{overstoryOverstorySatelliteVegetation}, but concerns remain around data quality and false-negative rates from remote sensing. Utility companies have begun using drones to inspect power infrastructure, but data collection is still largely manual, requiring a human pilot to drive to each inspection site and operate the drone~\cite{xcelenergyHearingExhibit1022020}. In addition, these surveys yield large amounts of data that can be difficult to process and analyze. Automating this inspection using computer vision and machine learning methods is an active area of research~\cite{nguyenAutomaticAutonomousVisionbased2018,persianiDronebasedFaultRecognition2025}, and future research may explore using active perception to enable robots to make real-time decisions on which pieces of infrastructure to re-visit and inspect in greater detail, improving coverage and avoiding the need for costly human re-inspection. In addition to asset inspection needs, transmission system operators can also benefit from real-time monitoring of transmission line sag, wind loading, and ice build-up~\cite{chenSurveySagMonitoring2023}, which can help with real-time outage prediction and dynamic line rating (where the system operator can push additional power through a line under favorable weather conditions~\cite{fernandezReviewDynamicLine2016}).

In addition to these utility inspection needs, generator owners and operators
also need to regularly inspect their power plants to detect emerging issues and
plan repairs. In this area, wind turbines are a promising use case for robotic
inspection, since turbine blades currently require expensive and dangerous
manual inspection from human operators rappelling down the side of the
blade~\cite{renOffshoreWindTurbine2021}. To date, there has been extensive work
using drones to carry out these
inspections~\cite{liuReviewRobotbasedDamage2022,wangAutomaticDetectionWind2017,renOffshoreWindTurbine2021,khalidApplicationsRoboticsFloating2022,pierceQuantitativeInspectionWind2018}.
There has been similar work on drone inspection for solar
panels~\cite{raptormapsRaptorMaps,nooralishahiDroneBasedNonDestructiveInspection2021}.
It is possible that some of these inspection technologies could also be used
prior to construction; for example, to automate time-consuming and
labor-intensive environmental surveys.

\vspace{1em}
\begin{boxMarginLeft}
      \subsubsection*{Future Directions: Perception in Energy}
      \begin{itemize}
            \item \emph{Collaboration with end users.}
                  Some applications, such as inspection for wind and solar farms,
                  are relatively mature and in the process of commercialization.
                  In these cases, there are opportunities for robotics researchers to
                  collaborate directly with industry and end users to
                  understand their needs and develop new capabilities.

            \item \emph{Application-specific datasets.}
                  Emerging applications like environmental reviews and utility
                  vegetation management require collaboration with industry to build
                  high-quality datasets.

            \item \emph{Active perception for inspection.}
                  Active perception research can enable inspection robots to be more
                  efficient, making real-time decisions about areas to inspect in
                  greater detail.

            \item \emph{Large-scale multi-modal mapping.}
                  Mapping and sensor fusion research can be used to create
                  high-quality, large-scale maps of utility infrastructure. Along
                  with methods for incorporating unconventional sensing modalities,
                  such as ground-penetrating radar, researchers in this area can
                  enable substantial improvements in the cost and speed of
                  underground construction.
      \end{itemize}
\end{boxMarginLeft}


\titledsubsection{Field robotics \& manipulation}\label{energy:autonomy:fieldrobotics}


\titledsubsubsection{Autonomous construction \& assembly}\label{energy:autonomy:fieldrobotics:construction}

Building a large-scale wind or solar farm requires the efforts of a large workforce, often spread over hundreds of acres of remote terrain.
This section emphasizes opportunities for thoughtfully-designed robotic systems to support human workers, particularly in dull, dirty, or dangerous tasks or tasks that require a high degree of precision. For example, to build a solar farm workers must drive hundreds of metal supports (piles) into the ground within millimeters of their intended position. This task is currently done using mechanical pile drivers and has seen early attempts to commercialize a fully autonomous solution to improve precision and repeatability~\cite{builtroboticsAutonomousSolarPiling}. After driving these piles, workers need to install (``rack'') the solar modules onto mounting hardware. Modern solar panels are already near the limit of what a single person can safely lift, so assistive robots can help reduce worker strain through partial or full automation~\cite{aesAESLaunchesFirst,chargeroboticsChargeRobotics,pottLargescaleAssemblySolar2010}. Wind turbines require heavy equipment to install the towers and blades, but there are similar opportunities to support human operators and improve precision through automation~\cite{hoffmanPrecisionAssemblyHeavy2020}. While there are opportunities for focused research on these problems, practical success will also require substantial commercialization effort to integrate existing technologies into robust systems that can operate reliably in challenging field conditions.

Beyond opportunities to automate existing construction processes, there are also areas where robotics research can enable entirely new capabilities, particularly in underground construction. Currently, installing distribution lines underground usually requires digging an open trench, laying the cable, then back-filling and re-paving the hole. This process is not only expensive but also creates substantial disruptions at the surface. Research aimed at developing novel systems for installing underground cables, particularly with the ability to maneuver around existing underground utilities (e.g., natural gas pipes), could have a substantial impact on utilities' ability to improve resiliency and reduce wildfire risk by moving distribution lines underground~\cite{pacificgas&electric2024RDStrategy}. Similar research could also help reduce the cost of installing low-temperature geothermal and district heating networks, supporting new investment in decarbonized thermal networks.

\titledsubsubsection{Autonomous inspection \& maintenance}\label{energy:autonomy:fieldrobotics:inspection_maintenance}

Since renewable generators like wind and solar do not consume fuel, maintenance
accounts for a large fraction of their operational costs. Unfortunately, these
generators are often located in rural or remote areas, increasing the cost of
regular maintenance. Offshore wind turbines are an extreme example of this
difficulty, requiring a multi-hour boat journey for routine maintenance. Several
works have explored robotic inspection and maintenance systems for offshore wind
turbines, as surveyed
in~\cite{renOffshoreWindTurbine2021,khalidApplicationsRoboticsFloating2022}.
These studies have applied
drones~\cite{pierceQuantitativeInspectionWind2018,wangAutomaticDetectionWind2017}
and climbing
robots~\cite{sahbelSystemDesignImplementation2019,rodriguezClimbingRingRobot2008,leeMaintenanceRobot5MW2016}
for blade and tower inspections, remotely-operated and autonomous underwater
vehicles for subsea inspection and cable
installation~\cite{albiezRepeatedClosedistanceVisual2016,choEvaluationUnderwaterCable2020},
and multi-robot systems capable of carrying out multiple tasks at
once~\cite{bernardiniMultiRobotPlatformAutonomous2020}. Solar farms have similar
opportunities for autonomous inspection and maintenance, particularly for
cleaning solar
panels~\cite{solarcleanoSolarPanelCleaning,ecoppiaRoboticSolarPanel} and
trimming
vegetation~\cite{renuroboticsRenuRoboticsAutomation,vectormachinesVectorAutonomousMower},
but mature robotic solutions already exist in these areas.

As discussed in Section~\ref{energy:autonomy:perception}, electric utilities
also spend substantial amounts of time and money inspecting distribution and
transmission infrastructure with the aim of reducing the risk of power outages
and
wildfires~\cite{panossianPowerSystemWildfire2023,xcelenergyHearingExhibit1022020}.
A typical utility may have hundreds or thousands of miles of power lines to
inspect for issues like rotting poles and nearby trees, and the current practice
is for a human operator to drive a drone to each inspection site in
turn~\cite{xcelenergyHearingExhibit1022020}. Reliable autonomy for beyond visual
line of sight operations, where enabled by new regulations, could be a key
enabling technology for increasing the frequency and reducing the cost of
inspection. Thermal utilities, both for traditional natural gas delivery and
emerging businesses like district heating and geothermal, also face challenging
inspection needs, complicated by the fact that most of their assets are
underground. Research in pipe inspection and repair robots could help these
utilities extend the life of existing pipelines (avoiding new investment in
fossil fuel infrastructure) and support new decarbonized thermal networks in the
form of district heating and
geothermal~\cite{ismailDevelopmentInpipeInspection2012}.

\vspace{1em}
\begin{boxMarginLeft}
      \subsubsection*{Future Directions: Field Robotics \& Manipulation in Energy}
      \begin{itemize}
            \item \emph{Reliability and systems engineering.}
                  Areas like drone inspection or material delivery on large
                  construction sites are relatively mature, but may require systems
                  engineering, integration, and ruggedization effort. The robotics
                  community should prioritize sharing best practices and
                  highlighting engineering efforts in building robust, field-ready
                  systems.

            \item \emph{Construction and assembly automation.}
                  Applied research in manipulation can help reduce costs and improve
                  working conditions, for example through autonomous solar panel
                  installation.

            \item \emph{New platforms for underground operations.}
                  Research into novel robotic platforms could enable
                  valuable new use cases, particularly for underground
                  construction. This could help utilities move distribution lines
                  underground, improving resiliency and reducing wildfire risk, and
                  could also support new decarbonized thermal networks.
      \end{itemize}
\end{boxMarginLeft}


\titledsubsection{Conclusion}

In this section, we provide an introduction to the challenges of energy sector decarbonization and survey several areas where we see an opportunity for autonomy research to have a positive impact. These include opportunities both for traditional robots to help build and maintain energy infrastructure and for cross-disciplinary transfer of robotics and autonomy algorithms (in fields like controls and estimation) to the energy domain, building upon the rich history of research in power systems and other energy domains. In each area, we provide references to the state of the art in both energy systems and autonomy research, and we highlight impactful open research questions for the interested reader to explore. We also provide a guide to the various stakeholders whose needs, challenges, and incentives will drive adoption of innovations based on robotics and autonomy research. We hope that this discussion will provide a starting point for future collaboration between the energy and autonomy communities.

Throughout this section, our focus has been on applications of autonomy research to electric power and thermal energy networks. As a result, we chose not to cover important areas of research outside of robotics and autonomy; for example, power flow~\cite{cainHistoryOptimalPower2012,capitanescuStateoftheartChallengesFuture2011} and forecasting for both load and generation~\cite{dontiMachineLearningSustainable2021,hongProbabilisticElectricLoad2016}. Further, while we see many areas where autonomy research can have a positive impact in the energy sector, the energy transition is not purely a technological problem. There are several areas where technological innovation alone is not sufficient and policy change is needed. For example, the ability to build long-distance transmission lines in the United States is limited primarily by regulatory requirements and local permitting rules~\cite{doeofficeofpolicyQueuedNeedTransmission2022}. Similarly, while there are opportunities to optimize the operation of the electric grid, privately-owned utilities will typically only adopt these technologies if regulators provide appropriate financial incentives~\cite{lebelImprovingUtilityPerformance2023}. Fundamentally, decarbonizing the energy system is a problem at the intersection of technology, economics, and public policy, and researchers can maximize their impact by understanding not only the opportunities for technological innovation but also how those technologies fit into the larger context of our changing energy system.

\clearpage

\autsection{Built Environment}{Norhan Magdy Bayomi, Shide Salimi, Matthew Kramer, and John E. Fernandez}
\label{sec:built-environment}

\begin{boxMarginLeft}
      Cities, while home to more than half of the world's population and serving as centers of economic and cultural growth, are nevertheless responsible for 37\% of global emissions. The built environment faces pressing challenges for both climate change mitigation (reducing the emissions due to constructing, renovating, and operating the world's buildings) and adaptation (becoming increasingly resilient to extreme heat and storms). This section provides an introduction to these challenges in the built environment and identifies high-potential areas for robotics and autonomy to support climate mitigation and adaptation, including:
      \begin{itemize}
            \item \emph{Controls \& Planning}: Optimizing the control of both individual buildings and city-scale infrastructure to improve energy efficiency, reduce impacts on the electrical grid, and be more robust to extreme weather.
            \item \emph{Estimation}: Modeling thermal performance of existing buildings to guide targeted energy efficiency retrofits and monitoring cities for urban heat islands and other climate risks.
            \item \emph{Perception}: Supporting inspections and maintenance of existing buildings and infrastructure, and monitoring progress in new construction.
            \item \emph{Field Robotics \& Manipulation}: Automating retrofitting tasks like installing insulation in cramped spaces and supporting disaster response in the built environment.
      \end{itemize}
\end{boxMarginLeft}

The buildings sector is responsible for 37\% of global emissions~\cite{UNEP2022Buildings}, including emissions from both construction (and key inputs like steel and cement) and operations. In addition to needing to address these emissions, the built environment also faces pressing adaptation challenges: rising temperatures and increasing prevalence of extreme weather mean that cities will need to become increasingly adaptable and resilient~\cite{nasaEffectsNASA, voxAmericaWarming}. While urbanization helps support rising global standards of living, it is progressing fastest in less developed regions that may be more vulnerable to extreme heat~\cite{un_wup2018}, with associated risks of heat-related illness and death~\cite{magdy2019urban, kilbourne1997heat, cdc2019natural, kenny2010heat}.

Meeting the dual challenge of improving comfort and resilience while minimizing energy use and environmental impacts will define the coming decades of global development in the built environment. In this section, we identify opportunities for robotics and autonomy to support mitigation and adaptation on both building and urban scales. These opportunities include:

\textbf{Resilient \& Efficient Building Systems:} Robotic technologies enable building-level adaptation through automated envelope adjustments, smart maintenance systems, precision retrofits, and advanced building controls. These systems enhance occupants’ adaptive capacity during disruptions while optimizing energy efficiency for climate mitigation.

\textbf{Urban Infrastructure Management:} Robotic platforms monitor and maintain critical urban infrastructure, from automated stormwater systems to structural health monitoring. These systems enable more resilient cities while reducing energy consumption through optimized operations.

\textbf{Resource Efficiency:} Robotics support material and energy efficiency through automated waste sorting, selective demolition for material recovery, and precision construction techniques that reduce waste. These approaches address embodied carbon while supporting circular economy principles.

\Cref{sec:buildings:challenges} provides an introduction to the
challenges of climate mitigation and adaptation in the built environment,
structured around these core themes. Following this introduction,
\Cref{sec:buildings:controls,sec:buildings:estimation,sec:buildings:perception,sec:buildings:fieldrobotics-and-manipulation}
explore how research in different robotics and autonomy fields can help solve
these challenges and support sustainable development in the built environment.

\titledsubsection{Challenges in the built environment}\label{sec:buildings:challenges}

Climate mitigation and adaptation in the built environment can be addressed at
multiple levels. At the physical level, this includes improving efficiency and
resilience in both new construction and retrofits. At the systems level, it
involves optimizing operations across individual buildings and entire cities.

\titledsubsubsection{Adaptation and efficiency for new building construction}

For new buildings, climate adaptation means designing structures that are
inherently resilient to extreme conditions. A key concept is passive
survivability~\cite{trundle2023urban} --- a building’s ability to maintain livable
indoor conditions without relying on mechanical systems. This is especially
critical during prolonged power outages or fuel disruptions, which are becoming
more frequent due to climate change. Passive strategies also reduce overall
energy use, supporting both adaptation and mitigation goals.

Robotics and autonomy can enhance passive survivability during both design and
construction. Advanced tools can optimize building orientation, materials, and
envelope systems to improve natural ventilation, daylighting, and thermal
mass~\cite{buhl2023review}. For example, Autodesk's TileGPT helps design
buildings that maximize passive heating and cooling~\cite{liyanage2024climate},
while The Edge in Amsterdam uses smart energy management and efficient design to
operate on roughly 30\% of the energy required by a typical office building of
similar size~\cite{andersson2022urban}.

At the construction stage, robotics can aid in precision construction of high-performance building components, such as double-skin facades or advanced insulation systems~\cite{dfabhouse2020}. For example, the Hadrian X robotic bricklayer can construct walls with high precision~\cite{peiris2023smart}, and the SAM100 (Semi-Automated Mason) has been deployed for bricklaying and facade retrofitting, which help enhance the building envelope and reduce heat transfer, contributing to increased thermal performance~\cite{madsen2019sam100}. Additionally, automated systems can be designed to adjust building shading, window apertures, or natural ventilation pathways based on changing environmental conditions, optimizing passive performance.

\titledsubsubsection{Adaptation of existing building stock}\label{sec:buildings:overview:existing_buildings}

Roughly 40\% of buildings that will exist in 2050 have already been built~\cite{UNEP2024}. These existing buildings, often constructed before modern sustainability and resilience standards, are ill-equipped to handle intensifying heat and extreme weather from climate change.
As a result, retrofitting existing buildings to enhance resilience and adaptability, as well as reduce emissions through improved efficiency, is critical~\cite{trundle2023urban, buhl2023review}.

Older buildings present unique challenges due to aging infrastructure, legacy materials, and design limitations that do not account for modern climate science and building technology. While these buildings often require substantial upgrades, there are financial, technical, and regulatory obstacles to achieving climate resilience at scale~\cite{peiris2023smart, tomrukcu2024climate}.
Adapting these buildings to improve resilience and reduce energy use will require retrofitting strategies that are both resource-efficient and minimally disruptive~\cite{liyanage2024climate, andersson2022urban}.

One of the primary challenges lies in enhancing structural resilience. Buildings that were not originally designed to endure extreme weather events, such as flooding or high winds, may need reinforcements to foundational and structural elements, adding complexity and cost to retrofit projects. Moreover, many buildings, particularly those with historical significance, face regulatory constraints that limit the type of modifications allowed~\cite{stephenson2014new, aigwi2023adaptive}.

Energy efficiency is another critical area, as older buildings frequently fall short of modern efficiency standards. Retrofits aimed at reducing emissions and enhancing energy performance require extensive upgrades to insulation, windows, HVAC systems, and lighting. However, integrating these improvements without compromising structural integrity or disrupting building operations poses technical challenges, especially in densely populated urban areas where construction space is limited~\cite{perera2023challenges, panakaduwa2024identifying}.

Finally, limited availability of sustainable materials and skilled labor exacerbates the challenges of retrofitting. The demand for eco-friendly materials and specialized skills in resilient construction often exceeds supply, driving up costs and extending project timelines. Addressing these challenges requires both innovation in retrofit technologies and supportive policy frameworks that incentivize climate adaptation measures and reduce financial and regulatory burdens on retrofit initiatives~\cite{chen2023artificial}.

Robotics and automation are emerging as tools to address these challenges. As retrofitting needs grow in scale and complexity, robotics can provide precision, efficiency, and the ability to perform tasks in environments that may be hazardous or inaccessible to human workers. For example, drones equipped with thermal cameras can identify temperature differences in a building's structure, helping to pinpoint areas of energy loss, leaks, or other potential issues. This approach is faster and safer than traditional methods, making it practical for large-scale energy efficiency assessments of urban areas~\cite{bayomi2023eyes}. Robots can also assist with precision retrofitting, applying insulation, drilling, and reinforcing structures~\cite{massy2018qbot, salonvaara2023robotically,madsen2019sam100}.
Precision in these applications minimizes material waste and reduces the likelihood of errors that could compromise the building’s resilience. Robotics can also work in confined or high-risk spaces, performing necessary upgrades in areas that would otherwise pose safety challenges for human labor~\cite{larsen2024robotic}. These capabilities make robotics an increasingly valuable asset in enhancing the resilience and sustainability of the built environment~\cite{fleckenstein2022climate, halder2023robots}.

\titledsubsubsection{Building operations and maintenance}

Building operations and maintenance are central to reducing environmental impact
and improving resilience. Emerging tools for optimization and estimation are
enabling smarter, more adaptive systems that manage energy use efficiently,
enhance occupant comfort, and lower operational costs --- without compromising
indoor environmental quality. As climate change reshapes building demands,
integrating these technologies will be essential for both adaptation and
mitigation.

To support building operations, particularly energy management, there is a need
to not only monitor building conditions but also use those measurements to
optimize the performance of heating, ventilation, and air conditioning (HVAC)
systems. Mobile robots may be used to gather real-time data on indoor
temperature, humidity, and air quality, as in~\cite{geng2022robot,adanRobotThermalMonitoring2023}, feeding data
into control and system identification algorithms can help optimize
building-level HVAC performance. Autonomous building systems could optimize HVAC
operations, adjust lighting based on occupancy, and regulate energy usage in
response to real-time environmental conditions. This dynamic energy management
can not only reduce emissions but also enhance the adaptive capacity of
buildings, allowing them to respond flexibly to changing climate conditions
\cite{ding2024potential, wolf2022data,cho2023intelligent,gabriel2024machine}.
This flexibility could also support the electricity grid, with some studies
estimating that grid-interactive buildings could save \$100–200 billion in U.S.
power system costs by 2040 and reduce \coo~emissions by 80 million tons annually
by 2030. However, barriers like technological interoperability, insufficient
consumer incentives, and limited workforce training need to be addressed through
policy reform, financial incentives, and enhanced technology integration
\cite{satchwell2021national}.

For building maintenance, robotics can help inspect, clean, and maintain building envelopes. These tasks are often dull, dirty, and dangerous --- for example, inspecting the facade on a high-rise building --- making them ideal candidates for automation~\cite{salamah2022effect}. Multi-robot systems can combine imaging with non-destructive testing for building envelope inspections, for example combining aerial drones, ground-penetrating radar for detecting defects in building materials, and quadruped robots to assist human building inspectors~\cite{cheneaseebot}.
These advancements can increase on-site productivity and enhance safety while also improving the consistency and quality of envelope maintenance, directly impacting building energy efficiency by contributing to better thermal performance.

Beyond physical robotics, autonomy algorithms can contribute in areas like predictive maintenance. By analyzing operational data from building systems, these methods can predict potential failures in HVAC equipment and other building systems before they occur. By ensuring that necessary repairs are conducted proactively, building operators can minimize downtime and extend the life of building equipment.

\titledsubsubsection{Urban systems: development, performance, and adaptation}

Cities face mounting adaptation challenges as climate change intensifies urban vulnerabilities. Key challenges include increasing frequency of extreme heat events, intensifying urban flooding, infrastructure strain from extreme weather, and the need for rapid emergency response systems. Climate adaptation is reshaping urban development, requiring more dynamic and responsive urban systems: building energy systems must adapt to more extreme temperature variations, infrastructure must withstand intensified weather events, and urban services must maintain consistency despite climate disruptions. These challenges are particularly acute in coastal megacities, where sea-level rise compounds existing urban stressors.

For example, many cities will experience greater rainfall and flooding. Stormwater management systems must be upgraded to handle these events.
Dynamic stormwater system enables cities to autonomously monitor and adjust their water systems during extreme precipitation events~\cite{shishegarOptimizationMethodsApplied2018,webber2022moving}. Recent implementations in Saudi Arabia demonstrated that robotics-enabled stormwater management systems help reduce urban flooding during extreme rainfall events~\cite{SWM2024StormwaterRobots}. Improvements in water quality monitoring sensors may also enable cities to more quickly detect health and environmental hazards in their stormwater and sewage systems. Complementing these systems, real-time infrastructure stress monitoring through networked robotic sensors can provide insights into structural health under extreme weather conditions, enabling proactive maintenance and reducing failure risks~\cite{ramirez-morenoSensorsSustainableSmart2021}%

A warming climate will also exacerbate existing problems like urban heat islands. Robotic tools like mobile weather stations can help cities understand where heat mitigation efforts are most needed. They can also provide people with hyperlocal temperature information so they can avoid dangerously hot areas.
To adapt, cities will need to identify areas for resilient development. Robotics and autonomy tools can help cities monitor their current physical state to build a digital twin: a computer-based model of the city~\cite{shahat2021city}. Scenario planning tools can then enable them to evaluate the effects of different climate-related events like storms, floods, and heat waves~\cite{hao2024empowering}. This allows for more robust city development plans in the face of an uncertain future. These tools may also help cities use real-time climate data to optimize building practices, while monitoring and inspection robots ensure compliance with climate-resilient building standards.

In addition to serving cities' adaptation needs, robotics is emerging as a dual-purpose solution that can also support urban performance optimization. These systems enhance cities' responsive capabilities through comprehensive adaptive infrastructure management and performance optimization approaches. In adaptive infrastructure management, advanced robotics enable automated building envelope adjustments that respond to microclimate variations, significantly improving thermal performance and energy efficiency~\cite{alkhatib2021deployment,favoino2022embedding}.

Finally, there is an acute need for improved tools for early warning and emergency response.
Early warning systems give people a chance to seek shelter before a disaster reaches them. Improvements in sensor technology can make these systems more reliable and less expensive so that they may save lives in cities that cannot currently afford them.
Unmanned robotic vehicles can improve emergency responses to disasters. Aerial robots can provide important information about the scene to responders on the ground. Ground vehicles can move through dangerous areas, like collapsed buildings, to gather information, find victims, and clear hazards for human responders. Improvements to these robots can help save lives while reducing risk to first responders~\cite{delmericoCurrentStateFuture2019}.

\titledsubsubsection{Waste management \& circular economy}

Construction waste management primarily involves reducing the amount of
construction, renovation, and demolition (CRD) waste generated throughout a
building's lifecycle. The existing building and infrastructure stock is
responsible for 89\% of CRD waste~\cite{perry2014characterization}, 40\% of
material resources~\cite{perry2014characterization, gorgolewski2017resource} and
40\% of generated solid waste~\cite{eberhardt2022building}. CRD is large in
volume, mixed in nature, and potentially hazardous, posing challenges for
landfills~\cite{ccme2019guide}. Even materials that can be recycled or reused,
like wood, concrete, and metals, often end up in landfills, with negative
environmental effects such as methane emissions and water
pollution~\cite{luciano2020resources}. Proper management of this waste,
including both reduction and recovery, is a promising climate mitigation
solution.  While traditional waste management focuses on handling debris from
construction and demolition, a circular economy (CE) approach emphasizes reusing
materials, components, and even subsystems to minimize waste and reduce demand
for raw resources.  This might involve choosing materials for their durability
and recyclability, employing modular designs, and using demountable connections
to ensure that components are reusable in future constructions.

While different, both waste management and CE play significant roles in climate mitigation. Waste management reduces landfill use, thus lowering methane emissions, which is a potent greenhouse gas. By reusing materials and reducing virgin material extraction, CE can reduce the need for energy-intensive processes and associated emissions~\cite{panizza2024building}. Together, these approaches help decrease the carbon footprint of the built environment and building operations.

In recent years, automation and digitalization have emerged as effective tools to support both CE and waste management. These tools, which include building information modeling, computer simulation, remote sensing, design automation, and deconstruction robotics, can support waste sorting and recylcing planning, optimizing, operations, leading to more efficient material recovery and reuse~\cite{allam2024supporting}.
For example, computer vision can be adopted to automate the identification and classification of CRD waste~\cite{li2022rgb, lin2022deep, na2022development}, and robotic systems can perform tasks such as selective demolition and waste sorting to improve precision in material recovery and waste management. Remote sensing and the Internet of Things (IoT) can help with material tracking and lifecycle management, and machine learning models can predict life expectancy for materials and components, enabling better planning for material reuse. Automation in manufacturing, mostly through prefabrication, modular, and offsite construction can both reduce waste during construction and support later de-constructability and recoverability of the components.

\titledsubsubsection{Urban forestry and carbon sequestration}

Urban forests represent critical infrastructure for climate change mitigation and adaptation, serving to both sequester carbon and reduce the effects of urban heat islands, yet urbanization pressures and increasing climate stress makes managing these forests increasingly challenging. Recent advances in robotics technology are transforming urban forestry practices, making tree maintenance more efficient and optimizing the carbon sequestration potential of urban landscapes. For example, high-precision robotic sensing platforms now enable continuous monitoring of urban tree populations, providing real-time data on growth rates, carbon sequestration capacity, and early detection of stress factors~\cite{ferreira2023sensing}. Autonomous pruning systems equipped with advanced vision algorithms can now perform selective maintenance while optimizing for carbon sequestration potential. Studies in metropolitan areas have shown that AI-assisted tree maintenance can help improve carbon storage capacity through more precise canopy management and improved tree health outcomes~\cite{nitoslawski2019smarter}. Autonomous soil monitoring and amendment systems now enable precise management of soil conditions that optimize tree growth and carbon storage. Recent implementations in Singapore demonstrated that robotics-enabled soil management increased soil organic carbon content by 30\% while improving urban tree survival rates~\cite{moraitis2022design}.

\titledsubsection{Controls \& planning}
\label{sec:buildings:controls}

\titledsubsubsection{Building energy management \& grid interaction}

Effectively managing building energy use is important for reducing operational costs and minimizing environmental impact. Robotics and autonomy research can support advanced building energy management systems (BEMS) by enabling real-time optimization, enhanced efficiency, and integration with renewable energy sources. These technologies can transform buildings into dynamic systems capable of responding intelligently to changing conditions.

There is a long history of applying control methods like model predictive control (MPC) to minimizing energy use from HVAC, in some cases reducing cooling energy loads by up to 30\%~\cite{lee2023artificial}. While these methods have most commonly been applied to commercial buildings~\cite{green2024paretooptimized,killian2016mpcbuildings,drgona2020buildingmpc}, increasing deployment of smart thermostats creates opportunities for optimization in residential buildings as well~\cite{shareef2018reviewhems,beaudin2015hemsreview}. Previous works have applied MPC~\cite{killian2016mpcbuildings,drgona2020buildingmpc}, adaptive control~\cite{nesler1986adaptive,Shaikh2014review}, robust control~\cite{Anderson2008robust,Shaikh2014review}, and reinforcement learning~\cite{wang2020rlreview} to building control problems. Rather than provide a comprehensive survey of this vast literature, we highlight two open areas where control research can support climate adaptation and mitigation.

The first focus area is more deeply integrating BEMS with the electric grid, allowing buildings to serve as distributed energy resources. This area has attracted substantial research interest over the years, but challenges remain both in theory and in practice~\cite{elsisi2023comprehensive}. By shifting energy use during peak periods, buildings can help stabilize the grid and enable increased renewable energy penetration. However, effectively shifting energy use without compromising occupant comfort will require new hybrid forecasting and optimization models that can simultaneously consider a building's thermal performance~\cite{cho2023intelligent,gabriel2024machine}, occupant behavior~\cite{salimi2020optimizing}, generation from rooftop solar photovoltaics~\cite{barchi2021photovoltaics, aleem2020review, singh2024adaptive}, and grid conditions~\cite{ghasemi2016novel}. On a practical level, cybersecurity concerns, computational demands, and high implementation cost have all slowed development of grid-interactive building technologies. Future efforts should focus on developing cost-effective, secure, and adaptable systems while fostering collaboration and standardization to streamline research and implementation \cite{elsisi2023comprehensive}.

The second focus area concerns climate adaptation: using advanced building control to improve resiliency to extreme weather. As climate change increases the frequency of extreme heat waves, building energy management will play an important role in preserving occupants health and safety. Active building control can layer with passive energy efficiency measures to improve survivability. Autonomous grid-interactive buildings can help reduce the load on an electrical grid during a heat wave~\cite{Choobineh2016energyheat, Mondal2024}. Adaptive building envelopes can improve thermal comfort and resilience by dynamically changing a building's fa\c{c}ade in response to environmental conditions, for example adjusting the positions of shading panels, smart glass transparency, and photovoltaic panel orientation to optimize thermal performance, occupant comfort, and resiliency~\cite{favoino2022embedding, tabadkani2020review, alkhatib2021deployment, al2017building, allen2017smart,svetozarevic2016soro, stelzmann2024development,wang2023framing}. Integrating the control of these active building components with the internal HVAC optimization is a particularly exciting opportunity for research.

In addition to these research challenges, there are also several practical challenges to adoption and commercialization of advanced building control, including data access, integration with existing systems, and cost and speed of deployment~\cite{Henze2025whyhvac}. Overcoming these challenges will require not only effort by controls researchers, but also cross-disciplinary collaboration with the building science community, HVAC engineers, and building operations professionals.

\titledsubsubsection{Building design optimization}

Generative design is an advanced architectural approach that leverages algorithms and computational tools to create building designs optimized for various criteria, including climate resilience. By simulating and optimizing thousands of design iterations for climate adaptability and sustainability, architects can create buildings that are not only energy-efficient but also better equipped to withstand the effects of climate change, such as extreme weather events, rising temperatures, and resource scarcity~\cite{weber2022floorplan,Weber2020embodied,Suphavarophas2024reviewgenerative,Zhang2021generative}. For example, the Al Bahr Towers in Abu Dhabi use generative design to develop a dynamic facade that responds to sunlight. The structure features a honeycomb shading system that opens and closes in response to the intensity of sunlight, reducing solar heat gain and minimizing the need for air conditioning. This adaptive shading is optimized to ensure that the building remains energy-efficient in extreme heat, reducing carbon emissions and promoting climate adaptation and mitigation~\cite{attia2018albahr}.

Generative design can also be applied to plan climate-resilient urban housing, simulating how different housing layouts, materials, and technologies would perform under various climate scenarios, such as flooding or extreme heat. These tools could produce optimized designs that integrate features like raised building foundations for flood protection and natural cooling to reduce urban heat island effects. The generative approach helps create a community layout that is adaptable, energy-efficient, and capable of mitigating the impacts of climate extremes~\cite{Zhang2021generative,Shafiei2022coastalgenerative}.

\titledsubsubsection{District heating systems}
More than 40\% of the energy used in commercial buildings is for heating and cooling~\cite{eiaEnergyCommercial}. For homes, it’s over 50\%~\cite{eiaEnergyHomes}. Many buildings burn fossil fuels on-site to generate heat and run their air-conditioning systems to cool down. Cities and other densely populated areas can significantly reduce the energy required to keep their buildings at comfortable temperatures by adopting district heating and cooling (DHC) systems. DHC heats or cools water at central locations and then distributes it to customers via a network of pipes. Individual buildings can use the heated or cooled water as needed to maintain their desired temperature.

Where DHC exists, it can be made more efficient by predicting demand. Smart thermostats in buildings can learn heating and cooling patterns based on their occupants’ behaviors and environmental conditions and share these patterns with the central plants~\cite{wirtz2021temperature}. This can be used to predict future loads, and match the generation more closely to demand to limit waste.

In areas that want to adopt DHC, robotic tools can help design the system to meet its customers’ needs efficiently. Smart thermostats and remote sensing technologies can be used to estimate the heating and cooling demand of each building that would be served~\cite{li2017district}. Network optimization tools can help to locate critical components like pumps, valves, and junctions to ensure various demand conditions can be served efficiently, and to ensure resilience against single points of failure. \Cref{sec:energy} discusses thermal energy networks in greater detail.

\titledsubsubsection{Stormwater management systems}
Facilities for capturing and managing stormwater are critical for protecting a city’s residents and infrastructure. Cities are facing challenges in adapting their aging stormwater systems to the needs of the present and the future. They rely on passive designs that cannot be easily manipulated to handle growing cities and extreme weather events. Robotic solutions can offer increased control, monitoring, and capacity of existing systems, and can offer guidance for the design of new systems~\cite{webber2022moving}.

The capacity of an existing sewer system can often be augmented by the use of active controls. If water levels and flow rates are measured at key points in the system, this information can be used to actuate flow controls to better direct water through the system. Short-term weather forecasts and monitoring tools can be used to predict future system states and proactively control the system to maintain acceptable water levels~\cite{xu2021enhancing}.

Cities can benefit from improvements in water quality monitoring technologies. Samples from sewer systems can be used to detect chemical and biological hazards in the city. This practice is currently limited by the time and cost required to detect contaminants. Tools to enable real-time water quality monitoring could be used to detect a wider range of contaminants and detect them earlier~\cite{razguliaev2024urban}.

\begin{boxMarginLeft}
      \subsubsection*{Future Directions: Controls \& Planning in the Built Environment}
      \begin{itemize}
            \item \emph{Predictive and online energy management.}
                  Real-time optimization and control can help minimize energy
                  consumption and electric grid impacts. Improved modeling that
                  considers the impact of building control on occupant comfort
                  can help support adoption and acceptance.

            \item \emph{Improved design tools.}
                  Optimal design tools can help optimize building and urban
                  layouts for both energy efficiency and resilience to extreme
                  weather events. New methods that integrate optimal design with
                  building control can help simultaneously advance mitigation
                  and adaptation for new construction.

            \item \emph{Adaptive infrastructure networks.}
                  At the city scale, active monitoring and control can help
                  optimize infrastructure networks, particularly stormwater
                  management (providing adaptation benefits) and district
                  heating networks (providing mitigation benefits through
                  improved heating efficiency).

            \item \emph{Focus on ease of adoption.}
                  Across these themes, researchers can support sustainable
                  global development and adaptation through a focus on
                  developing scalable, cost-effective tools (for example,
                  adaptive building controllers that can easily integrate with
                  existing systems).
      \end{itemize}
\end{boxMarginLeft}

\titledsubsection{Estimation}
\label{sec:buildings:estimation}

\titledsubsubsection{Building thermal, environmental, and occupancy modeling}

Most energy use in buildings is due to heating and cooling. As discussed in Section~\ref{sec:buildings:controls}, optimizing these loads can help improve energy efficiency and reduce strain on the electric grid, but this often requires a model of each building's thermal performance and occupant behavior to predict when and where to heat or cool to maximize comfort and minimize energy usage~\cite{chenBuildingOccupancyEstimation2018}. Estimation and system identification tools can help develop these models, combining data from networked and mobile sensors with physics-based building models to help predict and optimize energy needs~\cite{knudsenSystemIdentificationThermal2017,bouacheIdentificationThermalCharacteristics2013,radeckiOnlineModelEstimation2017,radeckiOnlineBuildingThermal2012}. Similar methods can be applied to monitor indoor air quality~\cite{kousis2021intra,meier2017crowdsourcing,esposito2022recent}, improving occupant comfort and health.

Historically, building model identification and estimation has relied on a network of fixed sensors, which may lack coverage or have high up-front costs that limit adoption. An emerging area of research is augmenting data from static sensors with data from mobile robots, which can combine thermal imaging, \coo~sensors, and computer
vision to create dynamic occupancy maps with high spatial and temporal
resolution~\cite{zou2018device,bayomi2021building,cheng2023multi}. While fixed sensors must be permanently installed in each building, detailed thermal models may only need to be updated periodically. As a result, one contractor with a fleet of mobile sensors would be able to cost-effectively support multiple buildings. Further research could help integrate active sampling with system identification to plan an optimal sequence of measurements to support thermal modeling and performance optimization~\cite{zhangDatadrivenBuildingEnergy2021}. Feeding these data back into building information modeling (BIM) systems could support continuously-updated digital twins that accurately reflect the state of a building over time, allowing for more accurate energy simulations and predictive maintenance of building systems.

\titledsubsubsection{Structural monitoring}

As climate change increases the frequency of extreme weather events, continuous
monitoring of critical infrastructure is essential for resilience and public
safety. Structures must be periodically inspected and repaired, but traditional
inspections are time-consuming, expensive, and may miss early-stage damage.
Automated estimation tools can constantly monitor buildings, roads, and other
infrastructure to detect defects early.

Structural health monitoring systems measure and analyze vibrations to identify
signs of wear or failure~\cite{cha2024deep, azhar2024recent}. These systems can
be installed in key locations --- especially those difficult to access --- and issue
alerts as soon as issues arise. They are particularly valuable for
infrastructure subject to heavy loads or environmental stress, such as tunnels
and bridges. Vehicle-based sensors can also monitor roads and railways during
normal operation, reducing the need for disruptive shutdowns and
resource-intensive inspections~\cite{mishra2020road}.

\titledsubsubsection{Urban heat islands and climate risk}

An urban heat island occurs where the air temperature within a city is significantly higher than the surrounding suburban or rural regions. Heat islands occur because buildings and roads retain heat, while natural features such as green spaces dissipate heat~\cite{kousis2021intra}. These conditions can pose a danger to a city’s inhabitants. It is important to understand the factors that cause urban heat islands so that we can identify and mitigate them in our cities today, and design cities that prevent them in the future.

Advanced monitoring tools, supported by estimation and autonomy tools, can help
build our understanding of heat islands and develop mitigation strategies.
Although most cities have a network of fixed-position weather stations, these
provide only a sparse set of data points that can miss important climate
variations within the city. Remote sensing satellite technologies can provide a
dense map of temperature measurements across a city, but the sampling frequency
may be limited~\cite{zhou2018remote,elmarakby2024prioritising}. Remote sensing
also struggles to provide data for other important measurements, such as wind
speed and relative humidity.
Augmenting traditional data sources with mobile weather stations (e.g., on
vehicles) and crowdsourced observations can improve spatial and temporal
resolution~\cite{kousis2021intra,meier2017crowdsourcing}, but also introduces
challenges in assimilating large volumes of mobile sensor data.
Estimation research could help integrate urban weather monitoring with
localization for mobile sensor data to create high-resolution maps of heat risk
real-time, with the eventual goal of learning city-scale models to predict
microclimate variation and heat
risk~\cite{kousisIntraurbanMicroclimateInvestigation2021,kousisEnvironmentalMobileMonitoring2022}.

In addition to helping understand and mitigate urban heat islands, pervasive, low-cost urban sensing networks can also support early warning and response systems for extreme weather and other natural disasters~\cite{esposito2022recent,acosta-collRealTimeEarlyWarning2018}.
Especially for coastal cities, early warning systems could also benefit from robotic sensors deployed outside the city boundaries; for example, uncrewed surface vehicles could help measure hurricanes to forecast their future evolution~\cite{zhangHurricaneObservationsUncrewed2023}. The role of robotics in collecting data for climate and weather modeling is discussed further in \Cref{sec:earthsciences}.

\begin{boxMarginLeft}
      \subsubsection*{Future Directions: Estimation in the Built Environment}
      \begin{itemize}
            \item \emph{Adaptive sampling for environmental monitoring.}
                  Active and adaptive sampling methods, especially to
                  co-optimize static sensor networks with measurements from
                  mobile robotic platforms, can support a wide range of use
                  cases in the built environment, from building modeling to
                  city-scale environmental monitoring.

            \item \emph{Estimation for digital twins.}
                  Better occupancy and thermal modeling can support both
                  targeted energy efficiency upgrades and optimized building
                  energy management systems. Further research in this area can
                  focus on updating these models over time to create dynamically
                  updating digital twins for optimizing building performance.

            \item \emph{Scalable mobile sensing.}
                  Lowering the cost of sensing, for example by using autonomous
                  mobile platforms rather than relying on fixed sensor networks
                  for coverage, can enable more widespread monitoring of both
                  infrastructure (e.g., structural damage in buildings, bridges,
                  etc.) and urban environments (e.g., heat islands and extreme
                  weather risks).
      \end{itemize}
\end{boxMarginLeft}

\titledsubsection{Perception}
\label{sec:buildings:perception}

\titledsubsubsection{Building energy audits}\label{sec:buildings:perception:audits}

As discussed in Section~\ref{sec:buildings:overview:existing_buildings}, retrofitting existing buildings with energy-saving measures can help improve both efficiency and resilience. However, since no two existing buildings are alike, there is a bottleneck in assessing these buildings' performance and retrofit needs. In the past decade, there has been an increasing use of unmanned aerial vehicles (UAVs) to enable rapid, detailed surveys of building exteriors, especially of facades, roofs, and other inaccessible areas. By capturing precise thermal imagery and 3D point cloud data, these systems can identify heat leaks, insulation gaps, or other issues, allowing contractors to tailor retrofits to individual building needs~\cite{jaw2013locational,van2018framework,shariqRevolutionisingBuildingInspection2020,bayAutonomousRoboticBuilding2017}. Opportunities for further research in this space include active perception methods to allow inspection robots to autonomously explore the exterior of a building, reducing the burden on the operator, and fusing the results of drone-based inspection with existing building information systems (e.g., CAD models, maintenance records, energy consumption data, etc.) to allow building operators to make informed retrofit decisions.

\titledsubsubsection{Construction site progress monitoring}

Construction sites are notoriously complex environments. As cities strive to
meet growing housing needs and implement low-carbon building practices, it
becomes even more critical to keep projects on schedule and minimize material
waste. Project managers must be able to monitor construction progress, verify
that completed tasks align with plans, and identify safety hazards. However, the
scale of most sites makes frequent walkthroughs prohibitively time-intensive.

Mobile robots like drones and quadrupeds can autonomously navigate construction
zones, capturing detailed visual and 3D data of ongoing work. By comparing this
data to Building Information Modeling (BIM) plans, AI-powered systems can
automatically detect discrepancies between planned and actual progress, enabling
early detection of delays or errors~\cite{choi2023overview,
      asadi2018vision,kropp2018interior}. Timely corrections help reduce rework and
avoid unnecessary emissions from duplicated effort.

Construction monitoring also poses challenges for robotic perception and
localization: environments change rapidly, sometimes adding new rooms or floors
in days. To support these applications, future research can prioritize SLAM
methods that remain robust in large-scale dynamic settings, with the goal of
enabling real-time model updates throughout
construction~\cite{yarovoiReviewSimultaneousLocalization2024}.

\titledsubsubsection{Underground utility mapping}

As cities work to adapt infrastructure for climate resilience --- such as expanding
electric grids for renewables or upgrading water systems for drought and flood
response --- accurate underground maps become critical.
Cities contain vast networks of underground utilities (electric cables, natural
gas and water pipes, etc.). When these need to be maintained, or when new
underground infrastructure is installed, we run the risk of damaging nearby
equipment. Projects rely on low-quality or outdated utility maps and sometimes
run into infrastructure that was not recorded at all~\cite{jaw2013locational}.
These incidents can cause environmental hazards and interruptions to critical
services, and they are a significant roadblock in the development of cities.
Robotic tools can provide a better understanding of where underground
infrastructure is located. Ground penetrating radar (GPR) is widely used to
detect obstacles, but its accuracy may be improved in complex environments
through the use of smarter scanning methods~\cite{jaw2013locational}. It may
also be supplemented with other mapping techniques, such as 3D GPR and
gyroscoping mapping~\cite{van2018framework}. Cities with existing underground
infrastructure may update their maps more frequently to allow new projects to
rely on them for planning. New and expanding cities may be able to use
previously undeveloped land without the risk of failed excavations.

\titledsubsubsection{Urban carbon sink \& biodiversity monitoring}

Urban carbon sinks, such as green spaces, urban forests, and constructed wetlands, can sequester carbon dioxide from the atmosphere, acting alongside natural ecosystems and agricultural land as important global carbon sinks. Accurate monitoring of these sinks is essential for quantifying their impact and informing urban planning strategies. Robotic systems are emerging as powerful tools for enhancing the precision, scale, and frequency of urban carbon sink monitoring. For instance, drones carrying LiDAR imaging sensors and hyperspectral cameras can create detailed maps of urban forests, providing accurate estimates of above-ground biomass and carbon storage potential~\cite{ma2024deep}. These systems can operate continuously, capturing temporal changes in carbon sequestration rates that might be missed by intermittent manual surveys. Recent studies have demonstrated the use of drones for estimating carbon stocks in urban trees with high accuracy~\cite{kuvzelka2018mapping}. These aerial platforms can access areas that may be challenging for ground-based robots, such as rooftop gardens or steep terrain, providing a comprehensive view of urban carbon sinks. These mobile platforms can augment stationary sensor networks, improving our understanding of urban carbon dynamics~\cite{singh2008mobile}. This multi-scale approach, combining ground-level robotic measurements with broader satellite observations, provides a more comprehensive picture of urban carbon sinks and their temporal variations.

Beyond their role in carbon sinks, robot perception can also support general maintenance and monitoring or urban green spaces like parks and gardens. For example, prior works have used UAV-based hyperspectral imaging to detect and map invasive plant species in urban areas~\cite{nasi2015using,dash2017assessing,ecke2022uav} and cameras carried by ground robots to continuously monitor urban plant health and growth patterns~\cite{droukas2023survey}. These monitoring systems could also be combined with automated maintenance systems, transferring methods from precision agriculture (discussed further in \Cref{sec:land-use}) to an urban context~\cite{lan2017current}.

\begin{boxMarginLeft}
      \subsubsection*{Future Directions: Perception in the Built Environment}
      \begin{itemize}
            \item \emph{Scalable scene perception.}
                  There are opportunities for perception to support use cases
                  throughout the built environment, from building-scale
                  construction and retrofits to city-scale infrastructure.
                  Perception and scene understanding methods that can adapt
                  across these different scales can support robust, scalable
                  deployments.

            \item \emph{Mapping and understanding dynamic urban environments.}
                  Urban environments are highly dynamic; at the extreme end of
                  the spectrum, construction sites can add new walls and rooms
                  on a daily basis. Perception research to support navigation,
                  inspection, and mapping in these changing environments can
                  help unlock new use cases for robotics in the built
                  environment.

            \item \emph{Multi-modal infrastructure monitoring.}
                  In areas like building energy audits and construction
                  monitoring, robotics can create new value by helping building
                  managers monitor changes in the built environment over time.
                  Researchers can support these use cases through tools that can
                  detect changes in buildings and urban infrastructure and
                  active perception methods that can autonomously decide which
                  regions of a building to revisit and inspect in greater
                  detail.
      \end{itemize}
\end{boxMarginLeft}

\titledsubsection{Field robotics \& manipulation}
\label{sec:buildings:fieldrobotics-and-manipulation}

\titledsubsubsection{Automated inspection}

Automated inspection requires both perception systems (discussed above) and mobile robotic platforms to access different types of infrastructure. In some applications, existing robotic platforms are sufficient; for example, tracked robots using ground-penetrating radar to inspect bridge decks~\cite{la2017development} or drones used to inspect building facades~\cite{bayomi2023eyes,choi2023overview}. However, in some particularly challenging settings, novel robotic platforms may be needed. For example, prior works have explored caged drones for inspecting inside HVAC ducts~\cite{borik2019caged}, snake robots for inspecting water and sewer pipes~\cite{dinh2022vision, kirchner1997prototype}, climbing robots for building exteriors~\cite{fang2023advances}, and crawling robots designed to inspect overhead transmission lines~\cite{pouliot2012field}. Even for mature platforms like UAVs, there are opportunities for multi-agent systems to conduct inspections of large-scale infrastructure like dams and bridges~\cite{schranz2020swarm,parrott2020simulation}. Across all of these use cases, novel robotic systems can not only reduce risk to human inspectors but even enable inspections of previously-inaccessible assets. While there is not a single clear agenda for future research on similar new inspection platforms, we hope that field roboticists will continue to collaborate with domain experts to discover new opportunities in this area.

\titledsubsubsection{Precision retrofits}

In addition to identifying regions of a building that need retrofits to improve energy efficiency (as discussed in Section~\ref{sec:buildings:perception:audits}), there are opportunities for robots to augment human workers in installing those upgrades. There is a large body of research on construction robotics~\cite{paraschoConstructionRoboticsAutomation2023,willmann2012aerial,vaha2013extending}; rather than reviewing this extensive literature, we focus on automated precision energy retrofits as a climate-relevant application of construction robots.

One of the most common retrofitting tasks, particularly in old buildings, is installing insulation in walls or crawlspaces. Humans cannot access the interior of walls without intense disruption to building occupants, and while many crawlspaces are technically accessible, they are cramped and difficult to work in. Prior works have developed robotic systems for inspecting and installing new insulation in crawlspaces~\cite{hollowayRobotSprayApplied2016,cebolladaMappingLocalizationModule2018} and walls~\cite{wangIntelligentSprayingRobot2022,lublasserRoboticApplicationFoam2018}. While blown-in insulation is a common, cost-effective method for retrofitting wall insulation, it can be challenging to install without unintentional gaps, which can lead to thermal leaks~\cite{lugano2000insulating}. As a result, there is potential for low-cost, easy-to-use robots with novel form factors (e.g., snake robots) to help retrofit insulation in existing walls.
Robots can also support the development of advanced building materials for retrofitting applications. For example, shape memory alloy (SMA) actuators, which change shape in response to temperature variations, can serve as dynamic shading systems for buildings, but they can be difficult to manufacture. Robotic fabrication systems for these advanced building components can enable more customized and efficient production~\cite{kwon20233d,yi20203d}.

As climate change intensifies the need for widespread building energy efficiency improvements, the use of robots for building retrofits is likely to expand. Robotic platforms have the potential to significantly accelerate the pace of retrofits while improving their quality and consistency. However, successfully integrating robotics into retrofit processes will require addressing challenges related to system reliability, cost-effectiveness, and integration with existing construction workflows.

\titledsubsubsection{Automated waste sorting, recycling, and material recovery}

Cities generate large amounts of waste, both from construction and from daily life. Managing this waste is a critical challenge for both quality of life as well as for climate change mitigation, as material recovery from recycling can reduce the need for raw materials. Sorting robots can integrate with existing recycling facilities, but the primary challenge is reliably detecting and classifying different types of recyclable materials, often using a combination of visual cues and material properties like density~\cite{koskinopoulouRoboticWasteSorting2021,chin2019automated,nevzerka2024machine}. More advanced use cases like automated disassembly and urban waste mining (discussed further in \Cref{sec:industry}) can extract valuable materials from otherwise difficulty-to-recycle items like electronics, but require accurate manipulation abilities~\cite{obrienMeetDaisy,xavier2021sustainability,ongondo2015distinct}.
At a systems level, data from automated sorting systems throughout the waste stream can help cities optimize waste collection routes, predict patterns in waste generation, and streamline their operations~\cite{esmaeilian2018future}.

\titledsubsubsection{Disaster response}

As climate change leads to more frequent and intense severe weather, cities must be able to not only prepare for but also quickly respond to natural disasters. First responders face many challenges: for example, assessing damage across hundreds or thousands of buildings is time consuming and potentially dangerous, and changing post-disaster conditions can limit access and situational awareness~\cite{kediaTechnologiesEnablingSituational2022}. Robots can help emergency personnel respond to these events~\cite{delmericoCurrentStateFuture2019a}. For example, unmanned ground and aerial robots can quickly gather information about an emergency site without putting human responders in danger, building detailed maps of both individual buildings (supporting damage assessment~\cite{kruijff2014designing}) and entire neighborhoods (supporting situational awareness~\cite{kruijff2015tradr}).

Post-disaster environments pose a number of challenges for existing robotics tools. First and foremost, domain experts cite the reliability of robotic systems as the primary barrier to widescale adoption; to be successful, disaster response robots must be able to reliably accomplish their missions with limited human oversight~\cite{delmericoCurrentStateFuture2019a}. Beyond a general focus on ruggedization, targeted research on localization and mapping systems that are robust to changing environments (e.g., when damaged structures shift or debris is moved) or on human-robot teaming for disaster response can also help improve performance and reduce barriers to adoption.

\begin{boxMarginLeft}
      \subsubsection*{Future Directions: Field Robotics in the Built Environment}
      \begin{itemize}
            \item \emph{Novel form factors.}
                  Morphologies with non-traditional locomotion capabilities
                  (e.g., climbing, crawling, and snake robots)
                  could unlock new use cases for inspection and maintenance of
                  traditionally inaccessible infrastructure.

            \item \emph{Reliable and rugged platforms.}
                  Reliability and ruggedization are barriers to adoption in
                  several critical domains. Addressing these needs will require both
                  improved physical platforms and improved reliability of perception,
                  navigation, and control systems.

            \item \emph{Close collaboration with end users.}
                  While robotics can help automate manual processes like
                  installing insulation, it is important that roboticists collaborate
                  with domain experts and workers to understand the specific problem
                  and design solutions that are compatible with existing workflows.
      \end{itemize}
\end{boxMarginLeft}

\titledsubsection{Conclusion}

Better buildings and cities are key to improving global quality of life. In this section, we have provided an introduction to the challenges of sustainable development and climate mitigation and adaptation in the built environment. Equipped with this context, we have highlighted several areas where robotics and autonomy research can help make the built environment more efficient, resilient, and comfortable. These include not only opportunities for physical robots to support inspection, retrofits, and monitoring of urban infrastructure, but also opportunities for computational tools from controls, optimization, and estimation to support a vision for ``self-driving'' buildings and cities that can autonomously improve efficiency and quality of life for their residents. Fully realizing this vision will require deep collaboration between roboticists and domain experts in the built environment, including not only researchers but also practicioners like architects, urban planners, and building managers. We hope that this section provides both a roadmap and a starting point for roboticists hoping to explore this promising intersection.

\autsection{Transportation}{Shreyaa Raghavan, Cameron Hickert, Jung-Hoon Cho, Vindula Jayawardana,  Matthew Kramer, and Cathy Wu}
\label{sec:transport}

\begin{boxMarginLeft}

    Transportation systems, spanning roads, rails, ports, and airways, are
    vital to global mobility but also major contributors to emissions,
    pollution, and resource use. Robotics and autonomy offer tools to reduce
    these impacts through smarter control, adaptive infrastructure, and
    low-emission operations. We identify pathways through which the following
    robotics research communities can drive climate-relevant progress:

    \begin{itemize}
        \item \emph{Controls \& Planning}:
              Develop scalable decision-making methods that incorporate
              behavioral feedback, exploit network structure, and integrate
              learning and optimization, with a focus on high-impact domains
              like ports and aviation.

        \item \emph{Estimation}:
              Enable condition-aware maintenance, behavioral demand modeling,
              and real-time estimation pipelines, particularly those that
              directly inform routing, pricing, and scheduling.

        \item \emph{Perception}:
              Scale up deployment of mature technologies like transit signal
              prioritization, and develop robust perception for safety-critical
              domains such as long-range rail obstruction detection and
              infrastructure monitoring during live service.

        \item \emph{Field Robotics}:
              Focus on the development of deployable, reliable, low-maintenance
              systems for areas such as port operations, last-mile delivery,
              and asset maintenance and inspection (e.g., ship hulls or rails).
    \end{itemize}

    These research thrusts can both shift transportation usage to
    lower-emissions modes and build transportation systems that are cleaner,
    smarter, and more resilient, aligning mobility with climate and
    sustainability goals.

\end{boxMarginLeft}

Transportation remains the largest contributor to global greenhouse gas (GHG)
emissions in the United States, accounting for 28\% of all emissions. Within
this sector, road vehicles, including cars, trucks, and buses, dominate, making
up 80\% of the total emissions \cite{epa2018fast}. Air travel, shipping, and
rail transport contribute the remainder, posing unique challenges due to their
scale and operational characteristics. While decarbonizing transportation
requires major efforts in policy, we focus in this paper on how automation can
reduce emissions across all modes of transportation and across both passenger
travel and freight operations.

\titledsubsection{Challenges in transportation}

\begin{figure}[tb!]
    \centering
    \includegraphics[width=\linewidth]{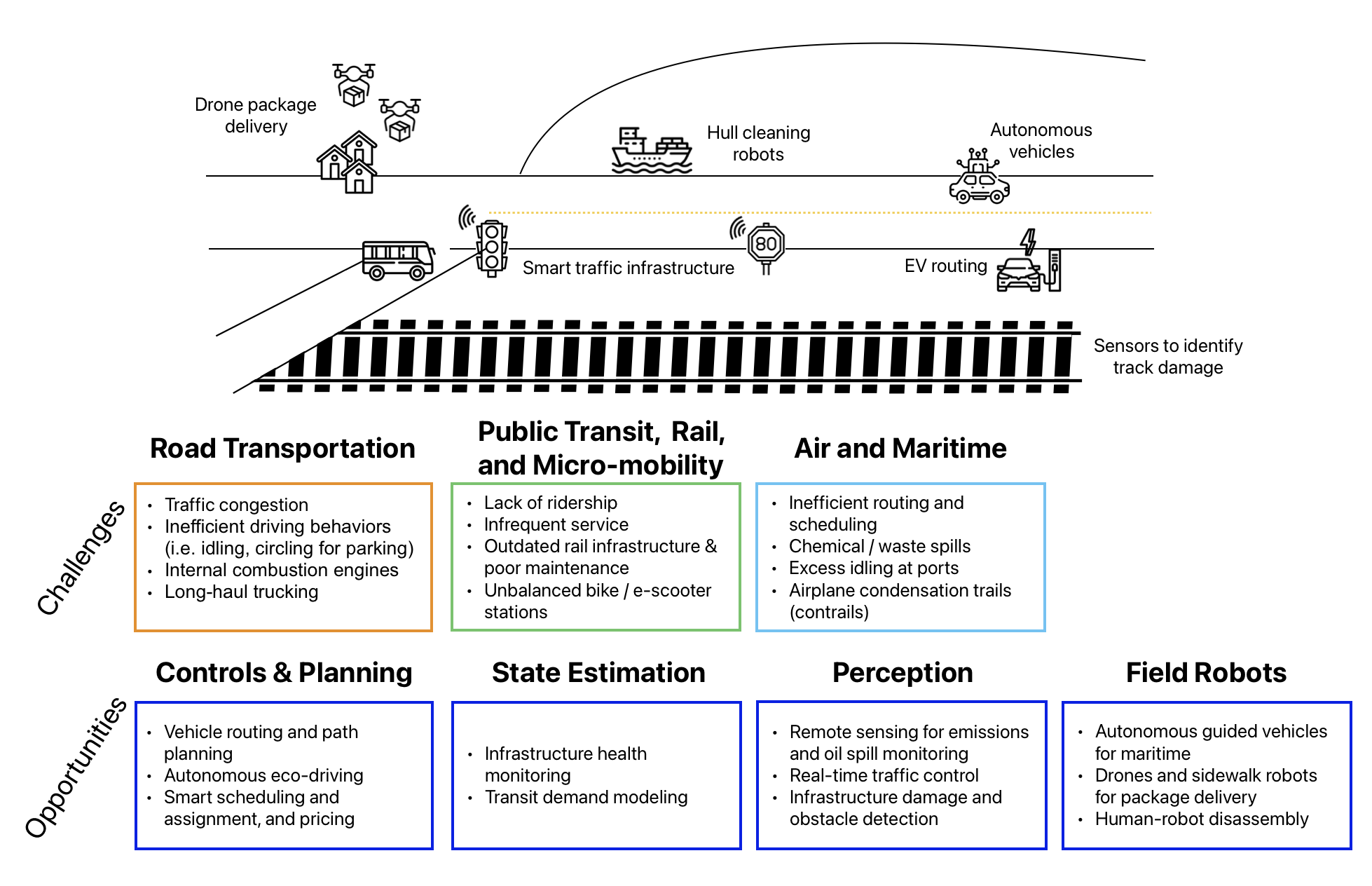}
    \caption{The climate-relevant challenges in transportation systems and
        opportunities for robotics communities to address them.}
    \label{fig:transportation}
\end{figure}

The climate challenges within transportation stem from widespread reliance on
internal combustion engines, inefficient infrastructure, and a lack of adoption
of greener alternatives. Traffic congestion, idling vehicles, long-haul
trucking, shipping, and air travel all burn vast amounts of fuel, which produces
an excess of carbon emissions. Additionally, underutilized public transit
systems and outdated infrastructure compound these issues, preventing an
effective shift to lower-emission modes of travel. Robotics and automation
technologies provide significant opportunities to mitigate the climate impact of
transportation systems and address these challenges. We identify three
categories within the transportation sector:
{Road Transport};
{Rail, Public Transit, \& Micromobility}; and
    {Air \& Maritime}. Each of these categories has a distinct set of challenges, as shown in
\Cref{fig:transportation}. We discuss these challenges in more detail in
\Cref{sec:road,sec:rail,sec:airmar}.

\titledsubsubsection{Road Transport} \label{sec:road}

Road transportation is a major contributor to greenhouse gas emissions, driven
primarily by passenger light-duty vehicles and long-haul trucking. It accounted
for 80\% of the U.S. direct emissions from the transportation sector in 2022
\cite{epa2018fast}. Within personal light-duty vehicles, the largest challenge
is the dependence on internal combustion engine vehicles (ICEVs). ICEVs made up
roughly 79\% of all new light-duty vehicle sales in 2024
\cite{cbtnews2024evmarket}. While the market share of ICEVs is declining, these
engines are still used in a majority of vehicles on the road. They work by
burning fossil fuels, typically gasoline or diesel, and as a by-product, produce
carbon dioxide (\coo) and other pollutants that contribute to climate change and
air pollution. These tailpipe emissions are further increased due to inefficient
driving patterns, such as repetitive accelerating and decelerating, traffic
congestion, and idling in urban areas (i.e. at a traffic signal or in a parking
lot). These behaviors worsen fuel efficiency and exacerbate the climate impacts
of road transportation.

Beyond personal vehicles, the freight industry, particularly long-haul trucking,
is a significant source of transportation emissions. Trucks often travel long
distances, passing through urban centers and consuming large quantities of
diesel fuel. This makes the freight sector disproportionately responsible for
emissions despite representing a smaller share of vehicles on the road. For
instance, medium- and heavy-duty trucks account for 21\% of transportation
emissions in the U.S., despite making up only 5\% of all vehicles on the road
\cite{nrel2024trucks}.

One of the major drivers in reducing emissions from road transport will be
encouraging the use of electric vehicles (EVs) to reduce reliance on interal
combustion engines. The IPCC reports that 85\% of the total emissions (supply
chain and tailpipe) could be reduced when comparing similarly-sized ICEVs and
EVs \cite{IPCC_AR6_WGIII_Chapter10}. Additionally, smart and dynamic city
infrastructure, such as dynamic highway speed limits, traffic signal control,
and connected or autonomous vehicles, provide opportunities to optimize
transportation systems to generate lower emissions. These infrastructure
strategies also help avoid inefficient driving behaviors and excess idling on
roads. While these solutions reduce vehicle-based emissions in the near term,
they may also result in induced demand, which is when more people are encouraged
to drive because of better driving conditions \cite{hymel2019if}. The exact
climate benefit of smart traffic infrastructure has not yet been quantified and
remains an open question. Therefore, the largest known driver of reducing
emissions from driving will be shifting passengers from personal vehicles to
public transit.

\titledsubsubsection{Rail, Public Transit, and Micromobility} \label{sec:rail}

Shifting travel away from personal vehicles and toward public transit, rail, and
micromobility is a critical strategy for reducing emissions from the
transportation sector. Studies indicate that replacing short car trips with
biking can reduce travel emissions by approximately 75\%, while opting for
trains over cars for medium-length distances can cut emissions by around 80\%
\cite{ritchie2023form}. However, these modes face persistent challenges that
limit their adoption. Public transit systems, especially buses and light rail, often struggle with low ridership due to poor accessibility, limited
coverage, and unreliable service. Additionally, in many cities, bus and light
rail vehicles are stuck in the same congested traffic as cars, leading to delays
and inefficiencies and making it less likely for passengers to use these
systems. In fact, some research has shown that commute times for
transit-dependent workers is on average twice as long as workers with car access
\cite{stacy2020unequal}.

Rail, for both passenger and cargo use, faces various maintenance and safety
concerns, especially when it comes to the condition of its infrastructure. Poor
track health and fleet condition (of train cars) reduce operational quality,
which makes rail a less appealing option. Even current inspection requirements
struggle to identify defects because visual inspection is limited and often
fails when it comes to preemptive detection  \cite{schlake2011train}.  Outdated
railway infrastructure conditions can also force trains to go slower or stop
more frequently on the tracks and idle, which results in higher fuel
consumption. For example, single-track lines can cause delays due to meet and
passes (where one train must stop to let the other pass by), and has been shown
to reduce fuel efficiency \cite{fullerton2015sensitivity}. Unfortunately, public
transit and rail systems, especially in the United States, are often outdated
and in subpar condition, so many of our current railway systems face these
problems of reduced operational efficiency and excessive idling. For example,
the American Society of Civil Engineers gave America's rail and public transit
infrastructure a B- and D, respectively, on their annual assessment report card
\cite{asce2025rail}.

Micromobility options like shared bikes and e-scooters offer a low-emission
alternative for short urban trips, particularly for first- and last-mile travel.
A study conducted in Barcelona estimated that Bicing shared bike stations
(similar to Citi Bike in New York City or Bluebikes in Boston) helped reduce the
annual carbon emissions by 9000 metric tons \cite{rojas2011health}. Shared
micromobility stations can also help reduce reliance on cars and ease
congestion, both of which mitigate the emissions generated from driving.
However, while these options are promising low-emission alternatives for urban
dwellers, they face challenges that limit their broader uptake. One major barrier
is unpredictable availability, often caused by uneven distribution of shared
bikes and e-scooters as well as poor fleet rebalancing. This unreliability
reduces user confidence and the likelihood of routine use. In parallel,
infrastructure shortcomings, such as the lack of protected bike lanes, limited
parking or docking stations, and unsafe or poorly lit pathways, create real and
perceived safety risks. Studies have shown that cities with well-integrated
micromobility infrastructure (e.g., bike lanes) and stations with proximity to
metro stations generally experience higher and more consistent ridership
\cite{ghaffar2023meta}. Without deliberate investment in physical infrastructure
and improved operations, micromobility will remain underutilized and struggle to
scale as a meaningful alternative to car-based travel.

Automation and robotics can significantly enhance public transit systems, making
them more efficient, responsive, and appealing to riders. By leveraging
real-time data and intelligent algorithms, transit agencies can better model
rider demand and peak usage hours, optimizing routes, stops, and service
frequencies to align with them. This dynamic adjustment not only reduces wait
times but also improves reliability --- two critical factors in encouraging more
people to shift from personal vehicles to transit. For rail infrastructure,
robotics can enable improved detection and maintenance to prevent delays and
idling. In micromobility, automation can support bike- and scooter-sharing
systems by managing fleet rebalancing and maintenance, ensuring availability
where and when it is needed. Additionally, mobility-on-demand services, such as
autonomous shuttles, can fill service gaps in areas with low fixed-route
coverage. Integration of ride-sharing and car-based mobility with transit
systems, especially for first- and last-mile connections, can create seamless,
multimodal journeys that reduce the need for car ownership and cut overall
transportation emissions.

\titledsubsubsection{Air and Maritime} \label{sec:airmar}

The air and maritime sectors pose unique challenges in the effort to reduce
transportation-related emissions due to their heavy reliance on fossil fuels and
the scale of their operations. Maritime shipping, which handles 90\% of global
trade by volume, runs exclusively on combustion engines, resulting in
substantial greenhouse gas emissions \cite{wef2024shipping,
    lindstad2021reduction}. The emissions of a ship comes from three phases of
operation: cruising, maneuvering, and hotelling \cite{nguyen2022ship}. The
cruising mode refers to when a ship is traveling from port to port, typically at
a constant speed. To reduce emissions from cruising, we can make use of better
routing methods that allow ships to take more fuel efficient paths. On the other
hand, the hotelling and maneuvering phases refer to the behavior of ships in the
port itself. Maneuvering occurs as a ship changes speed and direction to enter
or depart a port, allowing it to navigate among the docks and other ships,
whereas hotelling refers when the ship is moored at port. During the hotelling
phase, the engines are still running to support loading / unloading cargo,
cleaning, and maintenance \cite{toscano2019atmospheric, nguyen2022ship}. A study
of cruise ships in Greece indicated that the hotelling phase generally
contributes to the majority of at-port emissions ($\sim$ 90\%), although this
can vary depending on the type of ship and length of the hotelling and maneuvering
phases \cite{papaefthimiou2016evaluation}. Emissions from the hotelling phase
also negatively impact the health of people living in port area
\cite{ramacher2020impact}. Therefore, there is increased interest from the
academic community and maritime engineers on quantifying and mitigating
emissions from the hotelling phase. Robotic tools like automated guided vehicles
(AGVs), automated mooring systems (AMS), and hull cleaning robots can shorten
the time needed to keep the engine on while docked, making the maintenance and
cargo loading process more efficient.

In addition to carbon emissions, shipping can also cause environmental harm through chemical and waste spills, which threaten marine ecosystems and coastal communities. These incidents may result from equipment failure, human error, or poor monitoring, and their effects are often long-lasting, resulting in harmful effects throughout the entire food chain \cite{sharma2019enabling}. Robotics offers promising solutions to mitigate these risks. Advanced perception systems and real-time environmental monitoring can detect leaks, monitor vessel health, and support faster responses to contamination.

On the other hand, air transportation has its own set of challenges that makes
it difficult to decarbonize. Air travel is the most fuel-intensive mode of
travel, with take-off and landing requiring disproportionate amounts of fuel
compared to the rest of the flight time \cite{c2es_transportation,
    stanford_transportation}. Most aircraft are powered by kerosene (a fossil fuel)
and emit not only large volumes of \coo~but also produce contrails and other
high-altitude pollutants that exacerbate climate impacts
\cite{rmi2023contrails}. However, it is uniquely challenging to design
low-emission alternatives for air travel because aircrafts require an
energy-dense fuel source to achieve appropriate range and speed. For example,
electric and hybrid options are not as energy-dense as
kerosene, making them currently impractical for much of air travel \cite{xue2025sustainable}.
Another promising sustainable fuel option is hydrogen, and while reports show
that it is three times as energy dense as kerosene, it has a much lower
volumetric density, meaning it would require four times as much storage space as
kerosene to fuel the same aircraft \cite{hoff2022implementation}. This would
require redesigning planes to support additional room for fuel. Additionally,
the average lifespan of a
passenger plane is 25 years, and freight planes
typically are replaced once every 32 years, making it hard for these novel sustainable
alternatives to saturate aircraft fleets \cite{bbc_aircraft_retirement}. While
electric, hybrid, and hydrogen-based options are being designed, we need more
efforts to make them feasible and widespread alternatives.  The robotics
community has an opportunity here to leverage its skills to enable low-emission
and low-contrail routing for aircraft, as a means to both mitigate emissions and
reduce the amount of fuel needed.

\titledsubsection{Controls \& Planning} \label{sec:transport:controls-and-planning}

\titledsubsubsection{Routing}
Routing is a foundational component across transportation sectors. More
efficient freight routing (e.g., for trucks and ships) can reduce fuel
consumption and emissions, while optimized routing for passenger services (e.g.,
buses, trains, and mobility-on-demand systems) supports mode shifts away from
private car use. Routing is also critical for electric vehicle (EV) adoption,
where problems like the Electric Vehicle Routing Problem (EVRP) seek paths that
respect charging constraints.

These problems go far beyond computing shortest paths. Real-world
constraints  like required stops, traffic, and weather  must be incorporated.
Some decisions are discrete (e.g., station placement), while others are
continuous (e.g., speed profiles along a trajectory). Routing is often
formulated as a combinatorial optimization problem, typically using
mixed-integer linear programs (MILPs) \cite{braekers2016vehicle}. These problems
are large and riddled with many locally optimal solutions
\cite{braysy2005vehicle}, making them computationally challenging to solve to
global optimality.
As a result, meta-heuristic algorithms are commonly employed to find
high-quality solutions. Techniques include simulated annealing
\cite{lin2021integrating, omidvar2012sustainable}, large neighborhood search
\cite{hao2021electric}, ant colony algorithms \cite{tsou2013ant, sama2016ant,
    li2019improved}, and particle swarm optimization \cite{li2020optimizing}.

Rather than focusing solely on faster heuristics,
the field should prioritize richer models that integrate uncertainty,
multi-objective trade-offs (e.g., emissions vs.\ cost), and real-time
adaptability. Opportunities exist in designing solvers that exploit problem
structure (e.g., spatial or temporal locality, network topology), incorporating
live sensor or traffic data into optimization pipelines, and quantifying
solution robustness under uncertainty. These efforts will improve both the
tractability and trustworthiness of routing strategies in real deployments.

Alternatively, routing problems can be formulated as Markov Decision Processes
(MDPs) \cite{azaron2003dynamic, khani2018real, yue2022estimation, yu2019markov}.
Dynamic programming (DP),  including stochastic and approximate variants,  has been
applied to optimize routes for time or energy efficiency
\cite{cho2012intermodal, ghasempour2020adaptive, papageorgiou2015approximate,
    pourazarm2014optimal}. However, DP requires known state transition dynamics,
which are often unavailable in practice.
To overcome this, reinforcement learning (RL) has been successfully applied to
ship routing \cite{moradi2022marine}, EV routing \cite{lin2021deep,
    basso2022dynamic}, trucking \cite{adi2020interterminal, farahani2021online},
transit network design \cite{darwish2020optimising}, and mobility-on-demand
systems \cite{guo2020deep}.

A key research challenge is to combine the
reliability of optimization with the adaptability of learning. Promising
directions include hybrid approaches that blend offline planning with online RL,
methods for generalizing RL policies across network topologies, and frameworks
that embed physical or operational constraints directly into learning
architectures. Safety, interpretability, and robustness must be emphasized to
ensure real-world deployment viability.

\titledsubsubsection{Vehicle and Infrastructure-Level Controls \& Planning}

Control systems in transportation govern the real-time behavior of vehicles and
infrastructure, shaping both energy efficiency and system-level performance. A
prominent example is autonomous vehicles (AVs), which must operate safely amid
real-world uncertainty and nonlinear, multi-agent dynamics. AVs offer the
potential for significant emissions reductions through `eco-driving' strategies
that minimize idling, smooth traffic flow, and reduce stop-and-go
behavior~\cite{meng2020eco, jayawardana2022learning, jayawardana2024mitigating,
    sun2020optimal}. Notably, studies suggest that even limited AV adoption (as low
as 5--10\% of vehicles) can lead to system-wide benefits~\cite{wu2021flow}.
However, safety and reliability remain key barriers to deployment, motivating
research into supervisory control and cooperative schemes~\cite{hickert2024data,
    hickert2023cooperation}. Encouragingly, reductions in congestion often
correspond to emissions reductions~\cite{beevers2005impact,
    neufville2022potential, wang2015research}.

Beyond the vehicle level, infrastructure-level control strategies like
adaptive traffic signal timing or variable speed limits   can improve throughput
and reduce emissions~\cite{cao2016unified, ferreira2011impact}. Reinforcement
learning (RL) has been used to learn these strategies directly from interaction
data; for example, one field study on a Tennessee freeway showed that
RL-controlled speed limits reduced congestion relative to baseline traffic
patterns~\cite{zhang2024field}.

While such control strategies promise efficiency gains, they may also induce
unintended behavioral shifts. Improved traffic flow could incentivize private
vehicle use over public transit, while driverless AVs might enable longer
commutes~\cite{wadud2016help}. These rebound effects could offset or even
outweigh direct emissions savings.

Autonomous control is also gaining traction in maritime contexts. Emissions from
trans-oceanic shipping can be reduced through optimized steering and autonomous
routing. Additional gains are possible by automating cargo movement and mooring
operations, which can reduce emissions from idling vessels waiting to
dock~\cite{diaz2018reduction, duinkerken2015routing, piris2018reduction,
    tsolakis2022towards, stavrou2017optimizing}. Similar routing-based control
strategies also apply to public bus networks and rail
systems~\cite{bauer2010minimizing, hulagu2020environment,
    miandoab2020developing, moradi2022marine, tsou2013ant}.

In aviation, `eco-flying' strategies use control to reduce contrail
formation, which are  responsible for over a third of aviation's warming impact. By
integrating weather forecasts, satellite imagery, and flight data into advisory
systems, aircraft can be routed to altitudes less prone to contrail
formation~\cite{geraedts2024scalable, teoh2020mitigating}. One recent study
showed that such rerouting can reduce contrail coverage by over 50\% at a modest
2\% increase in fuel use~\cite{elkins2023ai}.

Finally, improvements in low-cost, real-time control technologies have enabled
widespread consumer drone use~\cite{kangunde2021review}. Looking ahead, drone
fleets may play a central role in delivery logistics. While they could reduce
emissions by shortening routes or replacing ground
vehicles~\cite{rodrigues2022drone, stolaroff2018energy}, large-scale deployment
may increase net emissions due to the energy costs of flight and supporting
infrastructure~\cite{goodchild2018delivery, stolaroff2018energy}.

The field should push beyond individual vehicle efficiency to understand and
design for system-wide behavioral responses, especially rebound effects that
may erase emissions reductions. There is a need for control strategies that are
not only energy-aware but also behavior-aware, anticipating how infrastructure
or autonomy changes traveler choices. Maritime and aerial domains have
particular potential for progress: in these contexts, domain-specific
constraints (e.g., port queuing, contrail avoidance) may open opportunities for
high-impact, targeted control. Research at the intersection of learning,
control, and domain modeling will be key to realizing climate benefits.

\titledsubsubsection{Scheduling and Assignment}

Scheduling and assignment are foundational decision problems in transportation
systems, central to optimizing operations and reducing emissions.

\emph{Scheduling} focuses on determining the optimal timing and sequencing of
operations. In transportation, this includes synchronizing public transit with
peak demand to reduce underutilized vehicles~\cite{hassold2014public,
    ghoseiri2004multi}, rebalancing mobility-on-demand fleets to anticipate
surges~\cite{wallar2018vehicle, smith2013rebalancing, spieser2016shared,
    alonso2017demand}, and coordinating electric bus maintenance and charging to
minimize downtime during high-demand periods~\cite{perumal2022electric,
    yao2020optimization}.

\emph{Assignment} refers to allocating resources (e.g., vehicles, drivers, or
infrastructure) to specific tasks, locations, or users. Key objectives include
matching resources to minimize energy use and emissions. Applications span
optimal infrastructure placement (e.g., EV chargers~\cite{liu2012optimal,
    lam2014electric}, bike-share stations~\cite{frade2015bike}, or communications
infrastructure~\cite{tubaishat2009wireless}) and real-time matching problems,
such as dispatching drivers to riders~\cite{alonso2017demand} or allocating EVs
to chargers~\cite{baouche2014efficient}.

Both scheduling and assignment problems are typically formulated as integer or
mixed-integer programs~\cite{wallar2018vehicle, wang2018coordinated,
    pavone2011load, alonso2017demand}, and are often NP-hard. To scale to real-world
settings, researchers have increasingly explored learning-based methods to
approximate optimal policies~\cite{ye2022learning, wen2017rebalancing,
    kim2024learning}. Nonetheless, heuristics remain prevalent for large-scale
deployments~\cite{sassi2017electric, ota2016stars, mounesan2021fleet}.

A promising direction is to develop anticipatory scheduling and assignment
frameworks that fuse short-term predictions with system-level constraints,
enabling proactive decisions and reliability under tight temporal and resource
limitations. Particularly valuable are methods that can generalize across
spatial and temporal regimes (e.g., different cities or event patterns)
supporting scalable, adaptive low-emission transport systems.

\titledsubsubsection{Pricing}

Pricing strategies are powerful levers for shaping transportation system
behavior, with direct implications for climate change
mitigation~\cite{proost2009will}. Tools such as congestion pricing, tolls,
transit fares, mobility-on-demand pricing, and parking fees serve as essential
actuators for aligning transportation use with environmental goals. Strategic
use of these mechanisms can shift user behavior toward more sustainable modes:
higher parking fees can discourage private vehicle
use~\cite{yan2019effectiveness}, incentives for shared mobility can increase
adoption~\cite{storch2021incentive}, and green premiums can reward
eco-driving~\cite{schall2017incentivizing}.

Effective pricing must account for both spatial and temporal variation.
Designing such strategies often involves mathematical
programming~\cite{xie2019optimal}, simulation-based
modeling~\cite{eliasson2001transport, sabounchi2014dynamic}, and game-theoretic
approaches to capture strategic interactions between users and system
operators~\cite{zardini2021game}. In dynamic settings, where responsiveness is
key, data-driven methods are increasingly used to support real-time or adaptive
pricing decisions~\cite{saharan2020dynamic, shukla2020dwara}.

There is a growing need for pricing strategies that account for behavioral
feedback and distributional impacts. Policies should be not only efficient but
also equitable. Promising research avenues include integrating pricing with
other decision layers (e.g., routing or scheduling), developing adaptive schemes
that respond to real-time demand and emissions data, and designing mechanisms
that balance environmental goals with social acceptance and accessibility.

\begin{boxMarginLeft}
    \vspace{1em}
    \subsubsection*{Future Directions: Controls \& Planning in Transportation}
    \begin{itemize}
        \item \textit{Behavior-aware control strategies.}
              Develop control and planning methods that explicitly account for
              system-level behavioral responses, such as rebound effects and
              induced demand, to ensure that efficiency gains lead to real
              emissions reductions.

        \item \textit{Structure-exploiting optimization.}
              Create routing and scheduling algorithms that leverage structural
              properties of transportation systems, such as spatial locality,
              network sparsity, or repeated demand patterns, to enable scalable
              and certifiable decision-making.

        \item \textit{Hybrid learning and optimization.}
              Advance methods that combine offline optimization with online
              learning, particularly for adaptive routing, dynamic pricing, and
              real-time control under uncertainty.

        \item \textit{Targeted control in high-impact domains.}
              Pursue domain-specific control strategies in maritime and aviation
              settings, where constraints like port congestion or contrail
              avoidance present opportunities for outsized climate benefits.

        \item \textit{Integrated and equitable pricing.}
              Design pricing mechanisms that are responsive, predictive, and
              fair. These strategies should be closely coordinated with control,
              scheduling, and user behavior models to shape low-emission
              transportation systems.
    \end{itemize}
\end{boxMarginLeft}

\titledsubsection{Estimation} \label{sec:transport:state-estimation}

\titledsubsubsection{Health Monitoring}

Maintenance is a major contributor to transportation infrastructure costs.
Reducing these costs can help make low-emission transportation modes more
competitive. Robotic tools that estimate vehicle health and predict failures
offer one path to more cost-effective, reliable service.

Electric vehicle (EV) batteries degrade over time and with use. Given the high
cost of battery servicing and replacement, many transportation sectors have been
slow to adopt EVs despite their climate benefits. Zhang et al.\ present a method
for predicting the remaining useful life of lithium-ion batteries using
real-time, non-invasive measurements~\cite{zhang2020degradation}. Incorporating
such models into battery management systems can help owners extend battery
lifetimes, reduce replacement frequency, and limit the risk of unexpected
failure.

Long-lived vehicles, such as trains and heavy-duty trucks, typically rely on
plan-based maintenance that takes them out of service at fixed intervals for
labor-intensive inspections. Robotic tools enable a shift toward condition-based
maintenance, where damage is detected and addressed in real time. This can
reduce labor costs, minimize service disruptions, and improve safety. For
example, Wang et al.\ demonstrate a structural health monitoring system for
detecting damage in high-speed train car bodies~\cite{wang2019trainshm}, while
similar approaches have been applied to train bogies~\cite{hong2014situ} and
other vehicle structures~\cite{okasha2010integration, notay2011wireless,
    hajizadeh2015anomaly}.

By reducing downtime and improving reliability, condition-based maintenance may
allow low-emission transport systems to offer higher service frequencies and
better overall performance.  Promising directions towards this aim lie in the
development of predictive maintenance frameworks that fuse real-time sensing,
machine learning, and domain knowledge (e.g., system models) to anticipate
failures before they occur. Particularly impactful are approaches that enable
low-cost retrofitting of legacy fleets and scalable deployment across diverse
vehicle types, helping accelerate the transition to more reliable,
lower-emission transportation systems and take advantage of the full useful life of
their vehicles.

\titledsubsubsection{Demand Modeling}

Accurately modeling transportation demand is essential for allocating resources
efficiently and delivering reliable service. However, demand is shaped by
external and often unobservable factors, making it difficult to predict. Robotic
systems and data-driven tools can improve responsiveness by enabling real-time
estimation and adaptation.

For high-frequency public transit, \textcite{sanchezmartinez2015realtime}
describes a framework that uses fare collection, passenger counting, and vehicle
location data to dynamically adjust operations and vehicle schedules. In
bicycle-sharing networks, \textcite{liu2021excess} propose a method for
predicting unmet demand to guide more efficient bike and dock allocation. These
approaches improve system reliability and availability, increasing the appeal of
low-emission transport options.

Demand modeling also plays a critical role in long-term planning. Automatic
traffic counts at key road network locations help calibrate demand
models~\cite{cascetta1997calibrating}, allowing agencies to identify underserved
areas. Public transit authorities can analyze smart card data to predict demand
stability~\cite{bass2011model} and assess customer
loyalty~\cite{trepanier2012transit}, informing investment priorities and
evaluating service improvements.

A key opportunity lies in integrating diverse data sources (e.g., mobile
devices, environmental conditions, and social activity patterns) to improve
real-time and long-range demand forecasts. Future models should account for
uncertainty, behavior change, and feedback effects, enabling transportation
systems to not only react to demand, but shape it toward more sustainable
patterns.

\begin{boxMarginLeft}
    \vspace{1em}
    \subsubsection*{Future Directions: Estimation in Transportation}
    \begin{itemize}
        \item \textit{Intelligent maintenance with embedded estimation.}
              Develop systems that fuse onboard sensing, learning, and modeling
              to enable real-time, condition-aware maintenance of transportation
              assets and infrastructure. Emphasis should be placed on scalable,
              retrofittable solutions that reduce downtime and extend vehicle
              lifetimes.

        \item \textit{Demand modeling with behavioral and system feedback.}
              Create demand models that integrate multi-source data with
              insights about behavior change, policy response, and uncertainty.
              These models should enable planners not only to forecast demand
              but to calibrate and anticipate how demand evolves in response to
              interventions.

        \item \textit{Estimation systems that inform action.}
              Design estimation frameworks whose outputs directly support
              real-time decisions in routing, pricing, and scheduling.
    \end{itemize}
\end{boxMarginLeft}

\titledsubsection{Perception} \label{sec:transport:perception}

\titledsubsubsection{Real-Time Traffic Control}

Public transportation vehicles that operate in mixed traffic,  like buses and
light rail,  face challenges in maintaining reliable schedules due to fluctuating
traffic conditions. One effective strategy to improve their speed and on-time
performance is to prioritize them at traffic signals.

\textcite{furth2000buspriority} describe a system that detects buses as they
approach intersections and grants them a green light if needed to stay on
schedule. Crucially, this priority is only given when a delay would cause the
bus to fall behind, minimizing disruption to other traffic. Similar techniques
can be adapted to support a range of low-emission transportation modes,
encouraging greater adoption through improved reliability.

Many of the core components for real-time traffic control already exist. The key
challenge now is system-level integration: coordinating signal priority across
corridors and modes, aligning control with service reliability and emissions
goals, and embedding these tools into urban transportation infrastructure at
scale.

\titledsubsubsection{Damage and Obstruction Monitoring}

Monitoring transportation networks for damage and obstructions is essential for
safety and reliability. However, the high cost and operational disruption of
manual inspections often limits their frequency. Robotic monitoring tools offer
the potential for more continuous, scalable detection of faults before they
compromise service.

While obstacle detection for autonomous cars has advanced rapidly, similar
capabilities for rail remain underdeveloped.
\textcite{risticdurrant2021obstacle} highlight key domain-specific challenges:
trains travel on fixed tracks and require long distances to brake, demanding
obstacle detection systems that operate at significantly greater ranges than in
road vehicles. These systems must also be trained to detect rail-specific
hazards, such as fallen trees or rocks, which are not commonly represented in
road-based datasets.

Rail tracks themselves also require ongoing monitoring for structural damage.
Traditional approaches, such as manual inspections or dedicated track inspection
vehicles, are labor-intensive and often disrupt service.
\textcite{lederman2017track} propose an alternative approach using sensors
mounted on in-service trains. By aggregating repeated measurements over time,
this method enables transit agencies to detect damage trends early and
prioritize targeted inspections in high-risk areas.

A key opportunity in this domain lies in developing long-range, rail-specific
perception systems that can reliably detect rare but high-consequence hazards.
Embedding these systems into real-time operations and maintenance workflows,
especially through integration with in-service vehicles, will be essential for
scalable, low-disruption deployment.

\begin{boxMarginLeft}
    \vspace{1em}
    \subsubsection*{Future Directions: Perception in Transportation}
    \begin{itemize}
        \item \textit{Traffic signal prioritization.}
              Transit signal priority systems can improve reliability and reduce
              emissions by keeping buses and light rail on schedule in mixed
              traffic. The perception technology is mature; the primary need is
              deployment at scale, with system-level coordination and
              integration into multimodal traffic control strategies.

        \item \textit{Rail obstruction and damage detection.}
              Trains require long braking distances and cannot steer around
              obstacles, making early hazard detection and track monitoring
              critical for safety and service continuity. Research opportunities
              include long-range perception for rare obstruction events, and
              scalable, in-service systems for detecting gradual infrastructure
              degradation over time.
    \end{itemize}
\end{boxMarginLeft}

\titledsubsection{Field Robotics} \label{sec:transport:field-robotics}

Field robotics can reduce transportation emissions by automating operationally
intensive tasks that would otherwise rely on fossil-fueled vehicles, engines, or
manual labor. Applications span maritime logistics, last-mile delivery, and
infrastructure maintenance.

\titledsubsubsection{Autonomy in Port Operations}

In maritime logistics, robotic systems can significantly improve the efficiency
of cargo handling and docking. Automated guided vehicles (AGVs) streamline
loading and unloading, reducing ship idle times and associated
emissions~\cite{winnes2015reducing, styhre2017greenhouse,
    stavrou2017optimizing}. Integrating AGVs with AI-driven coordination systems in
``smart ports'' can further increase throughput and reduce
emissions~\cite{johansson2023smart}.

Automated mooring systems (AMS) reduce the need for engine power and manual
handling during docking~\cite{ortegapiris2018reduction, tsiulin2023how}. Because
docking is one of the most emission-intensive phases of ship operation, AMS
technologies represent a high-leverage opportunity for emissions
mitigation~\cite{ortegapiris2018reduction}.

\titledsubsubsection{Autonomous Delivery and Urban Mobility}

Drones offer a potential low-emission alternative to conventional last-mile
delivery vehicles~\cite{stolaroff2018energy, goodchild2018delivery, li2021life,
    rodrigues2022drone}. However, their environmental advantage depends on route
efficiency and task planning~\cite{verge2023google}. Intelligent control
algorithms are needed to ensure that flight energy use does not offset emissions
gains.

On the ground, sidewalk delivery robots can reduce short-trip vehicle use,
helping to alleviate congestion and cut idling emissions in dense urban
areas~\cite{jennings2019study}. While promising, these systems still require
advances in multi-agent coordination, safety, and integration with existing
urban infrastructure.

\titledsubsubsection{Robotic Hull Cleaning and Maintenance}

Robotic maintenance systems can reduce both emissions and operational downtime.
Hull-cleaning robots remove biofouling from ship surfaces, reducing drag and
improving fuel efficiency. Studies show regular robotic cleaning can improve
fuel economy by up to 17\%~\cite{adland2018energy, nassiraei2012development}.

These systems reduce the need for drydocking or diver-based inspection and
repair, offering safer and more cost-effective maintenance. Similar
emissions-aware robotic approaches could be extended to other domains, including
rail and road infrastructure.

\begin{boxMarginLeft}
    \vspace{1em}
    \subsubsection*{Future Directions: Field Robotics in Transportation}
    \begin{itemize}
        \item \textit{Advance system readiness for high-impact domains.}
              In many field robotics applications (e.g., automated mooring, port
              cargo handling, and drone delivery) the core autonomy exists.  The
              research frontier lies in scaling these systems for dependable,
              low-maintenance deployment in unstructured, high-throughput
              environments.

        \item \textit{Integrate with real-world operations.}
              Field robots must work alongside legacy infrastructure and human
              labor. Future efforts should focus on interfaces, fallback modes,
              and integration strategies that minimize disruption while
              maximizing emissions reductions.

        \item \textit{Quantify climate impact in deployment.}
              Robust emissions accounting frameworks are needed to assess when,
              where, and how robotic systems displace emissions. This includes
              considering indirect energy use, marginal emissions savings, and
              the effects of deployment scale and frequency.
    \end{itemize}
\end{boxMarginLeft}

%
%
%
%
%
%

\def\IndustryName{Industry}
\def\IndustryAuthors{Kristen Edwards, Yanwei Wang, and Jingnan Shi}
\section[\IndustryName]{
  \textbf{\IndustryName}%
  \hfill
  \parbox[t]{0.55\textwidth}{\raggedleft\small\itshape \IndustryAuthors}%
 }
\label{sec:industry}

\begin{boxMarginLeft}
    The industrial sector contributed 24\% of global emissions in 2019, much of which come from heavy industrial processes like steelmaking~\cite{IPCC_2022_WGIII_Ch_11}. While decarbonizing heavy industry is primarily a chemical and industrial engineering problem, we identify applications in other areas of the industrial sector where robotics can have a positive impact by \emph{reducing demand for raw materials} and \emph{increasing material efficiency}.
    Towards these two goals, we identify perception, planning and manipulation as the three main interest areas:
    \begin{itemize}
        \item \emph{Perception}: Defects and anomalies detection can be used to prevent
              failures and reduce manufacturing waste. Waste characterization is
              important for increasing material recovery rates.
        \item \emph{Planning \& Manipulation}: Automatic disassembly can help recycle materials from complex products like automobiles, electronics, and batteries. Robotic
              waste sorting can help boost material
              recovery rates.
    \end{itemize}
\end{boxMarginLeft}

\begin{figure}[ht!]
    \centering
    \includegraphics[width=\linewidth]{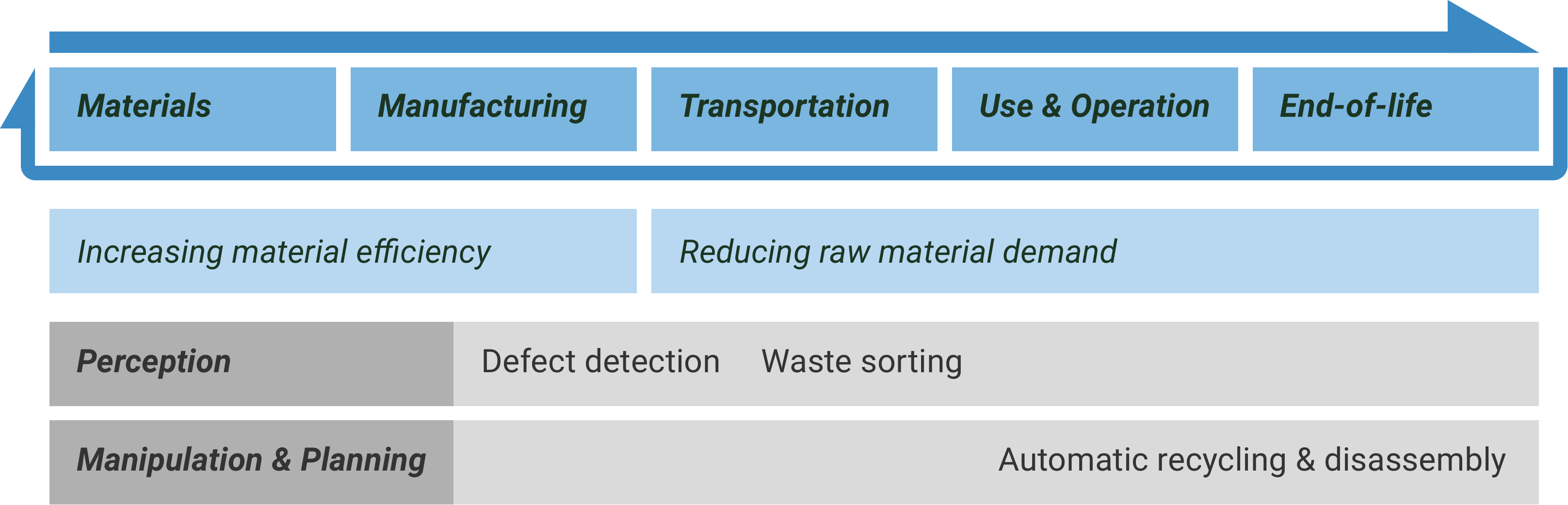}
    \caption{Across the material lifecycle, we identify opportunities for robotics to help reduce waste by reducing demand for raw materials and increasing material efficiency (two climate mitigation strategies identified by the IPCC~\cite{IPCC_2022_WGIII_TS}).
    }
    \label{fig:industry:summary}
\end{figure}


Industrial activities, which include the entire lifecycle from extracting raw materials through manufacturing, use, and end-of-life disposal, are essential for modern human society. These activities support rising global standards of living and drive economic growth and employment. However, industry is also a substantial source of carbon emissions. By some estimates, since 2000 emissions from the industrial sector are growing faster than those for any other sector, reaching 24\% of global emissions in 2019~\cite{IPCC_2022_WGIII_Ch_11}.
There are six commonly recognized mitigation pathways for the industrial sector~\cite{IPCC_2022_WGIII_Ch_11}:
\begin{enumerate}
    \item Reduce material demand
    \item Increase material efficiency
    \item Promote circular economies
    \item Improve energy efficiency
    \item Electrification and fuel switching\label{sec:industry:list:electrification}
    \item Carbon capture, utilization, and storage
\end{enumerate}


While all six are essential to reduce carbon emissions  in the industrial sector, we will focus on the first three strategies, which
are all closely tied to reducing net emissions from material use and present clear opportunities for robotics research to have an impact.\footnote{Some of the other strategies are also discussed in other sections; for example, in \Cref{sec:energy}.} We recognize that a substantial fraction of industrial emissions come from heavy industry (e.g. manufacturing cement, steel, and fertilizer). Decarbonizing the production of these materials is critical for overall climate mitigation, but we do not see a clear pathway for robotics and autonomy research to contribute in those areas; rather, they are primarily areas for chemical and industrial engineering research. Instead, we focus on areas where robotics can help improve efficiency and reduce the need for raw materials, particularly in recycling, waste management, and preventative maintenance.

In this section, we will first use life-cycle analysis to contextualize the
major challenges in the industrial sector and describe how each stage of a
product's life cycle contributes to the sector's GHG emissions
(Section~\ref{sec:industry:sust-manu}).
We then position autonomy research in the context of mitigating the industry
sector's GHG emissions, focusing on perception, planning, and manipulation.
The
main takeaway is that robotics and autonomy research can contribute to
enabling a circular economy (hence reducing material demand) and novel
manufacturing technologies that use less materials (increasing production
efficiency).  We end this section by discussing the broader policy and societal
considerations that affect climate mitigation and adaptation for the industrial sector.

\titledsubsection{Challenges in Sustainable Manufacturing}
\label{sec:industry:sust-manu}
Manufacturing is generally defined as the process of turning raw materials or
parts into finished goods. Within this context, sustainable manufacturing is
performing this process in a way that minimizes environmental impacts while also
being economically and socially
advantageous~\cite{sartal2014sustainablemanufacturing}. Governments~\cite{EPA2021}, international organizations~\cite{UN_2015, SMEP2024}, and major private companies are~\cite{Apple2020,Ford2020} have all identified sustainable manufacturing as a priority.
A common way to
analyze the environmental impact of manufacturing and using a product is through
life-cycle assessment (LCA). LCA examines a product's carbon emissions from
``cradle to grave,'' accounting for every step from raw material extraction to end-of-life disposal and including manufacturing, transportation, and distribution. In the following section, we expand on how
sustainable practices can be incorporated into each of these steps.


\titledsubsubsection{Materials}
Extracting and processing raw materials accounts for a significant portion of a product's energy consumption~\cite{aramendia2023global}.
Increasing demand for materials like steel, cement and plastics has led to a significant increase in industrial GHG emissions in developing countries such as China and India~\cite{krausmann2017-globalMaterialStocks}.
Combined with the already substantial material consumption by wealthy countries, estimated to average ten times more than that of the poorest nations~\cite{Schandl2018-globalMaterialFlows},
materials demand in 2050 is predicted to range from 140 Gt per year to 218 Gt per year~\cite{Krausmann2018-resource}, all well above the potentially sustainable threshold of 100 Gt per year~\cite{Bringezu2015-sustainableMaterialCorridor}.
Therefore, it is critical that we adopt strategies to reduce the demand for raw materials to achieve a sustainable level of industrial GHG emissions.

Several approaches can reduce the demand for materials.
One approach is to improve material efficiency through strategies such as light-weight design or predictive maintenance and repair for longer lifetimes~\cite{IPCC_2022_WGIII_Ch_11}.
Another approach is to reduce the raw material input
through strategies such as circular economy and recycling~\cite{IPCC_2022_WGIII_Ch_11}.
One can also reduce emissions by choosing the \textit{type} of material appropriately. Since the energy consumption at this stage is largely dictated by the embodied energy of a material (i.e., the total energy required for the extraction and processing), choosing different materials can greatly affect the overall sustainability of a product.
\begin{figure}[t!]
    \centering
    \includegraphics[width=\linewidth]{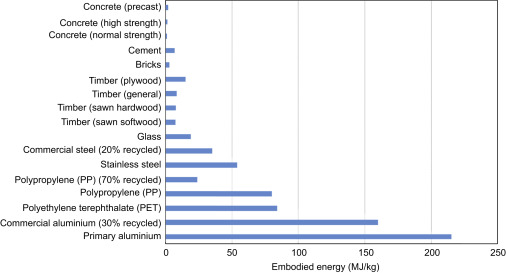}
    \caption{The embodied energy of different common materials. Lower embodied
    energy means less energy is required for extraction and processing. Image
    taken from~\cite{TULADHAR2019441} with permission.}
    \label{fig:ind-embodied-energy}
\end{figure}
Figure~\ref{fig:ind-embodied-energy} shows the embodied energies estimated for
different materials.  Note that the same material can have different embodied
energies based on whether it is completely raw or has been recycled, as 
shown in Figure~\ref{fig:ind-embodied-energy} for steel, aluminum, and
polypropylene.  In
Section~\ref{sec:industry:autonomy:planning-and-manipulation}, we discuss how
robotics can help with recycling to reduce embodied energies.

\titledsubsubsection{Manufacturing processes}
Manufacturing processes are the steps that transform raw materials
into finished products~\cite{klocke2009-manufacturing}.  Typical manufacturing processes
include subtractive manufacturing (e.g., machining), additive manufacturing
(e.g., 3D printing), and sheet metal forming.  The choice of manufacturing
process is dependent on a number of things including the design of the product,
the required function of the product, the total production volume, and
accessibility of materials and manufacturing processes.  One of the many
considerations here is the environmental impact of different manufacturing
processes.
\begin{figure}
    \centering
    \includegraphics[width=0.6\linewidth]{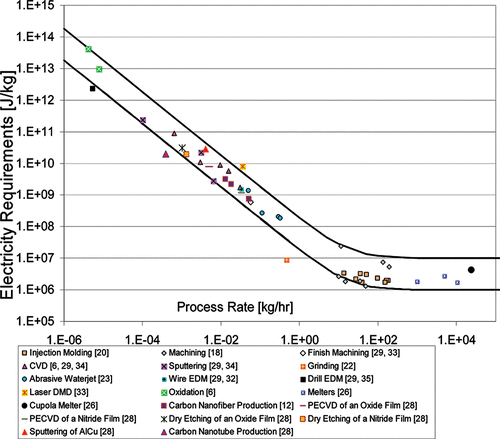}
    \caption{The electricity requirement [J/kg] for different manufacturing processes vs. their process rate [kg/hr]. Image taken from~\cite{gutowski2009thermodynamic} with permission.}
    \label{fig:ind-electricity}
\end{figure}
As seen in Figure~\ref{fig:ind-electricity}, different manufacturing processes have widely varying energy requirements, depending on whether and how much a material needs to be heated, melted, chemically
altered, or physically sheared or bent.

In addition to process' energy use, it is important to consider material efficiency across different
manufacturing process. For instance, studies estimate that approximately 40\% of
all aluminum cast into extrusion billets is scrapped before completion of a
fabricated product.  This inefficiency increases the cost of the fabricated
product by approximately 16\%, while also raising the greenhouse gas emissions
and cumulative energy demand by approximately 40\%
\cite{oberhausen2022-ReducingEnvironmentalImpacts,MILFORD20111185}. Robotics and autonomy can help improve material efficiency in manufacturing by automating certain manufacturing steps to minimize waste and reduce error rates and by detecting defects early in the manufacturing process.

\titledsubsubsection{Transportation and distribution}
The transportation and distribution portion of the LCA includes the energy required to transport a product through different stages of the supply chain as well as distribute the product to warehouses and final retailer locations. Measures such as reducing the total distance traveled and choosing more efficient transportation modes (e.g. shipping rather than air freight) can help improve supply chain sustainability. \Cref{sec:transport} provides much more detail on reducing the climate impacts of transportation systems.

\titledsubsubsection{Use \& operation}
To provide a complete picture, lifecycle analysis must include the impact of using and operating a product. For some goods, such as food and clothing, the energy and carbon contributions from use are negligible compared to those from manufacturing and transportation~\cite{arunan2021-greenhouseFoodPackaging,liu2023-foodGHGoverview}. However, for others products like cars and household appliances the impacts from use can be much more substantial.


Product design can play a role in creating products that minimize the energy
and carbon cost of use or encourage users to operate more sustainably.
A dedicated branch of research focuses on design for sustainable behavior
(DfSB), ranging from the anthropological aspects of sustainable
design~\cite{harper2018aesthetic, leube2024future} to design choices
that encourage sustainable behavior~\cite{chiu2020design, fischer2012human,
    oakley2008motivating}. DfSB aims to reduce a product's environmental and social
impact by moderating how users interact with the product.  While autonomy does
not significantly factor into DfSB, examples of autonomy that adjust user
behaviors to reduce environmental impacts can be found in building environmental
control and HVAC systems (\Cref{sec:built-environment}) and traffic management
(\Cref{sec:transport}).

\titledsubsubsection{End of life}
\label{sec:industry:end-of-life}
The end of life of a product can either contribute negatively or positively to
its LCA. For example, if the product is designed for easy disassembly and is
composed of recyclable materials, it can potentially be recycled or reused,
thereby reducing its total energy consumption and GHG. On the other hand, even
if a product is made of recyclable materials, certain manufacturing processes
could prohibit proper disassembly or effective material separation, making
recycling more challenging and ultimately leading to negative environmental
impacts~\cite{zhu2021-ComingWaveAluminum}. Recycling rates vary by material
type, as seen in \Cref{tab:recovery-rates}, so this early design decision
impacts the end of life sustainability of a product.

The automotive sector provides a clear example of this issue. Cars contain large amounts of
aluminum and steel, which have very high embodied energies as shown in
\Cref{fig:ind-embodied-energy}.  However, current
scrapping techniques often result in significant material losses due to the
difficulty in separating different metal alloys. As demand for aluminum for carmaking is expected to rise  in the coming decades due to a need to save weight~\cite{zhu2021-ComingWaveAluminum}, efficient recycling will be important to minimize potential environmental impacts.
Advances in robotics and autonomy could help disassemble and sort scrap materials more efficiently, potentially increasing recovery rates and reducing energy consumption during
the recycling process \cite{oberhausen2022-ReducingEnvironmentalImpacts}.

\begin{table}[htb]
    \centering
    \begin{tabular}{|l|p{7cm}|}
        \hline
        \textbf{Material}            & \textbf{Recovery Rate (\%)}                                                       \\ \hline
        {Aluminum}                   & \makecell[l]{76\% global recycling efficiency~\cite{international_aluminium_2021} \\ 50\% (cans) to 90\% (industrial)~\cite{aluminum_recycling}}
        \\ \hline
        {Steel}                      & 80\%–98\%~\cite{usgs_iron_steel_scrap}                                            \\ \hline
        {Rare Earth Elements (REEs)} & 60\%–80\%~\cite{fleming2021recovery, YU2024}                                      \\ \hline
        {Plastics (Overall)}         & 8\%–9\%~\cite{oecd2022}                                                           \\ \hline
        {PET and HDPE Plastics}      & 30\%~\cite{boulder2023-impactrecycling, oecd2022}                                 \\ \hline
    \end{tabular}
    \caption{Recovery rates of selected materials}
    \label{tab:recovery-rates}
\end{table}

\titledsubsection{Perception}
\label{sec:industry:autonomy:percep}


Perception is increasingly ubiquitous for robots in manufacturing environments. We identify two specific areas where research in perception can have a positive impact by improving material efficiency through anomaly detection and increasing recycling recovery rates through waste characterization.

\titledsubsubsection{Defect and Anomaly Detection}

Parts must be inspected not only during manufacturing, for quality control, but also during operation to prevent failures and extend lifetimes through preventative maintenance. Perception research aimed at improving automated defect detection could enable scalable and cost-effective solutions for reducing material waste and extending component lifetimes.


Many existing approaches rely on training deep neural networks to classify and detect defects on RGB images~\cite{zou2018deepcrack, bergmann2019mvtec}.
The MVTec Anomaly Dataset~\cite{bergmann2019mvtec} is the de-facto standard dataset for anomaly and defect detection in industrial settings behind these learning-based approaches, covering categories ranging from household items to industrial parts like transistors and screws, complete with defects consistent with those found in real-world manufacturing settings.
Since the release of the dataset, many works have been published and the performance on this dataset has been improved significantly~\cite{cohen2020sub,defard2021padim,roth2022towards,liu2023simplenet}.

Some approaches use depth cameras to reconstruct the 3D shapes of parts to
understand the specific geometries of the
defects~\cite{jahanshahi2013innovative}.  However, for metallic parts commercial
depth sensors suffer from poor accuracy and may not be efficient enough for
fast-paced manufacturing process~\cite{Wang2021-additive3DScan,
    agarwal2023-RoboticDefectInspection}.  To overcome these challenges, there have
been recent advances in combining tactile sensors with vision cameras for
multi-model defect detection~\cite{agarwal2023-RoboticDefectInspection}.
However, currently there are limitations, as modern touch sensors physically
degrade with continued use and the robustness of the method under real-world
conditions remains to be seen~\cite{agarwal2023-RoboticDefectInspection}.

Another area of significant interest is unsupervised defect detection, which aims to detect manufacturing defects without any prior knowledge of what types of defects are present~\cite{bergmann2019mvtec}.
The advantage of unsupervised methods is that they take advantage of pre-training on large-scale image datasets.
Many methods combine deep neural networks with non-learning components such as feature memory banks~\cite{cohen2020sub, roth2022towards} or bags-of-features~\cite{defard2021padim}.
We note that such approaches focus solely on the detection problem without trying to understand the geometry and type of detects. A potential future research direction is to combine unsupervised defects detection with multi-view 3D reconstruction to simultaneously detect and quantify the severity of defects~\cite{agarwal2023-RoboticDefectInspection}.

\titledsubsubsection{Waste Characterization}
\label{sec:industry:percep:waste}
Waste management facilities receive and process waste materials and separate them into different categories.
Some of these materials may be sold to downstream processors or discarded.
They operate complex assembly lines involving different steps to identify and separate various materials to avoid contamination.
Many of such facilities employ workers to manually sort undesired and recyclable materials into different streams.
These facilities are labor-intensive and pose health hazards for the employees~\cite{Li2022-expandPlasticRecycling}.
Currently, automated optical sorters with cameras to identify and separate different materials can replace a portion of manual sorting, but they incur high capital costs, especially for hyperspectral imaging systems that achieve high accuracy~\cite{Li2022-expandPlasticRecycling}. Near-infrared sorters are commonly used for plastic separation, but they do not work well with black plastics~\cite{Xia2021-cnnPlastic}.
Companies have also reported difficulties in obtaining the large amount of training data required to improve their deep learning models~\cite{sanneman2020-StateIndustrialRobotics}.

Significant efforts have been put into improving perception systems in terms of
reliability, robustness and accuracy.
Of particular interest are works on category-level object pose and shape estimation,
where the goal is to estimate the pose and shape of objects for a particular category, without knowing the exact instance~\cite{Wang19-normalizedCoordinate,Feng20-convCategory,Li20-categoryArticulated,Chen20-learnCanonicalShape, Burchfiel19-probabilisticCategory}.
This is contrary to the problem of instance-level object pose estimation, where the goal is to estimate the pose of an object given the exact shape of the object~\cite{Lim13iccv, Chabot17-deepMANTA, Newell16-stackedHourglass}.
Recent works such as OpenShape aim to build large multi-modal networks for general object shape representations~\cite{Liu2024-openshape}.
Such large-scale models have the potential to reduce the barrier of entry for non-expert users who can then incorporate state-of-the-art perception capabilities for waste reduction.

There have been also significant efforts to collect domain-specific datasets for waste characterization.
TACO (Trash Annotations in COntext) is an open image dataset for litter detection and segmentation in the wild~\cite{proenca2020-TACOTrashAnnotations}.
The images are manually annotated based on a hierarchical taxonomy of waste types (28 top-level categories, 60 sub-level categories), and the dataset currently contains 1500 images with annotations and 3918 images without annotations.
The ZeroWaste dataset provides annotated images of highly cluttered scenes of waste items from a single stream recycling facility in Massachusetts, USA, segmenting and classifying objects based on material (cardboard, soft plastic, rigid plastic, and metal) in 10715 labeled images. Future research may seek to provide methods that are robust to distribution shift (e.g. as the composition of the waste stream changes) and easy to adapt to the particular needs of each waste management facility (e.g. through transfer learning or fine-tuning).

Several commercial companies have seen success combining object perception with robotic systems for waste sorting and recycling,
such as EverestLabs~\cite{everestLabs}, Glacier~\cite{glacierRobotics} and AMP Robotics~\cite{ampRobotics}.
To the best of our knowledge, these existing systems rely on 2D detection and classification models to identify the waste type without understanding the geometry of the items.
We believe further material recovery rate improvements may be achieved by incorporating 3D geometry information of the waste items into the characterization process.

\begin{boxMarginLeft}
    \vspace{1em}
    \subsubsection*{Future Directions: Perception in Industry}
    \begin{itemize}
        \item \emph{Monitoring for defects and wear.}
              Perception systems can improve material efficiency in both
              manufacturing and operations by detecting defects and early signs
              of wear.  Research in multi-modal perception (e.g. combining
              visual and tactile sensors) can help improve accuracy, while
              self-supervised approaches can help create scalable systems that
              can be easily deployed in new environments.

        \item \emph{Waste characterization.}
              Automated waste characterization can help increase recycling and
              material recovery rates in waste management facilities.  There is
              room to improve the state-of-the-art in waste characterization by
              incorporating 3D geometry information of the waste items into the
              characterization process and developing multi-modal perception
              approaches that can be easily adapted to new waste streams.

        \item \emph{Domain-specific datasets.}
              Domain-specific datasets that are driven by real-world
              applications will be of substantial value for supporting
              application-inspired research in these domains to improve material
              efficiency, increase recycling rates, and reduce waste.
    \end{itemize}
\end{boxMarginLeft}

\titledsubsection{Planning \& Manipulation}
\label{sec:industry:autonomy:planning-and-manipulation}

Planning and manipulation are core capabilities in industrial robotics,
historically enabling high-throughput automation in structured
settings~\cite{Hagele2016-industrialRobotics}. To contribute meaningfully to
climate change mitigation, these capabilities must extend to more dynamic and
unstructured tasks.
Key opportunities include robotic waste sorting to increase recycling and reduce
landfill use~\cite{glacierRecyclingBot}, and automated disassembly to recover
high-emissions materials and support circular manufacturing
\cite{tanHybridDisassemblyFramework2021}. Unlocking these applications at full
scale will require advances in planning, manipulation, and decision-making under
uncertainty.

\titledsubsubsection{Automatic Disassembly}

Automatic disassembly will be crucial for large-scale recycling.  Existing large-scale metal recycling
techniques like shredding incur significant material losses and struggle to recycle mixed-material items.  While
automated disassembly systems like Apple's Daisy robot exist, they are designed specifically for a
small set of products~\cite{appleDaisy}.  Better metal separation techniques can
drastically improve the recovery rates and reduce energy consumption during the
recycling process~\cite{oberhausen2022-ReducingEnvironmentalImpacts}.
Automated disassembly is also important for lithium-ion battery recycling, as most battery systems are packaged together with integrated cooling and battery management systems~\cite{wegenerDisassemblyElectricVehicle2014}, and automated disassembly can also reduce costs and occupational safety risks relative to manual disassembly~\cite{tanHybridDisassemblyFramework2021}.

To realize this potential, automated disassembly systems must be able to handle complex assemblies and adapt to
different configurations.  While there is a significant amount of prior work on robotic assembly planning, there is relatively little work on automated disassembly.  Fortunately, for
rigid parts with only geometric constraints, assembly
and disassembly planning are nearly identical problems, so many prior works on automated assembly,
should also be applicable to automated disassembly.

Planning for assembly or disassembly requires choosing both a feasible sequence of steps and collision-free paths for each step, making this problem a natural fit for integrated task and motion planning algorithms~\cite{garrett2021integrated, srivastava2014combined, toussaint2015logic,yoshikawa2023large}. When the planning problem also includes sequencing or ordering constraints, temporal logic planning methods~\cite{kress2018synthesis,    belta2007symbolic, wolff2013automaton, plaku2016motion, wang2022temporal} can plan motions that satisfy a temporal logic task
specifications.

The key challenge for automated disassembly is the complexity of the planning problem, which grows exponentially with the number of
parts~\cite{rashid2012-ReviewAssemblySequence}, although some works suggest that the disassembly
sequence may be easier to compute than the assembly
sequence~\cite{ghandi2015-ReviewTaxonomiesAssembly}.  Recent approaches have
demonstrated the effectiveness of physics-based planning in terms of
computational efficiency and capabilities to handle complex parts with various
assembly motions~\cite{tian2022-AssembleThemAll,
    tian2024-ASAPAutomatedSequence}.

Robots can also employ reinforcement learning to learn a policy for automatic
disassembly.  Many recent works have explored training robots in simulation
environments to learn generalizable assembly
policies~\cite{yu2021-RoboAssemblyLearningGeneralizable,
    narang2022-FactoryFastContact}.  Of particular interest is the ability to take
raw pixels directly as inputs to policies, avoiding the need for reliance on
potentially fragile object pose
estimators~\cite{zakka2020-Form2FitLearningShape}.  Another area of interest is
the ability to learn policies that are agnostic to robot embodiments, therefore
allowing for policy reuse on various different robot platforms without
retraining~\cite{luo2021-LearningApproachRobotAgnostic}.  General policies for
automatic disassembly that can be deployed across industries have the potential
to significant increase material recovery rates across different materials.

Several works have explored the use of automatic disassembly for electric
vehicle batteries.  Automatic disassembly of these batteries can enable
large-scale recycling of the expensive metals used in them such as nickel and
cobalt~\cite{beaudet2020-evBattRecycle}, thereby reduce the climate impact of
raw material extraction.  Overall, they report challenges in the nonstandard
design of battery packs with difficult-to-access fasteners, which require
carefully designed disassembly sequences for each model of battery
pack~\cite{wegenerDisassemblyElectricVehicle2014}.  Currently, battery pack
disassembly usually involves humans as main operators, with robotic arms for
unscrewing and handling battery
packs~\cite{wegenerDisassemblyElectricVehicle2014,
    tanHybridDisassemblyFramework2021}. Future research could explore policies that can easily adapt to new battery types and collaborate effectively with human operators.

\titledsubsubsection{Waste Sorting}
Robot manipulators have been used in waste management facilities to sort and
categorize waste types~\cite{glacierRecyclingBot}.  They are typically deployed
in the residual line (also known as the last-chance line), where the potential
recyclables are given one last chance to be sorted out before being sent to a
landfill~\cite{glacierLastChanceRecycle2024}.
Section~\ref{sec:industry:percep:waste} covered the challenges with respect to
the identification and characterization of waste. In this section, we
focus on the manipulation aspect of the recycling process.

Existing commercial robotic recycling systems use vacuum suction grippers
for reliable and fast gripping of waste items with various
shapes~\cite{ampRobotics,glacierRobotics,everestLabs}.  However, suction
grippers suffer from poor performance when gripping items with flexible or
non-planar geometries~\cite{kiyokawa2024-ChallengesFutureRobotic}.  In addition,
suction grippers require constant maintenance and replacement to maintain optimal performance.  Future research directions could
investigate the use of alternative gripper designs (e.g., parallel jaw grippers)
or nonprehensible manipulation~\cite{kiyokawa2024-ChallengesFutureRobotic}, which could complement vacuum grippers for handling waste of different materials and geometries.

In addition to developing improved grippers, improved manipulation policies could help robots handle a larger volume and diversity of waste items. Sorting systems using self-supervised learning could fine-tune a sorting and manipulation policy from real-world data collected in waste management facilities. Recent works
have applied self-supervision for manipulation on applications
like pick-and-place, non-prehensile pushing, grasp generation,
dynamic tossing, and flexible object
manipulation~\cite{pinto2016supersizing, agrawal2016learning, mahler2017dex,pathak2018zero, mousavian20196, zeng2020tossingbot}.
Future work could extend these approaches to waste sorting and recycling, which feature a wide variety of objects and materials, including challenging manipulation problems like picking from clutter and handling deformable objects.

\begin{boxMarginLeft}
    \vspace{1em}
    \subsubsection*{Future Directions: Planning \& Manipulation in Industry}

    \begin{itemize}
        \item \emph{Automated disassembly.}
              Efficient disassembly is a critical enabler for circular manufacturing,
              especially for complex, high-emissions products like EV batteries.
              Future research could explore self-supervised and imitation
              learning approaches for learning disassembly sequences from
              interaction, supported by dedicated data collection platforms and
              real-world evaluation.

        \item \emph{Waste sorting.}
              Robotic waste sorting and recycling can reduce landfill dependence
              and increase recovery of valuable materials. Research is needed to
              expand beyond suction-based grippers, improve manipulation of
              deformable and irregular items, and develop scalable learning
              pipelines using e.g., self-supervised data from high-throughput
              facilities.

        \item \emph{Datasets and benchmarks.}
              Datasets and benchmarks for real-world disassembly tasks, including of
              complex assemblies, and waste handling are needed to support research in
              generalizable and reproducible manipulation systems. This can drive progress and
              scalability in recycling and material reuse.  There would also be
              value in developing a data collection system that standardizes the
              collection of disassembly sequences for various products, as is
              becoming common in manipulation research
              \cite{fu2024mobile,chi2024universal,wang2024dexcap,hagenow2024versatile}.
    \end{itemize}

\end{boxMarginLeft}

\titledsubsection{Conclusion}

Autonomy research can help reduce material demand, improve
material efficiency, and support circular economies --- three key mitigation
pathways identified by the IPCC~\cite{IPCC_2022_WGIII}. We highlight several
climate-relevant applications where robotics and automation can make tangible
impact, including defect detection, waste sorting, automatic disassembly, and
sustainable manufacturing. While some commercial systems exist, there remains
significant room for improvement through advances in perception, learning, and
planning.

It is important to emphasize that autonomy is one piece of a broader
decarbonization puzzle for the industrial sector. Substantial categories of industrial emissions, for example from making steel and cement, require substantial research and developing in industrial and chemical engineering and cannot be reduced through robotics and autonomy alone. Complementary strategies, especially electrification,
fuel switching, and low-emissions feedstocks, are essential to achieve
net-zero goals.

\def\LandUseName{Land Use}
\def\LandUseAuthors{David Russell, Soumya Sudhakar,\\Bilha-Catherine Githinji,
    Joseph DelPreto,\\George Kantor, and Derek Young}
\section[\LandUseName]{
  \textbf{\LandUseName}%
  \hfill
  \parbox[t]{0.55\textwidth}{\raggedleft\small\itshape \LandUseAuthors}%
 }

\label{sec:land-use}
\begin{boxMarginLeft}
    Agriculture, forestry, and other uses of land are responsible for a significant amount of net anthropogenic greenhouse gas emissions, and these are also critical systems threatened by a changing climate. At the same time, there are many opportunities for land use to help with mitigation, for example by remediating degraded land and planting trees (afforestation). We describe relevant research directions in autonomy that address open challenges in this domain including:
    \begin{itemize}
        \item \textit{Controls \& Planning:} Many robotic applications in the
              land use domain involve in-situ measurements for environmental
              monitoring. Open research directions include optimizing tradeoffs
              between in situ sampling and remote sensing, navigating in unstructured
              environments, and 3D simulators for forestry and
              agriculture to accelerate development of planning, control, and RL
              algorithms.
        \item \textit{Estimation \& Perception:} Improved perception and modeling of thin, sparse
              plant structures and improved fine-grained hierarchical classification
              of species can support agricultural and ecosystem monitoring applications.
              New multimodal datasets could highlight relevant features to
              improve performance, while few-shot learning and active learning could
              help reduce the amount of labeled data needed.
        \item \textit{Field Robotics:} Many land use applications require robots
              to serve in harsh unstructured environments and on long-duration remote
              missions.  This raises challenges including robustness, fault tolerance,
              energy efficiency, standardizing modular multimodal sensor payloads,
              serviceability, and retrofitting existing machinery.
        \item \textit{Manipulation:} Tasks such as automated weeding, fruit
              harvesting, pest control, and pruning plants can benefit from
              energy-efficient soft actuators and improved control strategies.
              Advocating for new and existing forestry and agriculture datasets to be
              part of manipulation benchmarks alongside domains that are currently
              widespread such as homes or factories can help spur research progress in
              these areas.
    \end{itemize}
\end{boxMarginLeft}

\begin{figure}[h]
    \centering
    \includegraphics[width=\linewidth]{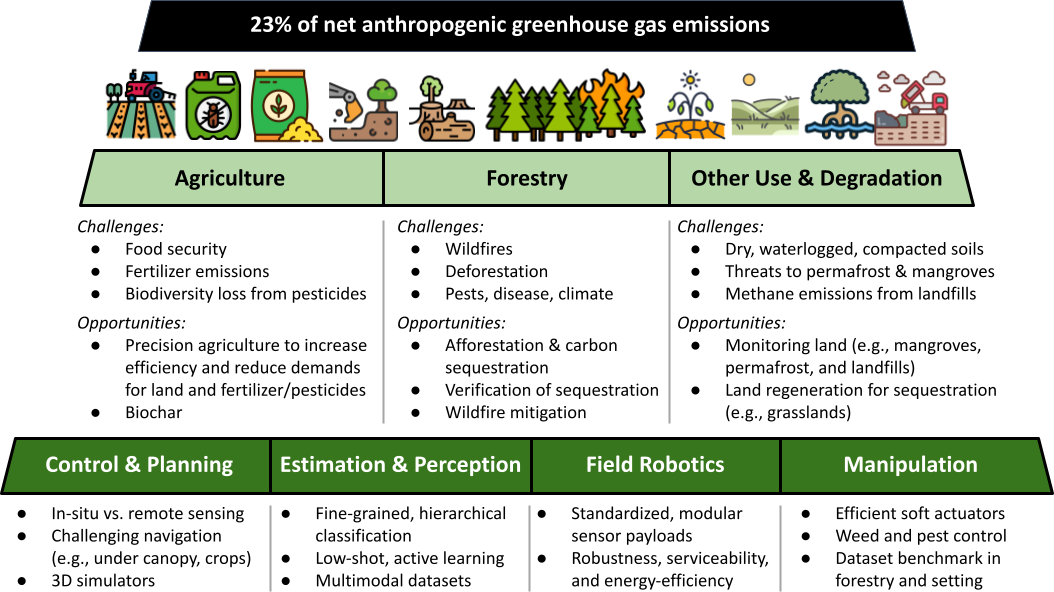}
    \caption{Challenges, opportunities, and future directions for autonomy in the land use sector.}
    \label{fig:land_use_overview}
\end{figure}

In this section, we discuss challenges and solutions for
climate change mitigation and adaptation in the land use sector. Human use of land, including
agriculture, forestry, and land degradation, are responsible for 23\% of net
anthropogenic greenhouse gas emissions and also provide opportunities for carbon
sinks, motivating mitigation measures that could help limit the impacts of
climate change~\cite{IPCC_summary_for_policymakers}. At the same time, this sector
is sensitive to the effects of climate change through increased wildfire risk, habitat
loss, and food insecurity, motivating adaptation measures to increase the resilience
of vital food systems and ecosystems. This section explores how robotics and
autonomy can address technical and implementation gaps for challenges spanning
agriculture (e.g., crop monitoring and automated harvesting), forestry (e.g.,
fighting wildfires, managing wildfire risk, managing forests for timber, carbon
sequestration, and ecological outcomes), and land degradation (e.g., soil
erosion, vegetation loss, and drying or overwetting of lands). Note that
\Cref{sec:built-environment} discusses the use of land for the built environment and
\Cref{sec:earthsciences} discusses the ocean, atmosphere, and cryosphere.

\titledsubsection{Challenges and opportunities in the land use sector}
\titledsubsubsection{Fertilizer and pesticide use in agriculture}
Global agricultural activities within the farm gate (before goods leave the farm) account for 14\% of global emissions~\cite{from2024systems}. Key drivers include deforestation for conversion to agricultural land (which produces \coo{} emissions), synthetic fertilizer production, livestock farming, and rice cultivation, which generate \chhhh{} and \noo{}~\cite{ipcc_ar6_afolu}.
Many agrochemicals (namely, fertilizers and pesticides) rely on manufacturing, distribution, and application processes that directly release greenhouse gases into the atmosphere, soils, and waterways \cite{panna23, Tubiello+2022, Saleh+20, cobb2022herbicides}.
A specific major contributor is the Haber-Bosch process for producing ammonia for synthetic nitrogen fertilizer, which alone accounts for over 1\% of global \coo{} emissions \cite{smith2020current, ZHANG2019100013}.
Precision agriculture enabled by autonomous robots could help make fertilizer
use more targeted and calibrated to the actual need, thereby reducing the total
usage. In addition, solutions to automate the creation and application of
biochar could reduce the dependence on synthetic nitrogen fertilizers and
increase the carbon uptake of agricultural land~\cite{olsson2022land}.
Precision agriculture can also help reduce the application of harmful
pesticides. Though many countries have begun banning unsustainable herbicides
and pesticides, many developing countries remain dependent on hazardous
pesticides since they are widely accessible~\cite{panna15,
    Kudsk+20,Ozkara+15,Delcour+15, Rani+21}. Meanwhile, elevating global
temperatures and genetic mutations in pests make pesticides less effective and
threaten food security~\cite{Matzrafi18, Khan+23}. Farmers lacking
cost-effective technologies
may overapply pesticides and risk harming insect pollinators that are essential
to maintaining crop yields~\cite{Khan+23}. Robotics could help enable new
solutions such as precision applications of pesticides, mechanical weeding, and
early detection of crop disease to help lower demand for pesticides while
protecting food security.

\titledsubsubsection{Food security and climate risks}
The rise of high-throughput agriculture addresses global food security, but
producers face difficult tradeoffs between sustainability and yields.
The current agricultural system is based on innovations
that stem from the Green Revolution in the 1940s-1960s, which made possible
economies of scale and motivated the consolidation of family-owned plots into
industrial farms. However, it also standardized an environmentally unsustainable
agricultural model that drives economies around monocultures, fuel, and chemical
production. This resource-intensive model has, in some respects, weakened
resilience to climate change impacts, but the growing demand for food makes it
infeasible to accept lower production rates during
transitions to more sustainable methods. Population growth is projected to
pressure agricultural sectors, particularly in equatorial regions
\cite[Fig. 2]{Kompas+24},
to increase supply by 35-56\% to meet demand in
2050 \cite{Dijk+21, Tonelli+24}.  At the same time, global warming threatens
productivity of arable lands with increased droughts and other catastrophic
weather events, diminished pollinator populations, and shortened harvest cycles
\cite{epa-cci25, Gornall+10}. Studies conducted with a collection of climate and
crop models predicted at least a 12-15\% reduction in global rice and wheat
production within the next two to four decades \cite{Ahmad+22}.
\par Separately, declining labor supply has incentivized robotic automation in agriculture, but not necessarily with sustainability in mind. Thus far, agricultural robots have facilitated scalable food production through autonomously guided machinery and crop monitoring \cite{Vougioukas19, YaoZhao+24}.
Nevertheless, researchers and engineers have yet to produce robots that operate reliably in unstructured environments. Sustainable agricultural automation will need robots that are adaptive, robust, and economically viable for automated resource-efficient food production.

\titledsubsubsection{Deforestation and declining forest health}
Forests around the world are facing increasing threats including extreme weather driven by climate change, invasive species, catastrophic wildfires, and human-driven deforestation ~\cite{IPCC2019ClimateReport}.
In many regions, increased droughts and heat waves due to climate change directly contribute to forest mortality \cite{allen2015underestimation}, while effects such as sea-level rise and increased tropical storms contribute to mangrove dieback ~\cite{friess2022mangrove}.
Pests and disease may be introduced by human activities, and may also spread incrementally from their historic ranges or increase activity within their historic ranges \cite{bentz2010climate} due to the changing climate. As a result of these threats, forests may face substantial declines in health and productivity \cite{herms2014emerald}.
Finally, forests are threatened by various human activities such as clearing land for agriculture, mining, housing, or harvesting timber and fuel.
Technology can play a key role in monitoring the health of forests and tracking deforestation. This information can be used to drive better management decisions or help drive policy to protect forests from logging.

\titledsubsubsection{Increasing wildfire risk}
Over the past decades, there have been dramatic increases in the frequency and severity of wildfires. In addition to ecological and human costs, these fires release massive amounts of \coo{}; for example, the 2023 Canadian wildfires produced emissions comparable to the annual emissions of a large industrialized nation~\cite{bryne2024carbonCanadianWildfire}. Climate change has led to hotter, drier summers and more frequent winds in many regions, increasing the risk of ignition and rate of fire spread. 
Humans have inadvertently increased these risks in many regions by actively suppressing wildfires, which are a natural and essential part of maintaining healthy vegetation in many ecosystems~\cite{sullivan2022spreading}. As a result, ``megafires'' fueled by dense, unhealthy trees have become more likely and more extreme. They threaten to burn vast areas, often with high severity, killing most of the established vegetation. If only a limited number of seed-bearing trees survive these fires, forests may experience ``type conversion'', changing semi-permanently or permanently to grassland or shrub ecosystems~\cite{guiterman2022vegetation}. 

These challenges present opportunities for robotics to prevent and safely manage forest fires. Preemptive thinning of forests, removing understory vegetation and dead material, and small-scale preemptive prescribed burns~\cite{Fire2021FuelsManagement}, also known as \textit{fuels treatments}, can mitigate the risk of catastrophic fire. Due to high costs, limited funding, difficult access \cite{10.5849/jof.14-058}, and risks of prescribed fires becoming uncontrolled, the scope of deployable fuel treatments is often small~\cite{north2015reform}. Autonomous monitoring systems can be used to survey a large area and prioritize limited resources for fuels treatments to the highest impact areas. Robots can also assist firefighters combating uncontrolled fires by providing situational awareness and directly intervening in the air and on the ground. These applications require sophisticated full-stack robot systems capable of performing in highly unstructured and often dynamic environments.


\titledsubsubsection{Scalable and sustainable forestry operations}
Commercial forestry often has negative connotations, but it can be conducted in a way that is sustainable and preserves ecosystem services and \coo{} sequestration. Agencies and organizations like the Forest Stewardship Council International provide guidance on and certification of sustainable forest management.
Maintaining economically viable commercial forestry can be one way to protect forests from other forms of development \cite{f9090547}. Forestry often involves challenging and dangerous tasks in remote areas, so labor costs and shortages are significant issues \cite{vsporvcic2023deliberations}. Not only must marketable trees be harvested, but new trees also need to be planted and subsequently thinned to ensure that the forest does not become too dense to be healthy \cite{reukema1975guidelines, brodie2024forest}.

Forests store a large amount of carbon worldwide, on the order of 860 billion metric tons according to a 2011 review \cite{pan2011large}. Since clearing land or declining forest health can release this sequestered carbon, there is increasing interest in incentivizing land owners to sustain these carbon stocks; for example, through forest-based carbon credits for behaviors that have a net positive impact on \coo{} sequestration, such as preserving forests, delaying harvest of trees, or planting new trees \cite{vanderGaast02012018}.
Despite the promising idea, ensuring that carbon credits actually lead to increased sequestration is very challenging \cite{https://doi.org/10.1029/2024EF005414}. These challenges include ensuring \textit{additionality}, that payments lead to behavior change beyond the business-as-usual approach, and avoiding \textit{leakage}, where negative effects that would have occurred within the credited site are shifted to another region. While technology alone cannot address these concerns, it can aid in accurately assessing the carbon stocks present in a given landscape.


\titledsubsubsection{Land degradation}
Land degradation caused by humans, either directly or indirectly
affects more than a quarter of the world's land cover~\cite{olsson2022land}. Land degradation can manifest in soils (e.g., physical erosion, compaction, chemical erosion, and salinization), depleted seed resources for regeneration, permafrost thawing, too much water in dry ecosystems and too little water in wet ecosystems (e.g., drying out of peatlands \cite{20093336601}), pest outbreaks, and coastal erosion among others, many of which contribute directly to emissions and have an impact on adaptation. Much attention is paid to soil, since soil holds 1.8 times the amount of carbon in the atmosphere and a decrease in carbon capacity of soil has a potentially large impact on climate.

Autonomous systems can help address human-driven land degradation, such as by providing more comprehensive monitoring solutions to better inform decision-making and planning.  This typically requires multi-scale solutions; for example, remote sensing can paint an overall picture of land degradation and capture vegetation indexes, but in-field measurements are needed for higher-resolution data that capture land health including nutrient depletion, microbial activity, and changes in ecology and biodiversity~\cite{olsson2022land}. By enabling these more comprehensive data fusions and reducing the cost of in-field measurements, robotics can play a role in decision-making algorithms about where to deploy scarce resources.  One challenge that will arise for using this data most effectively is a lack of consensus on defining land degradation, baselines, and standardized methods and sensors~\cite{olsson2022land}, but enabling scalable data sources and deployable solutions can help move towards more informed frameworks.


\titledsubsubsection{Vegetation regrowth and monitoring}

As climate change threatens critical ecosystems like peatlands, mangroves, and grasslands, it is important to develop solutions that can revitalize or restore these resources. Peatlands are exceptionally difficult to monitor without field surveys, since the peat thickness is a critical factor in determining \coo{} sequestration and cannot be determined by remote sensing \cite{20093336601}. This limits monitoring to an extremely small portion of all peatlands, making it challenging to assess ongoing degradation or the success of remediation efforts. Similarly, mangrove restoration requires monitoring vegetation and other environmental factors to assess the success of different restoration attempts~\cite{bosire2008functionality}. Finally,  grasslands collectively are responsible for one third of the terrestrial carbon stock and are used primarily for livestock grazing; their soil organic carbon capacity is most threatened by overgrazing by livestock, though quantitative data is limited~\cite{bai2022grassland}.

Robotics and automation can aid these efforts by providing detailed yet broad-coverage data about vegetation and land use. They can provide high-resolution maps of soil health
using platforms such as UAVs~\cite{ikkala2022unmanned} or ground robots, and they can help gather higher-resolution data across long time scales far more efficiently than manual efforts.  Such information can be critical for monitoring existing interventions and inferring impactful areas for future interventions.

\titledsubsubsection{Monitoring methane emissions}
Estimating and localizing methane emissions can help target mitigation efforts (for human sources like agriculture, landfills, and the fossil fuel industry) and provide valuable data for climate models (for natural sources like wetlands and peatlands).
Advances in remote sensing via MethaneSAT provide new levels of detail for methane measurements but are unable to reliably provide measurements during the night, cloud cover, and over bodies of water~\cite{hamburg2022methanesat}. Robotics can help address some of these gaps. For example, one high-impact source of methane emissions is landfills, estimated to be responsible for up to 20\% of global anthropogenic methane emissions~\cite{cusworth2024quantifying}. There is large uncertainty in measurements of landfill methane emissions, since most data in the U.S. is currently gathered by a person manually walking through the landfill and logs areas of high methane concentration using a methane sensor~\cite{cusworth2024quantifying}. These surveys are difficult, only conducted a few times each year, are dependent on the human operator's choices of where to measure, and can be unsafe.
Using autonomous robots for high-resolution in-situ methane measurements can fill gaps in satellite data to help landfill operators decide where to make repairs or how to tune existing landfill gas capture wells to produce fuel~\cite{xing2024satellite}.

\titledsubsection{Controls \& Planning}
\label{sec:land_use:controls-and-planning}

Autonomy systems for land use applications typically use a
hierarchy of decision-making tools for collision-free path planning, motion
trajectory optimization, and robust control. Further research on compute- and
data-efficient algorithms can improve performance and enhance applicability to
land use scenarios. For example, modern tools can use data-driven methods to
define robust state and dynamics models and to handle scenarios where
underactuated dynamics and unforeseen perturbations are inevitable.

\titledsubsubsection{Motion planning and control for unstructured terrain}

Unstructured terrain like dense vegetation, rocks, roots, and uneven soil
poses a major challenge for field robots. These environmental
features introduce uncertainty in both state estimation and terrain interaction,
which complicates motion planning and control.

In agricultural phenotyping tasks, variable soil and crop conditions disrupt
localization and make high-level navigation difficult~\cite{PanHu+22}. In forest
inventorying, a legged robot used reactive waypoint tracking to generate motion
trajectories that respected feasibility constraints at the control
level~\cite{Mattamala+24}. Motion planners that integrate learned terrain models
have demonstrated the ability to anticipate traversability and improve planning
through cluttered environments, including those with overhanging
vegetation~\cite{Frey+22}. Other approaches incorporate terrain traversability
directly into the cost function during path
planning~\cite{hameed2014intelligent}, while resource-constrained
Next-Best-Action planners have been used for tasks like soil sampling under
uncertainty~\cite{KanThayer+21}.

At the control level, stable locomotion on compliant and cluttered terrain
requires high-frequency, robust feedback strategies. Model predictive control
(MPC) has been deployed for outdoor navigation using data-driven inverse
kinodynamic estimation~\cite{Xiao+21}, and cascaded control schemes have been
used to switch strategies in response to terrain-induced state
uncertainty~\cite{DingHan+22}. Projected inverse dynamics can transform
constrained control problems into unconstrained ones, enabling fast LQR-based
control on soft terrain~\cite{XinXin+21}. Hybrid control methods have also
proven effective in aerial settings, where stability and contact planning must
be balanced in tight spaces~\cite{Aucone+23,Alexis+15}. In agricultural domains,
fuzzy logic controllers offer a rule-based alternative to model-heavy
strategies. These controllers have been applied to fruit picking under
high-variance dynamics~\cite{ChangHuang24} and to speed regulation on rough
terrain using closed-loop heuristics~\cite{Simon23}.

Future work should integrate terrain-aware planning with adaptive, high-rate
control strategies that explicitly handle perception uncertainty and dynamic
environmental transitions. Benchmarking on standardized datasets across diverse
agricultural and forestry settings will be key to driving robust, deployable
autonomy.

\titledsubsubsection{Decision-Making under Uncertainty}

Autonomous systems for land use and environmental monitoring must operate under
significant uncertainty from noisy or partial sensor observations, unknown
dynamics, and evolving environments. These conditions challenge traditional
planning and control approaches that assume full observability and accurate
models.

Learning-based approaches, particularly reinforcement learning (RL), have
emerged as effective tools for decision-making under partial observability. RL
policies trained directly from sensor data have enabled robust navigation in
contact-rich environments, including vineyard row-following with noisy
inputs~\cite{Martini+24}, UAV path planning in dynamic aerial
scenes~\cite{PeiAn+21,TuJuang23}, and forest quadruped locomotion via
sim-to-real imitation~\cite{Miki+22}. \textcite{DoukhiLee22} demonstrated
zero-shot transfer of a deep Q-network policy trained on fused LiDAR and depth
images for micro-aerial vehicle navigation in forested environments. To reduce
data requirements, \textcite{XieMeng+21} accelerated policy training through
prioritized sampling of high-information experiences. In agriculture-specific
applications, CropQL integrated Q-learning with a Random Forest model to
estimate yield and inform resource allocation~\cite{Raj+24}.

Complementing learning-based approaches, rule-based strategies such as fuzzy
logic have also been applied to cope with high variance and poorly modeled
dynamics. In fruit picking tasks, a fuzzy inference system translated noisy
inputs into smooth control commands through fuzzification and
defuzzification~\cite{ChangHuang24}. Similar logic-based controllers have been
deployed for terrain-aware speed regulation in agricultural
robots~\cite{Simon23}.

Finally, planning under uncertainty has also been approached via probabilistic
and information-theoretic techniques. A task allocation system using
Next-Best-Action planning guided soil sampling under resource
constraints~\cite{KanThayer+21}, while risk-aware planners have shown promise
for avoiding collisions and reducing mission failure rates in degraded
conditions~\cite{barbosa2021risk}.

Future work should focus on scalable uncertainty-aware learning frameworks that
integrate multiple sensing modalities, support few-shot adaptation, and allow
reasoning over long horizons. Decision-making strategies must also remain
interpretable and data-efficient to ensure robust field deployment across
diverse and evolving environments.

\titledsubsubsection{Multi-agent planning for environmental monitoring}

When deploying robots for land use monitoring, it is likely that
multiple robots will be used to cover large amounts of land.
Multi-agent planning presents many research challenges including computational
efficiency with number of agents, heterogeneous teams, communication
constraints, and handling exploration and
exploitation~\cite{rizk2019cooperative, dorri2018multi}. Given that robots may
be deployed as a heterogeneous team in hard-to-access areas (e.g.,
insect-inspired robots monitoring mangrove forests), resource management and
failure recovery are key research directions. For example, Mondal \etal\ use a
deep RL approach to manage planning of ground vehicles used to recharge for
multi-agent UAVs surveying land~\cite{mondal2025coordinate}. Meanwhile, many
works consider the robustness of the network to an adversarial attack on a
single agent~\cite{gao2022survey, deng2020robustness}. The efficiency of neural
policies was explored by \cite{HuangYang+24}, who used curriculum learning to
train deep RL agents for fast collision-free swarm navigation on low-compute
quadrotors. Given swarm robotics may have agents that are each extremely
resource-constrained, understanding how to manage multi-agent environmental
monitoring with in-field testing is desirable to understand remaining challenges
to deployment.
Future work should focus on real-world validation of multi-agent coordination
strategies under constraints typical of environmental monitoring: sparse
communication, heterogeneous sensing, and dynamic terrain. Field-deployable
methods must also prioritize robustness to agent failure and adaptive task
allocation in the face of environmental uncertainty.

\titledsubsubsection{Multi-objective planning for field deployment}

Given constraints the robot faces during deployment (e.g., battery or solar
panel capacity, wind, and difficult terrain), there is a deep field of research
investigating how to incorporate these constraints to plan better paths for the
robot to consume less energy overall and extend mission durations. There has
been work considering how to optimize locations of recharging
stations~\cite{Hassan+22, fendji2020cost} and paths~\cite{warsame2020energy} to
reduce recharging time and reduce fuel consumed~\cite{herynek2024multi}. In
addition, for UAVs, Ware \etal\  used computational fluid dynamics to model the
wind field which was then used to model the energy of a trajectory to find
minimum-energy cost paths~\cite{ware2016analysis}. Alyassi \etal\  uses a
combination of a learned regression model to estimate energy cost of a path
accounting for wind, battery and payload, and an approximate planner that
reduces flight time and recharging time~\cite{alyassi2022autonomous}.  \par
Depending on the nature of the planning problem, the computational cost of
traditional motion planning frameworks can become untenable when considering all
constraints faced in the field. Search in large state spaces can be efficiently
carried out by Monte Carlo Tree Search for footstep planning and contact
sequencing~\cite{Amatucci+22, ZhuMeduri+23}.
Marcucci \etal\  further showed how convex relaxation of motion planning
problems can make available powerful convex optimization tools for
collision-free trajectory generation in complex environments \cite{Marcucci+23}.
Understanding how these approaches can scale up to large environments and
non-convex costs and constraints would advance research towards more widely
deployable land-use automation.
Future research should focus on scalable planning frameworks that integrate
diverse field constraints (e.g., energy models, terrain interaction, and
atmospheric dynamics) while remaining computationally tractable. Emphasis should
be placed on developing approximate yet certifiable methods that balance
solution quality with deployment feasibility in large, complex environments.

\titledsubsubsection{Sampling in high-information areas for environmental monitoring}
\label{sec:land_use:planning:IPP}

Many monitoring questions in land use need to be answered at a spatial scale
that vastly exceeds the capacity for exhaustive monitoring. Therefore, results
need to be extrapolated from sparse observations to surrounding areas. Planning
sparse surveys is widely studied in many domains and often represents a
combination of sampling in fixed spatial patterns and stratifying samples across
attributes that are known a priori, such as elevation or forest type. Mobile
robots for environmental monitoring must possess the same type of capabilities
to ensure that their observations are as representative and informative as
possible. Automation can also facilitate updating planned paths online based on
current observations, to focus monitoring efforts on areas that are dynamically
deemed most important.  There has been a rich body of work looking at how to
optimize path planning to maximize information gain or mutual information
exploration methods~\cite{binney2013optimizing, popovic2024learning,
    charrow2015information, zhang2020fsmi}. More recent work has extended
point-based observations to realistic sensor models, such as modeling a
UAV-mounted camera for monitoring weeds in agriculture
~\cite{popovic2020informative}.
%
Future work should prioritize real-time informative planning under realistic
sensing and motion constraints, particularly in settings with limited prior
data. Informative path planning would also benefit from standardized benchmarks.
Current approaches differ widely in goals and evaluation metrics, hindering
direct comparison and slowing progress toward field deployment.

\begin{boxMarginLeft}
    \vspace{1em}
    \subsubsection*{Future Directions: Planning \& Control in Land Use}

    \begin{itemize}
        \item \textit{3D simulators for agriculture, forestry, mangroves, and peatlands.} 
              High-fidelity simulators present an opportunity to test existing and new algorithms that explore unstructured, complex environments in agriculture, forestry, and other land use cases. Developing these tools would require new data collection for these environments with appropriate sensing modalities to provide ``ground-truth'' datasets.
        \item \textit{Navigation in unstructured environments.} Under-canopy, below-crop, and unique environments such as mangroves require new types of planning to account for challenging terrain. Methods that efficiently estimate and resolve uncertainty would better support robots maneuvering around dense vegetation, foliage, and dynamic obstacles.
        \item \textit{Deciding when and where to sample.} 
              An important question is how to plan informative sampling for a target application using existing information, such as remote sensing or domain-specific simulations. Active perception and informative path planning can be applied to this area of research to more effectively deploy robotic solutions.
    \end{itemize}
\end{boxMarginLeft}

\titledsubsection{Estimation \& Perception}
\label{sec:land_use:estimation-and-perception}

Robust perception and estimation systems are a key part of nearly any autonomous
system in land use applications, since the environments are unstructured and the
current states are often largely unknown. These techniques are also widely
applied offline to extract useful insights from data to inform basic science,
land management, or policy.

\titledsubsubsection{Tree detection from aerial data}
Tree-level metrics that are spatially explicit are critical for answering
various forest management and ecology questions. Some work has shown that trees
can be detected in 0.5\,meter resolution satellite remote sensing
\cite{brandt2020unexpectedly}, but this study was conducted in a region with
exceptionally sparse tree cover. In denser forests, most approaches rely on
higher-resolution aerial imagery, often from small uncrewed aerial systems, to
obtain more detailed information than available from satellites.  Multiple
models exist for tree detection in aerial imagery, including the widely-used
DeepForest approach \cite{WeinsteinDeepForest2020}. This is a RetinaNet model
pre-trained on a large set of annotations that were automatically generated from
co-registered LiDAR data, and then it was fine-tuned on manual annotations; it
achieves 0.69 recall and 0.61 precision on a realistic U.S.-wide benchmark.
While DeepForest only produces bounding box predictions, Detectree2
\cite{Ball_2023_Detectree2} uses the MaskRCNN architecture to predict crown
polygons.  There has been recent interest in using foundation models like SAM~2
\cite{ravi2024sam2segmentimages} to generate crown predictions
\cite{teng2025assessingsamtreecrown,chen2025zeroshottreedetectionsegmentation},
because these models show strong ability to perform segmentation with limited or
no domain-specific training. As additional new foundation models are developed,
it will be beneficial to explore which ones are well-suited to tree detection
data.

After trees have been detected, it is common to predict additional attributes
such as species, live/dead status, or carbon content. For many classification
problems, the simplest approach is to fine-tune the detector model to produce
multi-class detections. While this approach may be sufficient for classifying
common distinguishable species, open problems remain around fine-grained
distinctions of similar species. This has motivated work on species
classification from higher spectral resolution data and/or multiple observations
over time \cite{weinstein2023capturing}.  For regression problems such as carbon
content estimation, a separate regression network may be used
\cite{reiersen2024reforestreedatasetestimatingtropical}. These attribution
classification problems pose interesting computer vision challenges, including
domain shift and long-tail distributions. This highlights a need for scaling
the available datasets, and an opportunity to use such domains as compelling
benchmarks for computer vision pipelines \cite{Beery_2022_CVPR}.

Across both tree detection and attribute prediction, there is an opportunity to
use observations from multiple points in time \cite{weinstein2023capturing} or
multiple viewpoints \cite{russell2024classifying} to improve prediction quality.
Multiple approaches have been explored to merge structural information, which
can be provided from either LiDAR or photogrammetry, with visual information
\cite{XU2025128696, rs12152379, teng2025assessingsamtreecrown}. Continuing to
explore approaches for fusing multiple data points or many sources of data while
maintaining computational efficiency is an area for future research.


\titledsubsubsection{Under-Canopy Mapping}

Aerial imagery is highly scalable, but cannot directly capture important
characteristics of under-canopy forest dynamics, such as tree diameters or
forest fire fuel densities. Under canopy, terrestrial LiDAR scanners (TLS),
i.e., stand-mounted spinning LiDAR, are commonly used because they capture dense
and highly accurate point clouds
\cite{liang2016terrestrial,POKSWINSKI2021101484}. Mobile LiDAR units (e.g.,
hand-held or backpack) have also been explored \cite{hyyppa_comparison_2020} and
make data collection more convenient at the expense of decreased quality.  To
obtain online estimates of forest structure, for purposes such as informing
robotic navigation, simultaneous localization and mapping (SLAM) algorithms are
required. These approaches often require multiple complementary sensors and are
sensitive to abrupt motion \cite{garforth_visual_2019}. One innovation is to use
tree trunks as additional landmarks in the SLAM optimization
\cite{chen_sloam_2019} to improve robustness and precision. This approach
produces a spatially-explicit map of tree trunks and additionally predicts the
diameter of each tree from point clouds as a post-processing step.  Key
directions for future work include incorporating learning-based tree
segmentation, techniques specifically developed for highly cluttered forests
which are commonly encountered, and the development of standardized, annotated
datasets for benchmarking forest mapping algorithms.

\titledsubsubsection{Under-canopy semantic understanding}
There has been impressive progress on semantic tasks using point clouds in the
last few years, such as individual tree segmentation and point classification.
Early successes in high-quality tree segmentation
\cite{TAO201566,https://doi.org/10.1111/2041-210X.13121} used multistep
algorithms and hand-crafted features with extremely high-quality point clouds.
More recent approaches rely on deep learning
\cite{xiang2025forestformer3dunifiedframeworkendtoend}. The work of Wielgosz
\etal\ \cite{WIELGOSZ2024114367} is notable for testing on downsampled
pointclouds to explore performance on data collected by lower cost or higher
coverage modalities such as aerial LiDAR or photogrammetry, and this area should
be further studied.

Online robot systems for forestry generally rely on multiple complimentary
sensors, and visual sensors are often used to infer rich semantic information
\cite{robotics12050139}.  Imagery-based perception is often the only feasible
option for classifying vegetation type \cite{andrada2020testing} or classifying
tree species \cite{Beery_2022_CVPR}, and can be a strong alternative to LiDAR
for other tasks such as detecting tree trunks \cite{10.1093/forestry/cpac043}.
However, downstream tasks generally require these 2D predictions to be
efficiently integrated into a constantly-updating 3D representation. This task,
often termed \textit{semantic mapping}, has been commonly studied in indoor and
autonomous driving settings. In these cases, the environment consists largely of
opaque objects with flat surfaces, and therefore mesh-based algorithms are
common \cite{9196885}. One alternative approach is to use a volumetric approach,
such as a semantic OctoMap \cite{hornung13auro} for mapping forest fire fuels
\cite{russell2022uav, andrada_mapping_2023}. Future work should explore what
representations are best suited to under-canopy semantic mapping. Currently, one
of the only multi-sensor under-canopy datasets is TreeScope
\cite{cheng2023treescope}. Additional datasets should also be developed, with a
focus on image-based prediction tasks.

\titledsubsubsection{High-throughput phenotyping via computer vision and 3D reconstruction}
Plant scientists and breeders can use robotic-assisted phenotyping
\cite{li_review_2020, Atefi21} to identify candidate crops that are resilient to
pests, disease, drought, and other environmental stressors. In the face of
climate change, this innovation must happen faster than ever to adapt to new
challenges. Lab-based image analysis pipelines have recently been used to
efficiently characterize plant root architecture and growth dynamics. Most
techniques extract features from vision-only sensing (e.g., RGB, thermal, LiDAR,
and spectral imaging), though alternative modalities like depth and contact
sensors have been used \cite{Atefi21,Ruppel+23,Chaudhury+17}. Phenotypic
analysis of row crops, for example, can be conducted using ground robots and
single-camera drones collecting terabytes of high-resolution image data
\cite{Esser+23, LiXu22, Nelson+23}.

Reconstructing plants presents a number of challenges including thin structures,
frequent self-occlusions, limited texture, and potential deformation between
views due to wind or vibration \cite{kochi_introduction_2021}. These challenges
have motivated the use of various techniques such as neural reconstruction
methods \cite{hu_high-fidelity_2024}, optimizations for thin objects
\cite{rs16244720}, or explicit template matching \cite{marks_precise_2022}. In
the wild, \cite{PanHu+22} demonstrate in situ phenotyping via a mobile robot
that autonomously navigates orchard and greenhouse environments and renders 3D
maps of crops using a NeRF model. Their proposed system detects plants with
reasonable precision, averaging 6\,cm position error and 2\,cm bounding box
error, suitable for highly localized analysis.

Future work should focus on robust, scalable reconstruction and analysis methods
that handle plant-specific challenges such as fine geometry, deformation, and
occlusion.  There is also a need for benchmark datasets and evaluation protocols
tailored to real-world agricultural conditions, enabling comparison across
sensing modalities and reconstruction techniques for diverse crop types.

\titledsubsubsection{Object Detection and Classification for Precision
    Agriculture} Robotics presents an opportunity for precise and efficient
production agriculture. While precision agriculture shares some commonalities
with high-throughput phenotyping, it focuses less on detailed analysis of
individual plants and more on reliably automating tasks such as navigating
fields, harvesting fruit, or automating weeding. These innovations could
increase food production while limiting the use of agro-chemicals. Object
detection and classification tasks are common in land use applications and are
largely tackled using imagery data and deep learning \cite{kamilaris_deep_2018}.
In agriculture, many tasks involve detecting individual plants
\cite{BARRETO2021106493} or specific portions of plants, such as fruits that
must be harvested \cite{wosner_object_2021}.  Classification tasks are often
between crops and weeds \cite{sa_weedmap_2018}, diseased vs. healthy plants
\cite{JACKULIN2022100441}, or growth or ripeness stages \cite{HAREL2020103274}.

\par It can be difficult to analyze differences between types of vegetation
using only visual spectrum features \cite{JorgeSanchez+19}, as they are less
reliable in variable lighting conditions and the presence of occlusions.
Customized sensors for plant volatility have been proposed \cite{Geckeler+24} as
part of a stress identification approach, but they require meticulous placement
and retrieval for sampling. Some of these limitations motivate the use of
machine learning approaches for disease detection and classification in fruit
\cite{Khan+23, ArifinRupa+24}. In 3D reconstruction problems, active perception
\cite{Burusa+24} can address occlusions and learned keypoints can increase
robustness of 3D plant structure estimation \cite{KiWang+23} for fruit
harvesting. Recently, data fusion techniques have been applied to the problem of
estimating vegetation indices from both spectral and spatial modalities, albeit
requiring excessive computation and network bandwidth \cite{Omia+23,WangChen+24}
and potentially increasing the barrier to adoption.

Future research should aim to develop robust detection and classification
systems that generalize across lighting, occlusion, and crop variability. Key
directions include integrating multi-modal sensing with lightweight models
suitable for real-time deployment.




\titledsubsubsection{SLAM for Ground-Based Crop Monitoring}

Long-term crop monitoring requires spatially and temporally consistent
localization to track changes in plant structure, health, or yield over time. In
structured agricultural environments, which are often GPS-denied due to canopy
cover, simultaneous localization and mapping (SLAM) provides the necessary
spatial reference frame to align observations across repeated passes and
seasons. This enables high-resolution, map-based tracking of individual plants,
growth trends, and treatment effects, which reactive navigation methods alone
cannot support.

Recent efforts in agricultural SLAM have leveraged crop-specific structure like row geometry or tree trunks to improve robustness and data
association~\cite{kim_p-agslam_2024,auat_cheein_optimized_2011}. These
approaches go beyond local row-following~\cite{Silva+23,LiXu21,Silva+24,Xue+12}
by maintaining persistent, globally consistent maps over time. However,
challenges remain in managing partial occlusions, structural deformations due to
wind or growth, and sensor viewpoint variability across missions.

Synthetic data generation has recently emerged as a tool to support
learning-based perception and localization in crop environments, helping to
scale development and simulation of field-deployable SLAM
pipelines~\cite{Martini+24}. Still, real-world deployment needs models that
generalize across crop types, seasons, and sensing conditions.

Future work should prioritize crop-aware SLAM frameworks that support long-term
temporal consistency, semantic tracking of plant instances, and integration of
spatial priors from field structure. Rich, timeseries datasets
will aid in developing robust, scalable solutions for longitudinal monitoring in
agriculture.

\titledsubsubsection{Advancing Data-Driven Remote Sensing}

Large-scale land use monitoring increasingly relies on remote sensing data to
guide, validate, or scale up robotic interventions. Satellite imagery and
spectral data offer minimally invasive means of characterizing terrain,
vegetation, and crop conditions at broad spatial and temporal
scales~\cite{ROGAN2004301, rollins2009landfire, global_forest_watch}. These
datasets are widely accessible through programs like
Sentinel~\cite{sentinel1_2012} and Landsat~\cite{RoyLandsat8_2014} and often
undergo preprocessing to enable pixel-wise analysis across time and
space~\cite{young_survival_2017}.

Traditional land use classification approaches have relied on spectral indices
and shallow machine learning models~\cite{KHATAMI201689}, but current research
is driven by deep learning~\cite{9553499}, with growing emphasis on more
granular, context-specific labels; e.g., crop type or harvest
stage~\cite{tseng2021cropharvest}. Foundation
models~\cite{lacoste_geo-bench_2023, khanna_diffusionsat_2024} and feature
learning strategies like MOSAIKS~\cite{rolf2021generalizable} are being
developed to reduce supervision requirements and improve transferability. Other
work focuses on task- and region-adaptive learning using meta-learning
frameworks~\cite{tseng2022timltaskinformedmetalearningagriculture}.

Despite progress, key gaps remain. Ground-truth datasets are globally
imbalanced, limiting model reliability in data-poor regions. Computational
efficiency is increasingly important given higher-resolution inputs and revisit
rates. Few systems close the loop between satellite-derived
insights and robotic action on the ground.

Future work should explore tight coupling between remote sensing and robotic
decision-making; for example, using remote observations to prioritize high-value
sampling locations, guide multi-robot deployments, or aid on-the-ground
estimates. Building this connection will require not only better models, but
also workflows and datasets that span both orbit and field scale.

\begin{boxMarginLeft}
    \subsubsection*{Future Directions: Estimation \& Perception in Land Use}
    \begin{itemize}
        \item \textit{Geometrically accurate plant modeling.} Robust 3D
              reconstruction of thin, deformable, and self-occluding structures
              remains a major obstacle to reliable plant state estimation. Advances in
              high-resolution, low-artifact depth sensing and learning-based methods
              for uncertainty-aware modeling from camera and LiDAR data are critical
              for downstream applications such as phenotyping, yield estimation, and
              manipulation.

        \item \textit{Hierarchical and fine-grained classification.} Most
              existing approaches focus on coarse semantic labels (e.g., crop vs.
              weed), but land use tasks often require species-level or
              condition-specific identification. Developing benchmarks that reflect
              intra-class variation and taxonomic hierarchies, and curating synthetic
              data to fill known gaps, will support more precise and biologically
              meaningful models.

        \item \textit{Learning from limited supervision.} Many perception
              problems are bottlenecked by expensive expert annotation. Few-shot,
              self-supervised, and active learning approaches, particularly those that
              leverage multimodal cues (e.g., RGB + depth + spectra), can enable
              scalable training from sparse or weakly labeled data. Simulation-driven
              pretraining and in-the-loop fine-tuning will also be key to deployment.

        \item \textit{Semantic mapping and temporal consistency.} Long-term
              monitoring of land use requires integrating perception outputs into
              spatially and temporally coherent maps. Research is needed on semantic
              SLAM, multi-pass alignment under structural change, and maintaining
              consistent instance tracking over time in the presence of
              occlusion, growth, or seasonal variation.

        \item \textit{Multimodal and cross-domain generalization.} Field robots
              and remote sensors collect heterogeneous data under varying conditions.
              Perception systems should be robust to changes in sensor type, crop
              type, geography, and lighting. Developing shared benchmarks, model
              adaptation strategies, and domain-aware training pipelines will be
              central to generalizing across diverse environments.
    \end{itemize}
\end{boxMarginLeft}

\titledsubsection{Field Robotics}
\label{sec:land_use:field-robotics}

\titledsubsubsection{Sensor Suites for Environmental Monitoring}

Understanding landscape characteristics is critical for guiding interventions,
informing policy, and supporting scientific research. While many sensing outputs
are analyzed through perception techniques (see
Section~\ref{sec:land_use:estimation-and-perception}), this section focuses on
hardware and data acquisition strategies for environmental monitoring.

Small uncrewed aerial systems (sUAS) are increasingly used in land management
due to their low cost, ease of deployment, and ability to rapidly survey large
areas~\cite{farrell2023eyes, tiberiu_paul_banu_use_2016, kim_unmanned_2019}.
Most are equipped with RGB or multispectral cameras, though directional LiDAR is
also common~\cite{hyyppa_review_2008}. sUAS platforms typically follow GPS-based
preplanned flights using commercial or open-source tools to ensure even
coverage. These systems are ideal for static environments with known areas of
interest, where flights can be conducted at sufficient altitude to avoid
obstacles.

For tasks requiring selective inspection or online navigation, more advanced
payloads are needed. Davila~\etal\ developed an open-source system integrating a
camera, GPS-IMU, onboard compute, and wireless
downlink~\cite{davila2022adaptopensourcesuaspayload}. This system performs
real-time semantic segmentation, georeferences predictions, and streams them to
the operator, allowing interactive monitoring. It was validated in a river ice
survey with human-controlled flight beyond line of sight.

Monitoring under forest canopies is essential for assessing wildfire fuels and
carbon stocks. UAVs with custom payloads have been used for fuel
mapping~\cite{russell2022uav, andrada_mapping_2023}, often built around DJI M600
drones equipped with stereo cameras, LiDAR, and onboard compute. These were
manually piloted within line of sight. Autonomous UAV deployment in cluttered
environments is more challenging, requiring onboard sensing and control to avoid
obstacles. Examples include systems with active stereo cameras and
IMUs~\cite{karjalainen_autonomous_2024}, or 360\,degree LiDAR and stereo
vision~\cite{DBLP:journals/corr/abs-2109-06479}, both capable of estimating tree
locations and diameters.

Due to the complexity of autonomous flight in dense vegetation, wearable sensor
backpacks, typically combining LiDAR and IMUs, have gained popularity for forest
surveys~\cite{xie_applying_2022, su_development_2021, ko_comparison_2021,
hyyppa_comparison_2020}.

Future research should broadly explore sensor suite strategies tailored to the
unique challenges of environmental monitoring, such as dense vegetation, uneven
terrain, and variable lighting. Designing systems that balance capability, cost,
and operational constraints will greatly enhance the ability to collect data for
land use applications.




\titledsubsubsection{UAVs for active wildfire response}
To adapt to a changing climate, we need better methods for containing
increasingly frequent and high-intensity wildfires to mitigate loss of human
life, ecosystems, and property in impacted communities. There are also numerous
chronic and acute risks to fire fighters including smoke inhalation and falling
trees such that increased automation could improve firefighter safety. Crewed
aircraft have been an integral part of wildfire response for decades by
providing situational awareness and applying fire suppressants. This has led to
extreme caution about using UAS, because even small commodity drones can
severely damage crewed aircraft. However, there is increasing interest among
fire response agencies in using remotely-piloted UASs as a supplement or
alternative to crewed aircraft for situational awareness
\cite{noauthor_mechanized_2022}.  One example application of autonomous flight
is using a multi-sensor UAV for monitoring wildfires \cite{jong_wit-uas_2023},
which allows the UAV to track individual fire fighters to monitor their safety.
Working near wildfires presents interesting challenges for robotic design,
including degraded perception from smoke, unpredictable air movement, and
requirements for highly reliable operation.
Future work should explore UAV designs and autonomy strategies that are robust
to smoke, heat, and turbulent conditions, while ensuring safe integration
alongside crewed aircraft during active wildfire response.

\titledsubsubsection{Under-canopy localization}
Precise geospatial localization is a critical component of many land use tasks.
Predictions from automated systems often need to be registered to other data
products such as manual surveys or prior observations to conduct analysis.
Precise localization is also important for navigation, especially if multiple
robots are coordinating on a shared task or robots are navigating a delicate
environment such as a crop field.

GPS data is the most common way to perform coarse localization, but it often
suffers meter-scale errors, which is often exacerbated in challenging
environments such as under forest canopies or mountainous terrain. Real-time
Kinematic (RTK) or Post-Processing Kinematic (PPK) GPS systems use a stationary
GPS ``base station'' to correct for atmospheric effects, providing localization
accuracy on the order of centimeters relative to the base station
\cite{tomastik_uav_2019}. If the base station is deployed in a fixed location,
data collected across multiple days can be effectively registered together.
Publicly available corrections from a network of fixed GPS receivers can also be
used, but provide less accurate corrections than an on-site base station.
Tomastik \etal\  \cite{tomastik_static_2023} achieved sub-meter localization
accuracy for a static receiver under a forest canopy using corrections from the
NOAA Continuously Operating Reference Stations (CORS) network
\cite{CORS2008Snay}.

Future work should develop localization methods that leverage specific
structures of under-canopy settings to remain accurate and reliable in
GPS-degraded environments, with a focus on long-term repeatability, multi-day
registration, and integration with other sensing modalities.



\titledsubsubsection{Retrofitting existing machinery for forestry and vegetation management}
Commercial forestry is a major industry worldwide, and there is also an
increasing understanding that forests must be proactively managed to mitigate
the risks and impacts of catastrophic forest fires. In both cases, there are
specialized industrial machines for these tasks that have been refined for
decades. However, in many cases, a major limitation is finding skilled operators
of these machines, especially given the demanding, dirty, and dangerous nature
of this work and shrinking rural workforce \cite{la_hera_exploring_2024}.

One solution is to retrofit existing machinery to be teleoperated, which allows
operators to have a safer and more comfortable work environment and avoid
traveling every day to often-remote job sites \cite{visser_automation_2021}.
This strategy is employed by the company Kodama \cite{Kodama2024} which is using
retrofitted teleoperated machinery to thin forests.  Retrofitting existing
machinery usually requires adding additional sensors and dedicated compute, but
often retains the original actuators, sensors such as encoders, and
communication protocol (e.g., Controller Area Network (CAN)). Interfacing with
these existing capabilities can be challenging because they are usually not
designed to be exposed to the user and may require reverse engineering.

Alternatively, there has been work that aims to develop fully-autonomous systems
using a retrofitted platform. Multiple projects in commercial forestry have
retrofitted specialized forest equipment such as a log forwarder
\cite{la_hera_exploring_2024} and a small tree harvester
\cite{jelavic_harveri_nodate}.  Another application is vegetation removal, and
one project retrofitted a medium duty Bobcat T190 with a grinding attachment for
semi-autonomous clearing \cite{ Portugal2021, andrada_integration_2022}.  While
these systems are impressive proofs of concept, especially for the engineering
work required to retrofit existing hardware, autonomous operation of these
systems is extremely preliminary and demonstrations are conducted in simple
scenarios.

There is a need for more work on retrofitting existing machinery (for both
teleoperation and full autonomy), with attention to tackling the real-world
challenges of cluttered environments, complex terrain, the need for complex and
open-ended manipulation, and reliable operation in remote locations.


\titledsubsubsection{Robot design for navigating challenging terrain}

Forests and other wildlands present a variety of navigation challenges not seen
in the built environment or managed ecosystems, such as highly cluttered
obstacles, steep slopes, and unknown terrain dynamics. This makes many regions
inaccessible for monitoring or interventions.  Taking cues from manually
operated machinery, many approaches employ wheeled robots with large wheels or
treads and a low center of gravity. In these cases, using proprioceptive sensing
(e.g., wheel speed and motor current) can be helpful for robust control across
diverse terrains \cite{LaRocque_2024}.  An emerging alternative is using legged
robots, which was shown by Lee \etal\ \cite{doi:10.1126/scirobotics.abc5986} to
offer robust performance across various terrain. While technically complex,
these legged systems benefit from years of study and increasing commercial
support.  Other novel designs have been explored, such as a multi-legged robot
that can traverse through rugged and steep terrain~\cite{he2025tactile}, a
forestry robot that moves between tree trunks using a grasping motion
\cite{RuralDelivery_grasping_robot_2015}, and a wheeled ground robot with an
integrated winch for extremely steep terrain~\cite{McGarey2018}. Soft robots can
also help with embodied physical intelligence strategies, such as twisting
\cite{ZhaoChi+22}, to efficiently address unstructured terrain locomotion.

Some environments require systems that can traverse multiple modes of
transportation (e.g., land, water, and air). For example, Nguyen \etal\
\cite{nguyen2024aerial} highlight the challenges of navigating complex costal
ecosystems such as mangroves where measurements are necessary from the air, on
the ground, and underwater. Other work has suggested the use of multi-robot
teams, such as the proposal by Couceiro \etal\ \cite{couceiro_semfire_2019} to
use aerial platforms to determine traversable terrain and goal destinations for
a larger ground vehicle in a forest vegetation management scenario.

Future work in this area should consider on designing terrain-adaptive robotic
systems that account for both mechanical reliability and perceptual robustness.
This is particularly relevant for navigating extreme or multi-modal environments
where conventional mobility strategies break down.

\titledsubsubsection{Resource-constrained autonomy for environmental monitoring}

For use cases where robots need to operate autonomously in the field for
long-duration missions, it is important that they both move and compute
energy-efficiently to best utilize their battery or solar panel capacity. Recent
work has considered insect-inspired robots that consume on the order of
milliwatts~\cite{wood2013flight, koh2015jumping, chukewad2020robofly}, including
platforms that can move untethered using solar
panels~\cite{jafferis2019untethered}, and in water and
air~\cite{chen2018controllable} with capability for trajectory tracking and
control~\cite{tagliabue2023robust}. One interesting example of optimizing for
energy-efficiency is in Hsiao \etal\  where they use the perching of a UAV on
walls and ceiling to reduce the power consumption of a UAV by
50-85\%~\cite{hsiao2023energy}. In addition to miniaturization as a means to
reduce energetic costs of actuation, there has been work to use buoyancy such as
in helium blimps~\cite{palossi2019extending} and ocean
gliders~\cite{rudnick2004underwater}. These platforms can benefit from further
research on lowering the expertise and cost needed for practical deployments,
such as making them easier to manufacture. In addition, given the small scale of
actuation energy, the energy spent on computation for autonomy algorithms is no
longer negligible; this necessitates more efficient
algorithms~\cite{li2024gmmap, han2015deep, fu2025dectrain, janson2015fast}, as
well as specialized hardware (e.g., ASIC chips)~\cite{suleiman2019navion} to
allow these robots to navigate intelligently and autonomously. Further research
is needed for full-system demonstrations of energy-efficient autonomy onboard
insect-scale robots, as well as for improving understanding of the energy and
financial cost trade-offs of deploying more advanced autonomy on these
platforms.


\begin{boxMarginLeft}
    \vspace{1em}
    \subsubsection*{Future Directions: Field Robotics in Land Use}
    \begin{itemize}
        \item \textit{Modular multi-sensor payloads. }Designing, constructing, and calibrating multi-sensor payloads is a laborious effort. Standardized open source or commercial options could streamline this process, at least for initial algorithmic prototyping.
        \item \textit{Simulators for full-stack autonomy.} Field robotics always requires substantial systems engineering, and the barriers to entry are exacerbated when testing must be conducted in specialized domains. Simulators could help advance algorithmic developments for these challenging domains. This is especially true if these simulators include task-based metrics that prioritize functional improvements.

        \item \textit{Retrofitting existing machinery.} Instrumentation, teleoperation, and automation of existing land use machinery is a clear area for commercial development. This presents various engineering challenges, as well as algorithmic challenges related to calibration and system identification with new platforms.
        \item \textit{Focus on robustness, serviceability, and energy-efficiency.} Land use applications often require robots to serve long missions under harsh conditions in remote areas. Deployable systems should be designed to tolerate damage, be easily serviced in-field, and be energy-efficient to extend the mission capabilities while operating on a battery.

    \end{itemize}
\end{boxMarginLeft}

\titledsubsection{Manipulation}
\label{sec:land_use:manipulation}

Robots capable of physically interacting with their environment can automate
procedures to develop and steward lands sustainably. Systems have been proposed
to accomplish tasks like wielding tools and handling plants. Open problems in
algorithm and mechanical design continue to limit progress on efficient mobile
manipulation, functional grasping, and dexterous manipulation of delicate
objects.

\titledsubsubsection{Mobile manipulators for cultivating crops}

Effective automation in agriculture and forestry depends on the availability of
highly capable mobile manipulators. Mobile manipulators have successfully been
deployed in fruit harvesting scenarios where robots navigate orchards, detect
ripe fruits, and carefully pick them \cite{Kuznetsova+20}. These systems rely
heavily on perception and grasp planning to jointly solve navigation and
manipulation. Synthesizing detailed large-scale maps (e.g., of trees in
vineyards) may require data fusion, such as integrating GNSS data into visual
odometry pipelines to localize point cloud representations of trees
\cite{SantosKoenigkan23, SantosGebler21}. Robot arms can carefully grasp and
detach fruit through reliable visual feedback and closed-loop control
\cite{ChangHuang24,ZhangLammers20}.  The proficiency of these systems at
navigating crop fields and picking may translate well to mechanical harvesting,
mechanical weeding, and targeted pesticide application \cite{Kant+23,Ahmadi+23}.
While there are few demonstrations of automated weed removal, promising proofs
of concept are paving the way to feasible solutions~\cite{wu2020robotic}. There
is an active conversation on developing standardized benchmarks for manipulation
(e.g., a series of workshops at premier robotics conferences over the last two
years), which presents an opportunity to incorporate manipulation benchmarks
that cover not only traditional manipulation applications inside homes and
factories, but also application areas for land use (e.g., mechanical weeding,
automated harvesting, and carbon soil sampling).


\titledsubsubsection{Gripper design for plant handling}

Since land management tasks require interaction with delicate objects (e.g.,
plants, animals, and organic materials), compliance becomes a necessary design
feature for robot hardware and software. However, the ability to carry out
forceful task execution remains important. Soft robotic grippers leverage
compliance of materials when performing tasks that, beyond a certain level of
complexity in object geometry and contact dynamics, are often unsolvable using
traditional grippers and control frameworks.  Suction-based grippers
\cite{Koivikko+21} provide reliable grasping of smooth convex objects, and their
efficient actuation makes them suitable for speedy object transporting.
Manipulation tasks, such as those in agricultural sectors, generally require
force-based motion to pick fruit, weed, and prune plants.  Additionally, organic
objects may have non-convex geometry, non-smooth material properties, or they
may be small, making high degree-of-freedom end effectors more appealing. Some
hybrid approaches \cite{Chin+20, Velasquez+24} achieve multimodal strategies for
carefully grasping objects with differing material properties, shapes, etc.
Roller grippers \cite{Chapman+21} are a new breed that may work well for weeding
tasks, where damaging the plant during extraction is permissible.  Soft fluidic
actuators form the basis of many soft robotic platforms.  These offer compliance
and delicate interactions, but can be expensive to manufacture and operate, are
susceptible to punctures, and often require bulky pumping infrastructure.
However, new techniques are promising for enabling soft fluidic actuators that
are strong, computationally designed, delicate, and inexpensively prototyped
\cite{li2017fluidDrivenOrigamiMuscles, li2019magicBallGripper,
becker2022tentacleGripper, bauer2022rapidSoftHands}. Alternative designs use
more traditional actuators as a structural basis. Pagoli \etal\  presented a
simple gripper design with three silicone fingers and an active palm for fully
dexterous in-hand manipulation \cite{Pagoli+21}. Unique to their design, the
fingers can dynamically stiffen via a system of retractable rods to increase
surface contact with grasped objects.  Tactile sensors
\cite{Dong+17,WangShe+21,LinZhang+23} enable precise force-based control
\cite{Arian+14, YuLiang+24,Wang+20} by capturing local observations of object
geometry and contact state. Based on tactile features, contact-rich tasks
requiring delicate regrasping and grip stabilization can be solved by simple
closed-loop controllers \cite{Veiga+20,Wang+21}.
Future work should aim to unify compliant design with precise force control,
enabling robust, low-damage manipulation across the wide variability of plant
geometry, material properties, and interaction tasks encountered in land
management.

\begin{boxMarginLeft}
    \vspace{1em}
    \subsubsection*{Future Directions: Manipulation in Land Use}
    \begin{itemize}
        \item \textit{Energy-efficient soft actuators.} While soft manipulators are highly appealing for delicate, conformal, or forceful tasks (e.g., handling fruits or pulling weeds), they often rely on actuation mechanisms that are impractical for land use scenarios, requiring high power, bulky and heavy machinery, and low frequency control.
        \item \textit{Extending applications via more dexterous end effectors.} Existing approaches for fruit harvesting could serve mechanical weeding, pest removal, and plant pruning applications. Designing robot form factors and control algorithms that maximize the end effector's reach would improve performance in scenarios where the target object is distant or dynamic. Possible improvements might include designing lightweight and more dexterous end effectors that are capable of articulating within dynamic environments while treating weeds, removing pests, or pruning plants.
        \item \textit{Datasets and benchmarks.} There is a need for collecting datasets and developing benchmarks with relevant domain-specific sensors (e.g., soil organic carbon, moisture, and compaction), and for hardware and data needed to apply robotic manipulation algorithms (e.g., soft actuator manipulators, occupancy maps, RGB-D data, pose estimates, and stiffness of the materials being gripped) for land use domains.
    \end{itemize}
\end{boxMarginLeft}


\def\EarthSysName{Earth Systems}
\def\EarthSysAuthors{Alan Papalia, Bianca Champenois, Shouyi\\Wang, Levi Cai, and Hanumant Singh}
\section[\EarthSysName]{
  \textbf{\EarthSysName}%
  \hfill
  \parbox[t]{0.55\textwidth}{\raggedleft\small\itshape \EarthSysAuthors}%
 }
\label{sec:earthsciences}

\begin{boxMarginLeft}

      The ocean, atmosphere, and cryosphere (i.e., all frozen water on Earth) are
      essential components of the Earth system, each playing a critical role in
      regulating climate. The ocean absorbs and redistributes heat and carbon dioxide,
      serving as a major climate buffer; the atmosphere mediates global temperatures
      through the greenhouse effect; and the cryosphere reflects solar radiation and
      influences ocean salinity, circulation, and sea level. Understanding and
      managing these systems is vital for both climate mitigation and adaptation.
      Robotics and autonomy can support this effort by enabling persistent monitoring,
      improved data collection for predictive models, direct support of human
      activities, and responsive environmental management. We identify four key areas
      of research where these technologies can contribute:

      \begin{itemize}
            \item \emph{Controls \& Planning}: Developing multi-agent systems for
                  collaborative data collection, decision-making algorithms for optimal
                  sampling strategies, and autonomous navigation in extreme environments
                  like deep oceans and ice-covered regions.
            \item \emph{Estimation \& Perception}: Advancing underwater
                  localization, mapping, change detection, and data assimilation
                  techniques to improve real-time monitoring and enhance our understanding
                  of dynamic Earth systems.
            \item \emph{Field Robotics}: Designing scalable, long-lasting, and
                  energy-efficient robots equipped with advanced sensors, payloads, and
                  instrumentation to measure new quantities of interest. These robots must
                  also be capable of operating in extreme and challenging environments,
                  with resilience to harsh conditions such as strong currents, ice, and
                  low visibility.
            \item \emph{Manipulation}: Sampling of important climate indicators (e.g.,
                  fragile biological samples or sediment cores) and inspection and repair
                  of critical infrastructure (e.g., marine energy platforms or aquaculture).
      \end{itemize}

      Advances across these themes offer the potential to transform our ability to
      observe, interpret, and manage Earth systems --- enabling more effective climate
      action, sustainable resource use, and long-term adaptation strategies. By
      addressing the challenges of data sparsity, expanding real-time monitoring,
      and supporting interdisciplinary collaboration, robotics can become a
      critical tool in understanding and protecting the ocean, atmosphere, and
      cryosphere in a rapidly changing climate.

\end{boxMarginLeft}

\begin{figure}[h]
      \centering
      \includegraphics[width=1\linewidth]{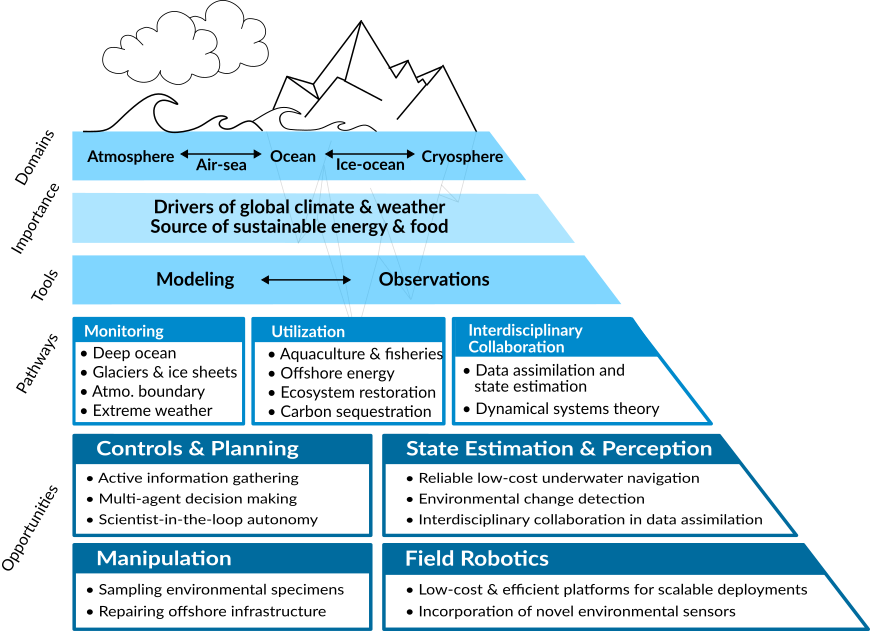}
      \caption{
            An overview of the discussion in this section.
            \textbf{Domains:} the scope of this section is the atmosphere, ocean,
            cryosphere, and their interactions.
            \textbf{Importance:} these systems are core drivers of
            Earth's climate and provide opportunities for sustainable utilization.
            \textbf{Tools:} the primary tools to study and work in these Earth
            systems are modeling and observational systems.
            Many of the contributions the robotics community can provide are
            through enhancing these tools.
            \textbf{Pathways:} we describe three major pathways for the robotics
            community to make an impact in these areas: improving monitoring
            capabilities, aiding in sustainable utilization efforts, and in
            interdisciplinary collaborations combining robotics skills and domain
            expertise (e.g., leveraging state estimation tools in data assimilation
            for climate modeling).
            \textbf{Opportunities:} we highlight specific
            areas where robotics can make a significant impact, such as developing
            low-cost underwater navigation to advance in situ ocean monitoring.
      }
\end{figure}

Earth's natural systems are central to Earth's climate and provide valuable
pathways for climate change mitigation and adaptation. Advancing the study of these systems (collectively, the Earth sciences) is therefore fundamental to
understanding and battling climate change. Towards this goal of understanding
and addressing climate change, this section will largely discuss the role that
the robotics community can play in advancing the Earth sciences.  Additionally, since the technologies useful in studying Earth systems are often beneficial to general activities (e.g., aquaculture and energy production) in these often remote and challenging environments, we will also discuss how robotics can support climate-positive activities in these domains.

This section focuses on three specific Earth systems: the
ocean, atmosphere, and cryosphere (all of the ice on Earth).  The focus
on these systems is due to their leading roles in the Earth's climate.
The atmosphere serves as the primary medium for heat exchange between Earth's
surface and space. It also facilitates the transport of water vapor, heat, and
other gases, playing a key role in regulating global climate patterns
\cite{trenberth2011changes}.
The ocean is the planet's largest sink for both heat and carbon, absorbing approximately 90\% of the excess heat generated by anthropogenic warming \cite{ipcc_ar6_oceans} and around 25\% of carbon dioxide emitted by
human activities \cite{gruber2023trends}. It also sustains significant biomass and biodiversity, while regulating weather and climate through the global transport of heat and moisture.
In addition to its critical role in Earth's climate system, the ocean presents major opportunities for advancing more sustainable human activities.
The cryosphere, which  includes ice sheets, glaciers, sea ice, permafrost, and
snow cover,  reflects sunlight back into space, houses large stores of
greenhouse gasses, influences ocean circulation patterns, and stores nearly 70\%
of the planet's freshwater \cite{IPCC_2019_SROCC_Ch-1}.

Altogether, these three systems are core drivers of Earth's climate. While our
understanding of these systems has improved significantly in recent decades,
there are still many open questions with real-world implications (e.g., how much
will sea levels rise under certain climate scenarios? How could global weather
patterns change in a warming world?  \cite{IPCC_2019_SROCC_Ch-TS}).
By gaining a deeper understanding of these systems, we can better identify climate impacts, more accurately answer scientific questions, and enable the creation of sustainable technologies and practices.

This section aims to highlight how roboticists can contribute to climate solutions through three pathways related to the ocean, atmosphere, and
cryosphere.
The first pathway for impact considers the use of robotics to enhance monitoring
capabilities in the Earth sciences, which can drive substantial improvements in
our understanding of these systems.
The second pathway will consider the application of robotics in
specific climate-relevant areas of ocean utilization (e.g., marine energy or
aquaculture).
The third and final pathway will be the application of skills and tools from
robotics to traditionally non-robotic problems (e.g., applying state estimation
techniques for data assimilation in Earth systems modeling).

Although most research topics map naturally onto one of the three pathways, many
promising technologies blur these boundaries. For example, long-endurance
stratospheric balloons equipped with miniaturized spectrometers could (i)
deliver continuous greenhouse-gas profiles that feed atmospheric-carbon models
(monitoring) and (ii) supply real-time wind and turbulence data that help
optimize the routing of transcontinental cargo flights, reducing fuel burn and
emissions (utilization).\footnote{While relevant here, we discuss
      transportation (e.g., aviation and shipping) in \Cref{sec:transport},
      rather than in this section.
}
Because overlaps of this sort are common, the pathways are meant as a conceptual
guide rather than strict bins for classifying every contribution.

\titledsubsection{Overview of Earth Systems}
\label{sec:earthsciences:overview}

\begin{figure}
      \centering
      \includegraphics[width=\linewidth]{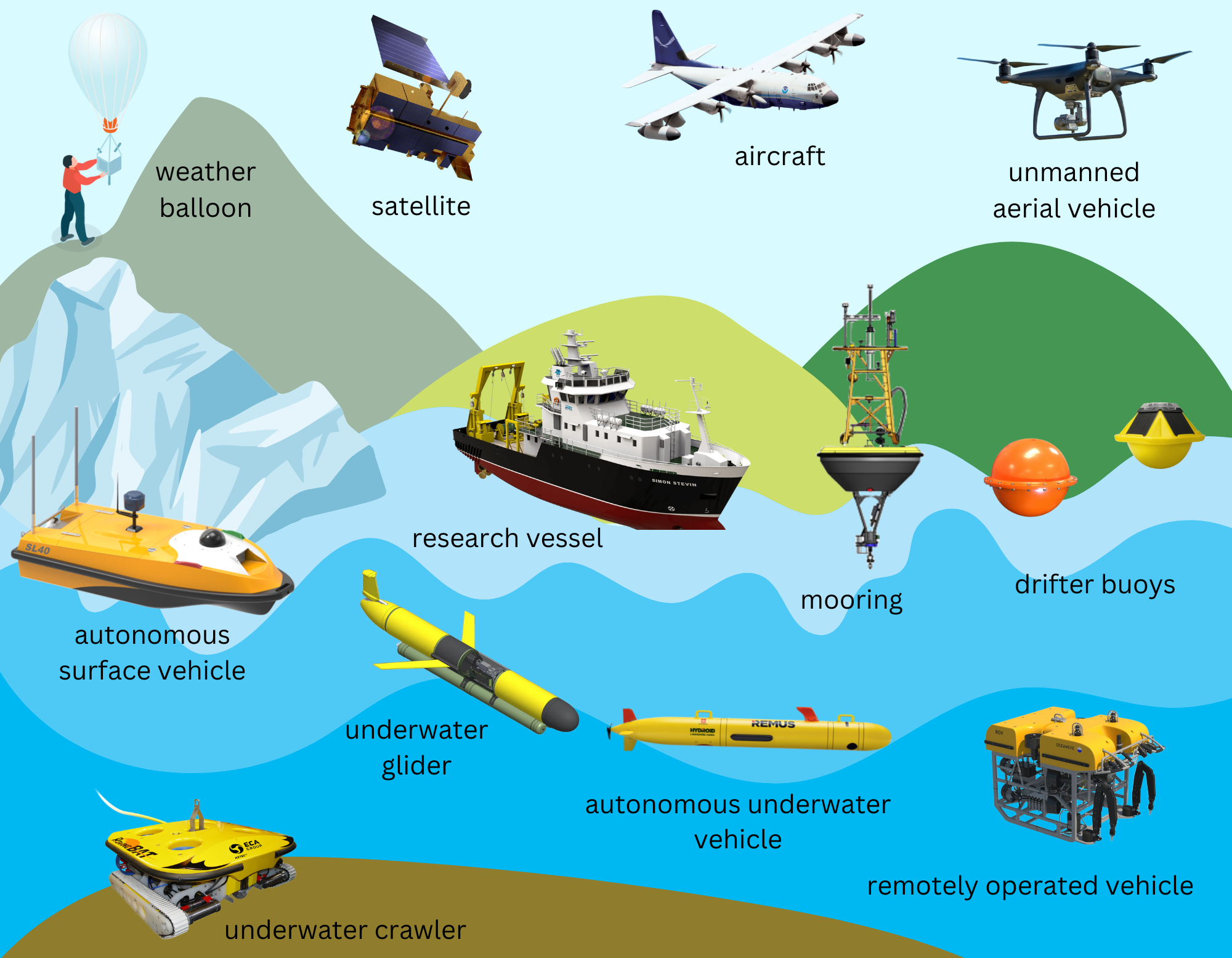}
      \caption{Common autonomous and semi-autonomous platforms used for
            monitoring of Earth systems}
      \label{fig:earth-sciences:overview-of-earth-sensing}
\end{figure}

In this section, we will provide a high-level overview of the Earth's systems
with the aim of providing sufficient context to understand how the robotics
community can make a climate impact in this area.
We will begin by discussing
the core technologies found across Earth systems work, namely modeling
and observational systems.
These two broad tools underpin most of the scientific work done in the Earth systems
and provide much of our understanding of these systems and how to work with
them.
We will follow up by discussing the scientific foundations of the Earth systems,
highlighting processes and phenomena that establish the importance of these
systems for climate change. Additionally, we discuss current data bottlenecks
that limit our understanding of these systems (and thus limit our ability to
predict and adapt to climate change).
We then discuss areas where human civilization (i) extracts resources from these
systems (e.g., energy or food from the ocean) and (ii) attempts to intervene in these
systems to mitigate or adapt to climate change (e.g., ecosystem restoration).
Finally, to showcase the impact of robotics in these areas, we will provide
some current success stories where the robotics community has made significant
contributions to the Earth sciences.

\titledsubsubsection{Modeling and Observations}
\label{sec:earthsciences:modeling_obs}

Earth scientists rely on two core technologies, \emph{modeling} and
\emph{observation systems}, to better understand natural processes and to
support responsible management of these systems. The interaction of modeling and observational data informs our scientific understanding of Earth's climate and enables the sustainable use of resources (e.g., offshore energy, aquaculture, transportation) \cite{Edwards2015}. This section offers a concise overview
of these two foundational tools, setting the stage for exploring how
the robotics community
can contribute to Earth systems science.

\paragraph{Earth Systems Modeling}

Modeling provides researchers with a powerful means to investigate scenarios and
locations that are otherwise expensive, difficult, or impossible to observe
directly, such as remote polar regions or future climate states
\cite{eyring2016overview,Fox_Kemper2019}.  Models help us characterize and predict complex,
large-scale processes in the Earth system by enabling capabilities such as
forecasts of weather events, assessments of offshore energy potential, and
simulations of global heat and moisture transport.

Nevertheless, these models carry inherent uncertainties and approximations. The
chaotic nature of many Earth processes makes them highly sensitive to initial
conditions and small errors, such that even slight deviations in the estimated
state can lead to large discrepancies in model predictions
\cite{Leutbecher2008}. Moreover, Earth systems are exceptionally
high-dimensional and governed by complex, interacting physical, chemical, and
biological processes. Capturing this multiphysics complexity often requires the
use of parameterizations: functional approximations of processes that are not
explicitly modeled because they are either subgrid-scale or insufficiently
understood to be explicitly resolved. While parameterizations help keep models
computationally feasible, they may introduce systematic biases if not carefully
constructed and validated \cite{McFarlane2011}.The multiscale features of the system also require very high-resolution discretization grids.

Observational data are critical to addressing model uncertainties and reducing
model errors.  Data-dependent tasks such as model initialization, calibration,
and validation are important to ensure the accuracy and reliability of models.
Observations can be directly integrated into the modeling process via \emph{data
      assimilation}, a technique analogous to state estimation in robotics that
combines observations with model predictions to produce more accurate estimates
of the system state \cite{barker2012weather,anderson2001ensemble}.

\paragraph{Observing Systems}

Observations, or measurements of Earth system variables (e.g., temperature,
salinity, wind velocity, or ice thickness) are key ingredients for Earth system
models and our broader scientific understanding. They provide evidence for newly
identified phenomena and allow scientists to support, refine, or replace model
assumptions. \emph{Observing systems} refer to the suite of
technologies and methodologies used to generate these measurements, spanning
everything from vertical ocean profilers and weather balloons to
satellite remote sensing \cite{moltmann2019global}.

Because observations are critical for both model fidelity and scientific
discovery, the Earth sciences use a diverse array of observing systems to capture
the wide range of natural processes at play. This diversity also ensures that
data from different methods can be cross-verified, minimizing bias and enhancing
the quality of measurement. Even so, substantial gaps remain in existing
systems, highlighting key areas where new techniques that fill data gaps
could be transformative
\cite{Revelard2022}. For instance, robots capable of autonomously surveying
remote or hazardous locations can capture vital data where other methods are too
risky, expensive, or imprecise.

Broadly speaking, there are two forms of observations. The first form, \emph{in situ
      observations}, involves taking measurements directly at the location of
interest. For example, this category includes instruments placed on ships or autonomous
underwater vehicles (AUVs) that record temperature profiles at depth.  The
second form, \emph{remote sensing}, gathers data from a distance, as with satellites
or aircraft that measure ocean color or atmospheric properties over large areas.
In situ methods can provide high-resolution ground-truth
measurements typically over smaller areas, whereas remote sensing offers broader spatial coverage, often at lower
resolution. Robotic platforms span both categories: an AUV or mobile land robot
would collect in situ observations, whereas a fixed-wing drone might conduct
large-area remote sensing surveys.

These various observation methods complement each other, enabling scientists to
gather information at multiple scales. Indeed, the ongoing shift toward
\emph{data-driven} methodologies (e.g.,
\cite{yuval2020stable,lai2024machine,bracco2024machine}) in Earth science
underscores the growing importance of flexible, robust, and cost-effective ways
to acquire data. This is where robotics can make a decisive impact by extending
observational reach, allowing adaptive placement of instrumentation, and
deploying specialized sensing payloads across challenging environments. Whether
capturing high-resolution data in turbulent seas or mapping ice thickness
beneath polar ice shelves, robotic technologies stand to greatly expand both the
quality and quantity of observational data needed for accurate modeling and
informed climate strategies.

\titledsubsubsection{Challenges: Processes, Phenomena, and Data Gaps}

Climate change manifests in diverse and interwoven ways across the atmosphere,
oceans, and cryosphere. Each of these Earth system components is subject to
complex physical and biochemical processes that are not always straightforward
to observe, especially given the harsh conditions or remoteness of many key
locations. We outline a few prominent phenomena that drive research
and highlight potential roles for robotics, from capturing hard-to-access data
on extreme events to exploring under-sampled marine and polar habitats.

\paragraph{Atmosphere and Extreme Weather}

Although the atmosphere is often associated with the greenhouse effect, its
influence on global climate extends far beyond the retention of heat. As the
primary medium for transporting heat, moisture, and momentum across the planet,
the atmosphere exerts a profound influence on both daily weather and longer-term
climate patterns. Climate change is expected to intensify many forms of extreme
weather, including heatwaves, droughts, floods, and tropical storms
\cite{cook2022megadroughts, stevenson2022twenty}, yet the precise distribution and magnitude of these
changes remain uncertain.
Accurate weather forecasting and climate projections rely heavily on
comprehensive and high-quality atmospheric observations. However, models are
still limited by the quality of observations. For example, satellite remote
sensing still suffers from known biases and limited resolution
\cite{khanal2020remote}, especially under cloud-covered or stormy conditions.

In this context, robotic technologies can enhance \emph{in situ} data collection
by providing adaptive and easily deployable platforms that gather temperature,
humidity, and wind measurements in or around developing storms
\cite{holbach2023recent}.  For instance, autonomous aerial vehicles could be
launched preemptively to sample the rapidly evolving microclimates of hurricanes
or typhoons, refining near-term forecasts and informing
disaster-prevention efforts. These capabilities build on historical lessons:
improved typhoon and hurricane monitoring has already contributed to fewer
storm-related fatalities, and extending these monitoring methods into more
extreme conditions will be essential to ensuring societies can adapt to an
increasingly volatile climate \cite{franzke2020risk}.
\alan{could include more atmospheric stuff here from e.g., CU Boulder and Nebraska-Lincoln}

\paragraph{Ocean Dynamics and Coastal Environments}
The ocean, Earth's largest carbon and heat sink, is central to climate
regulation. It absorbs up to 90\% of excess heat from anthropogenic warming
\cite{ipcc_ar6_oceans} and sustains roughly 80\% of all animal biomass. This
massive biological and thermal capacity drives planetary-scale processes such as
ocean currents, the global transport of heat and nutrients, and the cycling of
carbon through air-sea exchange. Yet, multiple oceanic phenomena are projected
to intensify under continued warming, including sea-level rise, marine
heatwaves, acidification, and deoxygenation
\cite{frolicher2018marine}.
These changes raise the specter of widespread
coral bleaching, reduced fish stocks, altered current patterns, and other
cascading impacts that affect societies around the globe.
\alan{could have citations for these possibilities}

Research priorities identified by the IPCC AR6 report \cite{ipcc_ar6_oceans}
underscore the importance of long-term ocean observations to track and model
these trends. However, many regions remain under-instrumented, particularly the deep ocean, coastal margins,
and dynamic boundary currents
\cite{loose2021leveraging}.
Conventional platforms
like Argo floats and research vessels have significantly expanded coverage of
the open ocean, but they can struggle in shallow or topographically complex
areas where strong currents may carry them away or where collision risks are
high
\cite{roemmich2009argo}.
For
example, monitoring in coastal zones may be complicated by increased
human activity, turbidity, and rapidly changing currents.  Robotic systems,
including autonomous underwater gliders and propeller-driven vehicles, can help
fill these gaps by offering persistent, fine-scale measurements of temperature,
salinity, dissolved oxygen, and other key variables
\cite{tanhua2019we}.
Combining such
in situ observations with other measurements such as satellite data can improve our
understanding of both short-term phenomena like marine heatwaves and longer-term
events like changing ocean circulation patterns.

\paragraph{Cryosphere}

Encompassing ice caps, glaciers, sea ice, permafrost, and massive ice sheets, the
cryosphere is undergoing rapid change due to rising
temperatures~\cite{IPCC_2019_SROCC_Ch-TS}.
Melting ice has direct consequences for global sea levels, ocean salinity, and
potentially the rate of warming itself. For instance, as the Greenland and
Antarctic ice sheets recede, they introduce large volumes of freshwater into the
ocean, which can modify global ocean circulation patterns and drive large
amounts of sea level rise. Vast permafrost regions are similarly vulnerable:
their thaw can release considerable methane and carbon dioxide, reinforcing
warming trends.

Autonomous robotic systems can play a transformative role in cryospheric science
by accessing environments that are dangerous, remote, or otherwise unreachable
for human researchers. For instance, autonomous underwater vehicles
(AUVs)~\cite{waahlin2024swirls} can operate beneath floating ice shelves to
collect data from the \emph{ice-ocean interface}, a critical but poorly
observed region where seawater interacts with the base of the ice.
These vehicles can collect data on how quickly the ice is melting from below,
map how water flows in the hidden space beneath the ice shelf, and monitor how
heat and currents interact right at the ice-ocean boundary. These processes
directly affect how stable the ice shelf is, but most of this environment has
never been mapped, and very few long-term measurements exist.
\alan{find some citations for these questions}

A particularly high-priority region is the \emph{grounding zone}, the
transition where the ice sheet goes from grounded on the seabed to floating.
This zone strongly influences how fast ice flows into the ocean, but it is even
less observed than the cavity beneath ice shelves. Robotic platforms can help
close data gaps by observing subglacial freshwater plumes, monitoring flexing of
the ice due to ocean tides, and tracking the position of the grounding line (the
boundary where the ice sheet lifts off the seabed) over time.  They can also
build high-resolution maps of the seafloor beneath the ice, track how sediment
moves through these environments, and identify features like channels that guide
water flow. These kinds of observations are critical inputs for climate models
that aim to predict how quickly ice will retreat and how much it will contribute
to sea level rise.
\alan{find some citations for these questions}

Beyond the ice-ocean interface, aerial drones can provide high-resolution
mapping of ice sheets and glaciers~\cite{teisberg22uav}, enabling surface and
subsurface structure analysis. While permafrost has seen less robotic attention,
similar techniques like high-resolution surface mapping or in situ
temperature profiles could support studies of thaw
dynamics, carbon release, and landscape transformation. As robotic tools mature,
they promise to help close long-standing observational gaps across the
cryosphere.

\paragraph{Key Interactions and Climate Tipping Points}

Because Earth's systems are deeply interconnected, changes in one domain can
cascade into others. Air-sea exchanges of heat, moisture, and gases shape global
weather patterns. Fresh meltwater entering the ocean modifies circulation
and nutrient flows. Observing these cross-system interfaces poses unique
technical hurdles, for example, where ice shelves extend for hundreds of
kilometers over open water. Direct measurements of warm ocean water contacting
the underside of ice are essential for tracking melt rates, but require
platforms that can reliably navigate such distances beneath thick ice.
\alan{could use more citations}

Understanding such processes is vital for detecting or predicting \emph{tipping
      points}: critical thresholds that, once crossed, can lead to major and often
irreversible shifts in Earth's climate
\cite{lenton2008tipping}.
Examples
include the potential collapse
of the Greenland or West Antarctic Ice Sheets, each carrying the risk of meters
of sea-level rise \cite{ipcc_ar6_oceans}. Similarly, rapid warming in the
Arctic, termed Arctic Amplification, could lead to substantial permafrost thaw,
which could release vast amounts of methane and carbon dioxide, further
accelerating warming.  Pinpointing the exact point at which these tipping points
are crossed remains a formidable challenge, partly because many occur in regions
with limited historical data. More robust monitoring, especially in remote
parts of the ocean and cryosphere, can yield early-warning indicators and
inform climate adaptation strategies, thus mitigating severe environmental and
socio-economic consequences.

\paragraph{Data Sparsity}


As discussed throughout this section, many regions of the planet remain
under-instrumented, leading to \emph{data-sparse} conditions that limit our
ability to model, forecast, and understand critical Earth system processes.
Such gaps may arise from a variety of factors, including seasonal disruptions to
remote sensing, the inaccessibility of certain locations, and logistical or
political constraints on sensor deployment.

We define data sparsity broadly as a lack of observations (e.g., sea surface
temperature) that meaningfully degrades our capacity to track or model key
Earth processes. While sparsity can arise for many reasons, we focus here on
three recurring forms that robotics is well-positioned to address:
\emph{spatial sparsity}, \emph{temporal sparsity} (or \emph{spatiotemporal sparsity}
in combination), and \emph{modal sparsity}. We believe this characterization is
useful to better contextualize the challenges of data sparsity in Earth sciences
and how robotics can help address them.

\subparagraph{Spatial sparsity.} This form of sparsity is often caused by either
limitations related to resolution or patchiness in sensing capabilities, or
remoteness and difficulty of access to physical regions, causing large permanent
gaps of knowledge. One example of this are deep-ocean basins (below 2,000\,m), which are difficult to observe due to the inability of satellite signals to penetrate
to such depths and unsuitability of most platforms (e.g., standard Argo floats
\cite{owens2022oneargo}) for deep-water deployment
\cite{zilberman2023observing}. Creative solutions have begun filling some gaps
in traditionally difficult to observe areas (e.g., attaching sensors to marine
animals \cite{boehme2009animal,harcourt2019animal}), however such measurements
remain limited, intermittent, and geographically patchy.

In the ocean, mesoscale eddies (50-200\,km in diameter) play a significant role
in heat and nutrient transport. While they can be observed through multiple
means, including satellite altimetry, drifting buoys, and autonomous underwater
vehicles, the spatial and temporal resolution of these observations is often
insufficient to fully characterize eddy dynamics across the global ocean
\cite{seo2023ocean}. Similarly, identifying and localizing sharp temperature
gradients (e.g., at ocean fronts where warm and cold waters meet) is important
to understanding weather patterns and marine ecosystems, but requires
high-resolution, in situ measurements that are rarely available
\cite{mccammon2021ocean}. Similar spatial gaps exist in polar and mountainous
regions where steep topography, ice cover, and extreme temperatures prevent
long-duration in situ instrumentation.

\subparagraph{Temporal sparsity.} Temporal sparsity in Earth observations can
arise from (i) limited historical records (e.g., no data prior to 2000), (ii)
seasonal or episodic conditions that block observation (e.g., monsoons, polar
winters), or (iii) coarse temporal resolution (e.g., monthly sampling).

Historical gaps often require reconstruction via proxy methods like
sediment cores, ice cores, or coral samples, but these introduce uncertainty.
Seasonal phenomena can further inhibit observation: persistent cloud cover
during the Indian monsoon \cite{wonsick2009cloud} and darkness or storms in the
Southern Ocean during polar winter \cite{pope2017community} can render
satellite-based monitoring ineffective for months, creating concurrent temporal,
spatial, and modal gaps.
Even when measurements are available, low-frequency sampling may miss transient
events, limiting insight into fast-evolving Earth system processes.

\subparagraph{Sensing modality sparsity.}
Many climate-relevant variables remain difficult to measure due to the lack of
scalable in situ sensors. Laboratory methods exist for some quantities (e.g.,
nutrient concentrations, respiration rates), but are often infeasible for field
deployment or degrade in fidelity during transport \cite{del2002respiration}.
Inference methods, such as using conductivity as a proxy for salinity, offer
partial solutions but rely on stable empirical relationships that may not
generalize across environments.

Critical gaps persist in observing ocean carbon fluxes
\cite{dai2022carbon,hauck2020consistency}, air-sea interactions (e.g., wave
spectra, gas exchange) \cite{centurioni2019global}, and marine biological
activity.  Robotic systems offer a path forward by enabling persistent,
high-resolution sensing in extreme or inaccessible regions. Embedding emerging
modalities like miniaturized chemical and genetic sensors in
autonomous platforms can help fill longstanding observational gaps.

\subparagraph{Takeaways on data sparsity.}
Addressing these forms of data sparsity is crucial for improving our
understanding of Earth system processes and advancing predictive capabilities.
Robotics has the potential to bridge many of these gaps by enabling persistent,
autonomous observations in remote or harsh environments, facilitating
higher-resolution data collection, and expanding access to variables that
traditionally require laboratory analysis.
One of the great possibilities offered by robotics is to enable
high-resolution, in situ spatiotemporal measurements, patching gaps in both
space and time and greatly aiding our understanding of Earth system processes.

\titledsubsubsection{Human Utilization and Interventions}
\label{sec:earthsciences:human-interaction}

Human interactions with the ocean, atmosphere, and cryosphere can broadly be
categorized into two areas: \emph{utilization} of these systems for resources
and economic activities, and \emph{interventions} designed to mitigate or adapt
to the impacts of climate change. The majority of such human interactions focus
on the ocean, largely due to its vast size and the long place in human history.

\paragraph{Ocean Utilization}

The ocean is increasingly seen as a source of sustainable energy, food, and raw
materials. For instance, offshore wind farms are rapidly expanding in regions
where steady winds and continental shelf configurations offer favorable
conditions \cite{RODRIGUES20151114, DIAZ2020107381}. However, developers must contend with harsh environments marked by saltwater corrosion, powerful waves, and the risk of extreme storms. Improved modeling of ocean conditions along with monitoring of offshore structures can help ensure the safety and longevity of wind farm installations \cite{shaw_2022, MITCHELL2022100146, Wang2024}.

The ocean is also a primary means of shipping goods and people around the globe.
High-quality ocean models can improve not just safety but also route efficiency
by predicting weather patterns and currents (see \cref{sec:transport} for more
discussion) \cite{Boren2022, jmse12101728}.

In terms of meeting future global food demands, both wild fisheries and
aquaculture play crucial roles. However, there are numerous challenges
such as overfishing, pollution, warming waters, and disease outbreaks. Robotic
systems equipped with various sensors (e.g., cameras, sonar, environmental
sensors) can routinely survey fish stocks, habitat condition, and water quality
parameters to provide `health snapshots' of both natural and man-made food
sources \cite{Antonucci2020, kelasidi2023robotics, FORE2024108676}. In wild fisheries, this high-resolution monitoring helps management
agencies set sustainable catch limits and quickly respond to phenomena like
harmful algal blooms \cite{Roessig2004}. In aquaculture settings, automated feeding and inspection
robots can detect early signs of disease, optimize feeding schedules, and reduce
negative environmental impacts. Ultimately, these advanced observation and
intervention tools can enable more resilient, climate-adaptive seafood
production and protect valuable marine ecosystems from further degradation
\cite{Aguzzi2022}.

\paragraph{Climate Interventions}

To mitigate or reverse climate-related threats, a variety of intervention
strategies have been proposed across Earth's oceanic, atmospheric, and
cryospheric systems. These range from habitat restoration to large-scale
geoengineering and differ widely in their maturity, feasibility, and risks.

\emph{Ecosystem restoration} efforts like coral gardening, artificial
reefs, and mangrove or seagrass restoration aim to rebuild coastal resilience
and enhance carbon sequestration
\cite{Layman2020,Rinkevich2021,Bayraktarov2020,Sinclair2020}. These ecosystems
provide habitat, reduce coastal
erosion, and act as significant carbon sinks. Robotic systems can support these
efforts through surveys to monitor growth, biodiversity, and habitat health
as well as by automating tasks like coral seeding
\cite{ridge2020unoccupied,madin2019emerging,AGUZZI2024195}.

\emph{Ocean-based carbon removal (mCDR)} strategies include enhancing the
biological carbon pump, artificial upwelling, and ocean alkalinity enhancement
\cite{national2021mcdr, renforth_assessing_2017}. These aim to increase the
ocean's long-term carbon uptake, but remain scientifically uncertain and
ecologically risky. Concerns include unintended disruption of marine food webs
and biogeochemical cycles. Monitoring, reporting, and verification (MRV) is
central to evaluating these interventions, and robotics offers a pathway to
cost-effective, high-resolution environmental observation and compliance
auditing to support this MRV \cite{bach_co2_2019, Doney2025}.\footnote{See \Cref{sec:land-use}
      for discussion of terrestrial carbon removal strategies, such as
      afforestation.}

\emph{Atmospheric interventions}, such as aerosol injection and cloud
brightening target radiative forcing directly
\cite{latham2012marine,smith2018stratospheric}. While technically distinct from
marine strategies, they share similar ethical, governance, and environmental
concerns. Most of these remain at the conceptual or pilot stage and would
require significant observational infrastructure and real-time modeling to
assess impacts and guide safe deployment.

Despite their diversity, all of these interventions depend on the ability to
observe complex Earth system responses over time and space. Robotic platforms capable of persistent sensing in remote, dynamic, or fragile
environments are likely to play a foundational role in enabling safe
experimentation, assessment, and, where appropriate, deployment.

\paragraph{Integrating Robotic Tools}

In all these utilization and intervention efforts, robotic systems offer unique
advantages for data collection, infrastructure management, and environmental
assessment. Autonomous underwater or aerial vehicles can perform routine checks
of offshore installations, detect and seal leaks in carbon capture and storage
facilities, or survey the success of a coral restoration project without
putting human operators into hazardous situations. By enhancing the efficiency,
consistency, and safety of such climate-positive activities, robotics can help
ensure that human interactions with Earth systems remain both economically
viable and environmentally responsible. Moreover, as we improve our ability to
observe and understand these interventions' outcomes, the knowledge gained will
guide more refined, adaptive strategies for climate resilience and mitigation.

\titledsubsubsection{Robotics Success Stories in Earth Systems}
\label{sec:earthsciences:success}

Robotics technologies have already played major roles in advancing our
understanding of the Earth's systems. We want to take the opportunity to
highlight some of these success stories as motivation for future contributions
from the robotics community. All of the examples we provide are of observational
technologies which provided novel capabilities for Earth systems.  A powerful
commonality among these examples is that their development was led by
non-roboticists, such as oceanographers and atmospheric scientists, who had deep
understandings of the observational gaps in their field and the impact that
these new technologies could have. This highlights the value and importance of
roboticists collaborating with domain experts to develop impactful technologies.
Below we discuss three well-established examples that have made major scientific
contributions.

\paragraph{Argo Floats}
The Argo program is a global
array of over 3,000 autonomous ocean floats that sample a range of ocean properties
(e.g., temperature, salinity) from the surface to a depth of 2,000 meters. These
floats drift with ocean currents and periodically dive to collect data, then
resurface to transmit data via satellites. The Argo program has revolutionized
our understanding of the ocean, e.g., allowing us to improve our certainty in
Earth's energy imbalance by a factor of four
\cite{wijffels2016ocean,johnson2016improving, Claustre2020}.  The impact of the Argo program
has echoed throughout the Earth sciences, having lead to over 5,000 scientific
publications and its extension to Biogeochemical Argo floats \cite{owens2022oneargo}. The success of the Argo program underscores
that relatively simple autonomous observation platforms can have a major impact
when focused on filling \emph{the right data gaps}.

\paragraph{Ice-Tethered Profilers}
Ice-Tethered Profilers (ITPs) are autonomous instruments that are deployed in
the Arctic to monitor ocean properties beneath the sea ice \cite{toole2011ice}.
These instruments are tethered to the ice and drift with the ice pack, using a
winch to move instruments up and down through the water column. ITPs have been
used to study a range of phenomena in the Arctic, driving insights into heat
transport mechanisms \cite{timmermans2008ice}, freshwater content and ice melt
\cite{proshutinsky2009beaufort}, oxygen circulation and biological processes
\cite{timmermans2010ice,laney2017euphotic}, sea ice variability
\cite{toole2010influences}, and gas exchanges and carbon cycling
\cite{islam2017sea}. Similar to the Argo program, the success of ITPs highlights
the value of targeted, autonomous observation platforms in Earth systems.

\paragraph{Small UAS for Meteorology}

Small uncrewed aircraft systems (UAS) have enabled targeted, in situ
measurements of atmospheric processes, including convection, boundary
layer dynamics, and storm supercell processes
\cite{kral2018innovative,reineman2016use,frew2020field}.  Their success
demonstrates how thoughtfully deployed robotic platforms can fill persistent
observational gaps in atmospheric science, improving forecasts and supporting
studies of weather extremes and climate-sensitive processes.

\titledsubsection{Controls \& Planning}
\label{sec:earthsciences:controls-planning}

Operating in Earth's dynamic, uncertain environments requires control and
planning methods that can account for partial observability, extreme
variability, and tight energy and communication constraints. At the same time,
the emergence of learned dynamical models offers promising tools for capturing
complex physical processes, paralleling similar efforts in Earth science to
develop data-driven parameterizations for poorly modeled phenomena.

\titledsubsubsection{Adaptive Sampling and Information Gathering}

Determining where, when, and how to collect data is a central challenge in
robotic Earth observation. Adaptive sampling and information gathering
strategies aim to direct measurements toward regions of highest scientific
value, often under tight constraints of energy, communication, and limited prior
knowledge. While these approaches are well studied in robotics, their deployment
in dynamic and transient environmental systems remains limited.

Many climate-relevant phenomena evolve on timescales comparable to the time
needed for robotic measurement. In such cases, platforms must balance
exploration and exploitation while updating plans in real time. Predictive
models are needed to anticipate environmental change and guide trajectory
updates \cite{flaspohler2019information, hitz2017adaptive, stache2021adaptive}.
These challenges are exacerbated for extreme events  like algal blooms,
severe storms, and rapidly moving ocean fronts  where timely, high-resolution
data are critical. Developing sampling strategies tailored to these
fast-evolving events is an open and important area of research.

Multi-robot systems present an opportunity to scale data collection while
improving resilience to platform failures. Heterogeneous teams (e.g.,
combinations of AUVs, gliders, buoys, and satellites)  can increase spatial and
temporal coverage if effectively coordinated \cite{edwards2023collaborative,
      salam2019adaptive}. Prior work has shown how teams can monitor ocean fronts
\cite{pinto2021boldly, faria2014coordinating, mccammon2021ocean} or biological
communities \cite{zhang2021system} by dynamically allocating tasks and sharing
local information. Collaboration can reduce redundant measurements, improve
robustness, and lower operational cost.

From a methodological standpoint, both homogeneous and heterogeneous teams have
been studied through coverage-based control \cite{karapetyan2018multi}, swarm
strategies \cite{duarte2016application}, and model-driven planning for static
\cite{hollinger2016learning,das2015data,kemna2017multi} and dynamic
\cite{salam2019adaptive,salam2023heterogeneous} fields. However, unified
frameworks for collaborative sampling remain underdeveloped. Real-world
deployments also demand robustness to intermittent communication and limited
shared belief. New algorithms are needed to address sensing and mobility
tradeoffs across platforms with varying capabilities.

Together, advances in adaptive and collaborative sampling will allow robotic
platforms to track evolving conditions, respond to rapid change, and fill
persistent observational gaps across the Earth’s oceans, atmosphere, and
cryosphere.

\titledsubsubsection{Flow-Aware Planning}

Mobile robots can harness environmental transport such as ocean currents and atmospheric winds to improve endurance, reduce energy use, and
expand coverage. Flow-aware planning exploits these dynamics to reach targets
more efficiently, especially in oceanographic and atmospheric missions.  Prior
work has demonstrated mobility gains by aligning paths with time-varying flows
\cite{ramos2018lagrangian,lermusiaux2016science,lolla2014time,knizhnik2022flow},
and more recent methods consider difficulties such as forecast uncertainty
\cite{kularatne2018optimal}.  Open challenges include partial observability of
both robot state and environmental conditions, integrating uncertainty,
determining appropriate environmental representations, and ensuring planners
remain tractable for onboard use.

\subsubsection{Scientist-in-the-Loop Planning}

As robotic systems become increasingly autonomous, incorporating human domain
expertise remains crucial for refining scientific objectives and ensuring
context-aware data collection. In Earth science, this typically manifests as
\emph{scientist-in-the-loop} autonomy, wherein researchers directly collaborate
with robots in real-time to guide data collection and interpretation. Such
approaches enhance robot adaptability, allowing robotic systems to adjust
sampling strategies dynamically in response to expert feedback. For instance,
interactive goal-refinement interfaces allow scientists to identify areas or
phenomena of particular interest, enabling robots to prioritize scientifically
valuable observations \cite{jamieson2020active}. Similarly, human experts can
improve semantic classification tasks by interacting directly with robots during
field operations \cite{samuelson2024guided}, thereby bridging the gap between
robot perception and scientific interpretation.

However, realizing effective scientist-in-the-loop autonomy becomes increasingly
challenging as robotic missions scale to multiple platforms operating
simultaneously. In such settings, communication efficiency becomes critical,
necessitating methods that convey information compactly and reliably.
Recent research has begun to address these issues: efficient topic modeling
techniques have been developed to reduce communication overhead while
maintaining shared understanding \cite{doherty2018approximate}, and methods for
streaming semantic maps have demonstrated the feasibility of collaboration with
reduced bandwidth \cite{girdhar2019streaming}. Further, strategies for achieving
semantic agreement among robots are critical for cohesive multi-robot observation
campaigns, particularly when explicit human labeling is
sparse or unavailable \cite{jamieson2021multi}.
Key directions for future research include developing low-bandwidth interfaces
for real-time scientist input, robust algorithms for semantic agreement in
sparse labeling scenarios, and scalable coordination frameworks for multi-robot
scientist-guided exploration.

\titledsubsubsection{Simulators and Benchmarks for Adaptive Sampling and Information Gathering}

Real-world trials of adaptive sampling are expensive and often infeasible at
early stages. Existing marine simulators provide valuable hydrodynamics and
sensor models \cite{potokar2024holoocean,lin2022oystersim,zhang2022dave}, yet
they seldom embed \emph{realistic environmental fields}.  Ideally, future
simulators could ingest historical data (e.g., temperature,
salinity, currents, chlorophyll, wind, sea-ice thickness) to create
replayable, physics-consistent scenarios across ocean, atmosphere, and
cryosphere that capture the relevant phenomena of interest.  Benchmarks that
score algorithms on metrics such as information gain or forecast improvement
would let researchers quantify the climate value of new adaptive-sampling
policies without overfitting to a single domain. Extending these tools beyond
the marine realm (e.g., stratified atmospheric boundary layers) will foster
cross-domain methods transferable across data-sparse Earth systems.

\titledsubsubsection{Learned Modeling of Dynamical Systems}

While the controls and planning community has been leveraging data-driven
approaches to capture complex dynamics, the Earth sciences have been making
similar pushes to develop data-driven parameterizations (representations of
key processes that cannot be directly modeled). Effectively these two
communities have been attacking the same problem: how to capture
difficult-to-model processes in a way that is accurate, data-efficient, and
generalizable
\cite{sayre2017three,kochkov2024neural,kolter2019learning,yu2024learning}.
There is potential for collaboration between these two communities to accelerate
progress in developing data-driven models that can be used in Earth system
models.

\begin{boxMarginLeft}
      \vspace{1em}
      \subsubsection*{Future Directions: Controls \& Planning in the Earth Systems}
      \begin{itemize}
            \item \emph{Scalable multi-robot coordination.}
                  Coordinating heterogeneous teams (e.g., gliders, AUVs, buoys) in
                  dynamic environments requires new algorithms for decentralized
                  planning, belief sharing under intermittent communication, and
                  role assignment under uncertainty.

            \item \emph{Flow- and forecast-aware sampling.}
                  Missions targeting dynamic ocean or atmospheric features must
                  incorporate real-time flow models and uncertainty-aware forecasts
                  to anticipate environmental change and adapt trajectories
                  accordingly.

            \item \emph{Extreme event-responsive planning.}
                  Capturing high-resolution data during fast-evolving events (e.g.,
                  algal blooms, storms, ice calving) demands planning methods that
                  can operate with limited prior information and make rapid
                  decisions under tight resource constraints.

            \item \emph{Scientist-guided autonomy at scale.}
                  Enabling effective human oversight in multi-platform deployments
                  calls for low-bandwidth interfaces, real-time semantic map
                  updates, and robust methods for resolving semantic disagreement
                  among platforms with partial supervision.

            \item \emph{Cross-domain learned models for physical systems.}
                  Closer integration between robotic learning and Earth system
                  science can accelerate the development of data-driven
                  parameterizations for unresolved dynamics in ocean, weather, and
                  climate models.

            \item \emph{Simulation and benchmarking infrastructure.}
                  Physics-informed, data-ingestible simulators and standardized
                  benchmarks are needed to evaluate adaptive sampling algorithms
                  across domains, enabling fair comparisons and accelerated method
                  development.
      \end{itemize}
\end{boxMarginLeft}

\titledsubsection{Estimation \& Perception}
\label{sec:earthsciences:state-estimation-and-perception}

\titledsubsubsection{Underwater Localization}

One of the greatest limitations in deploying underwater observing platforms is
the ability to localize them for long periods of time (e.g., hours to days).
This challenge is actually one of the greatest sources of cost for many
autonomous underwater vehicles, which rely on expensive equipment for accurate
localization \cite{rypkema2017one}.  Unlike terrestrial and aerial robotics,
where GPS often provides a reliable and globally available positioning system,
underwater robots must rely on alternative approaches due to the rapid
attenuation of electromagnetic signals in water.

The baseline for underwater navigation is inertial navigation. In the absence of
external positioning signals, underwater vehicles rely on inertial navigation
systems (INS), which integrate measurements from accelerometers and gyroscopes
to estimate position over time. However, inertial drift (the accumulation of
small errors in sensor readings) causes significant degradation in positional
accuracy over long deployments. This means that without external corrections,
even high-end INS solutions, such as those leveraging fiber optic gyroscopes
\cite{bergh1984overview} or ring laser gyroscopes \cite{chow1985ring},
experience large positional uncertainty after extended operations.
Beyond possible breakthroughs in the quality of low-cost inertial sensors, we
view two primary pathways to improve underwater navigation.

A first pathway lies in improving algorithms for inertial drift correction. In
pedestrian navigation, foot-mounted IMUs have been successfully stabilized using
zero-velocity updates \cite{wahlstrom2020fifteen}. Similar motion-constrained
priors exist in underwater vehicles, which often follow structured patterns such
as straight transits, periodic turns, or repeated dive cycles. These
regularities could be exploited to constrain drift in long-duration missions.
In parallel, learned approaches have shown promise in modeling complex inertial
error dynamics that are difficult to capture analytically. Recent work has
demonstrated that learning-based odometry systems can significantly reduce error
accumulation (e.g., \cite{liu2020tlio,herath2020ronin,jayanth2025neural}).
While these methods have demonstrated great capabilities, there is still a need
to push techniques specialized for compute- and power-constrained platforms
necessary for underwater operations.  Future models should allow for external
information (e.g., GPS fixes or acoustic positioning updates) to be integrated
into the learned inertial navigation framework to update the inertial drift
model and correct for accumulated errors.

The second pathway is integrating other sensing modalities (e.g., acoustic or
optical) to complement the inertial navigation system. Acoustic-based
techniques, such as long baseline (LBL) and ultra-short baseline (USBL) systems,
allow external positioning updates but require prior infrastructure deployment
and may be impractical for large-scale operations \cite{paull2013auv}. Doppler
velocity logs (DVL) provide velocity estimates relative to the seafloor,
significantly reducing drift rates when operating near the bottom. Acoustic
Doppler current profilers (ADCP) use similar technology to estimate velocities
relative to the water column (i.e., currents). Optical navigation techniques
(e.g., visual odometry \cite{shank2022advancing, rahman2019svin2}) and imaging
sonar-based simultaneous localization and mapping (SLAM) methods for
low-visibility conditions, can also provide critical localization cues.
Additionally, geophysical navigation approaches leverage known seafloor
topography, magnetic anomalies, or gravitational variations to provide
georeferenced position updates without requiring external infrastructure
\cite{liu2022comprehensive, djapic2015challenges}.  Emerging research also
explores the use of environmental cue-based localization, such as biogeochemical
markers, as potential references for
long-term navigation \cite{kumar2025flow}.
Future research should focus on what can be done with low-cost hardware and
develop techniques to incorporate these diverse sensing modalities into modern
navigation frameworks, which will undoubtedly require both research into how to
extract useful localization information from these modalities as well as how to
fuse them together in a robust and cohesive manner.

New datasets and benchmarks could greatly accelerate progress in this area.
Existing data currently emphasizes standard high-end sensors (e.g., INS, DVL,
multibeam sonar) which are cost-prohibitive for large-scale, low-cost
deployments \cite{bernardi2022aurora,wang2023underwater} or largely depend on
visual data \cite{ferrera2019aqualoc}, which is not effective in many underwater
environments (e.g., turbid waters or low-light conditions).  There is
value in new datasets that:
\begin{itemize}
      \item pair \emph{low-cost} sensing (MEMS IMUs, single-beam echo-sounders,
            one-way acoustic ranging) with ground-truth trajectories to spur research in
            cost-effective, energy-efficient navigation \cite{papalia2024certifiably};
      \item capture diverse acoustic propagation conditions (multipath,
            thermoclines, biological noise) to ensure robustness to environmental
            conditions; and
      \item include emerging modalities such as ambient soundscape navigation
            \cite{jang2023navigation}, terrain-relative sonar maps
            \cite{kimball2011sonar}, quantum inertial prototypes
            \cite{cheiney2018navigation}, and distributed acoustic sensing
            \cite{min2021optical} so algorithms can be benchmarked before field
            deployment.
\end{itemize}

In sum, advances in underwater localization will benefit from a concerted push
along three fronts: the improvement of compute-constrained inertial navigation,
integration of sensing modalities that have not been traditionally used for
localization, and the development of new datasets and benchmarks to catalyze
research in localization using low-cost sensors and novel modalities.

\titledsubsubsection{New Measurement Modalities}

Advances in robotic perception are expanding the range of sensing modalities
used in environmental monitoring, including optical, acoustic, biochemical, and
genetic techniques. This diversification helps address \emph{modal sparsity} by
enabling both novel measurements (e.g., in situ eDNA) and scalable versions of
existing methods (e.g., wide-area imaging, passive acoustics).

Camera-based perception has supported tasks such as species monitoring
\cite{liu2020towards,borowicz2018multi}, aquaculture inspection
\cite{kelasidi2020cagereporter,kelasidi2023robotics}, coastal wave and current
estimation \cite{dooley2025estimating,anderson2021quantifying}, phytoplankton
classification \cite{olson2007submersible}, snow accumulation tracking
\cite{harder2016accuracy}, and ice mapping \cite{fugazza2018combination}.
Acoustic sensing also plays a growing role in evaluating ecosystem health
\cite{mccammon2024discovering,mooney2024understanding}. New modalities such as
in situ chemical analysis and eDNA sampling are enabling autonomous observation
of biodiversity, nutrient cycling, and early ecosystem disruptions.

As these systems mature, future work must focus on deploying perception models
on low-power edge hardware, improving robustness under field conditions, and
extending capabilities to measure high-priority variables that remain
underobserved. One particularly critical area is the autonomous estimation of
direct carbon fluxes in coastal and open-ocean systems, which remain a major
source of uncertainty in climate models. Robotic perception research aimed at
integrating physical, biological, and chemical signals has the
potential to significantly advance Earth system monitoring.

\titledsubsubsection{Mapping and Natural Scene Reconstruction}

In many Earth science contexts, raw sensor data must be synthesized into
spatial or spatiotemporal reconstructions, either for physical assets (e.g.,
infrastructure) or natural processes (e.g., glacial retreat). These
reconstructions form the basis for analysis, monitoring, and decision-making.

For built environments, such as offshore energy infrastructure or aquaculture
installations, high-resolution 3D maps enable routine inspection without
requiring human divers or crewed support
\cite{kelasidi2020cagereporter,mai2016subsea}. At larger scales, natural scene
reconstruction plays a critical role in environmental monitoring. Applications
include tracking glacier and iceberg morphology
\cite{vargo2017using,piermattei2015use} and reconstructing permafrost layers
\cite{van2018contribution}: each requiring different sensing modalities,
spatial resolutions, and processing workflows.

These tasks pose significant technical challenges. Visual mapping often
struggles in low-contrast or poor-visibility conditions such as murky water,
snow, or sediment-covered seafloors \cite{solonenko2015inherent,wolfl2019seafloor}. Subsurface reconstructions using geophysical techniques like
ground-penetrating radar must contend with sparse and noisy data, requiring
robust signal processing and spatial inference pipelines
\cite{schennen2022seasonal}.

Future research should focus on robust scene reconstruction under adverse
sensing conditions, scalable algorithms for long-term, large-scale spatial
modeling, and multimodal fusion techniques for task relevant data (e.g., visual,
acoustic, and geophysical). Such capabilities will greatly aid persistent
monitoring of climate-relevant environments, particularly where manual
inspection is infeasible, challenging, or dangerous.

\titledsubsubsection{Change Detection in Natural Environments}

Many Earth science questions reduce to ``what has changed?'' and ``why?''. Even
with high-quality reconstructions of, e.g., glaciers \cite{ryan2015uav} or coral
reefs \cite{burns2015integrating}, both detecting and interpreting change
remains difficult.

One challenge is detecting specific changes between reconstructions. This often
requires registration of scenes that differ in shape or topology. Promising
directions include registration methods that leverage domain-specific knowledge
and change detection techniques that can bypass registration altogether.

A second challenge lies in causal attribution. Identifying \emph{what} changed
is often less valuable than understanding \emph{why} to isolate the precise
processes driving major changes (e.g., causes of ice melt). Frameworks
are needed that integrate (i) change detection, (ii) candidate causal
mechanisms, and (iii) domain knowledge as priors or constraints to infer likely
causes.

\titledsubsubsection{Data Assimilation}

Data assimilation is a modeling paradigm that updates a model's state based on
observational data to improve model performance. This is done by combining the
model's predictions with the observations in a way that is consistent with the
uncertainty in both the model and the observations.  Data assimilation,
typically in the form of optimal interpolation, 4D variational assimilation, or ensemble Kalman
filtering, has applications as broad as weather forecasting
\cite{gandin_objective_1965,mcpherson_operational_1986,lorenc_analysis_1986},
ocean modeling
\cite{bretherton_technique_1976,malanotte1996modern,thomson_chapter_2014}, and
past climate reconstruction \cite{Carrassi2018,Fang2016}.  Many of the inference
techniques underpinning modern approaches to data assimilation (e.g.,
variational inference and Kalman filtering) are intimately familiar to the state
estimation community which has been independently utilizing and advancing these
techniques for years.

This is a promising area to see if there is room for improvement through direct
collaboration. The robotic state estimation community has been largely driven by
computational restrictions, considerations of uncertainty, and the need for
real-time performance, which has led to advances in robustness, efficiency, and
scalability of these underlying algorithms. There is potential to combine these
insights with the deep domain knowledge of the Earth sciences community to
develop new data assimilation techniques that improve on current issues of
scalability and uncertainty quantification.

\titledsubsubsection{Task-Specific Datasets for Earth Monitoring}

Targeted, task-specific datasets can drive major advances in environmental
monitoring by enabling perception algorithms for hard-to-measure phenomena. For
example, estimating wave spectra from shoreline video could significantly
improve air-sea flux models, while tracking glacier and iceberg evolution could
shed light on their governing physical processes. Curated, well-scoped datasets
for these tasks would lower barriers for roboticists and accelerate the
development of deployable tools for Earth scientists.

One specific area of interest is in fish stocks, which are under growing
pressure from climate change and overfishing, yet population monitoring remains
sparse and labor-intensive. One possibly impactful direction could be
multi-modal datasets that, for example, combine video (for species ID and
health), acoustics (for biomass), environmental sensors (for habitat
quality), and eDNA (for biodiversity). This could power perception systems that
can aid in fisheries monitoring and management, ultimately supporting
sustainable fisheries and food security
\cite{swart2016ocean,masmitja2020mobile}.

\begin{boxMarginLeft}
      \vspace{1em}
      \subsubsection*{Future Directions: Estimation \& Perception in the Earth Systems}
      \begin{itemize}
            \item \emph{Low-cost and reliable underwater navigation.}
                  Advances are needed in lightweight, limited-drift inertial
                  navigation algorithms and multi-modal sensor fusion (e.g., DVL,
                  acoustic ranging, optical cues) that work on energy- and
                  compute-limited platforms. Future methods should integrate learned
                  models, motion priors, and intermittent external corrections to
                  maintain localization over long-duration missions.

            \item \emph{Benchmarks for resource-constrained localization.}
                  Progress in underwater state estimation is hampered by lack of
                  representative datasets. New benchmarks should include low-cost
                  sensor suites, challenging acoustic and visual conditions, and
                  emerging modalities (e.g., terrain-relative sonar, ambient
                  soundscape) to enable realistic algorithm development and
                  evaluation.

            \item \emph{Multimodal environmental perception for biogeochemical sensing.}
                  Robotic perception must extend beyond physical mapping to
                  integrate acoustic, visual, and chemical signals. One key target
                  is the estimation of ocean carbon fluxes. Depending on the
                  specific process being studied, this may requiring onboard fusion
                  of physical (e.g., wave state), chemical (e.g., dissolved \coo),
                  and biological (e.g., eDNA) cues to infer process-level dynamics.

            \item \emph{Robust scene reconstruction in adverse sensing conditions.}
                  Future mapping pipelines should combine acoustic, optical, and
                  geophysical data to reconstruct large-scale natural environments
                  (e.g., ice, seafloor, permafrost) under poor visibility,
                  low-light, or sparse data conditions. These reconstructions
                  underpin long-term monitoring and must be scalable and
                  fault-tolerant.

            \item \emph{Causal change detection in natural systems.}
                  Moving beyond ``what changed?'' to ``why did it change?'' requires
                  new frameworks that combine change detection, causal inference,
                  and domain priors. Techniques should account for spatial
                  variability, sensor noise, and ambiguous or sparse data to
                  attribute environmental change to physical drivers.

            \item \emph{Scalable data assimilation with real-time constraints.}
                  Roboticists bring deep expertise in real-time state estimation,
                  uncertainty quantification, and approximate inference. These
                  insights can be applied to develop scalable data assimilation
                  tools for Earth systems, improving model accuracy while
                  maintaining operational feasibility.

            \item \emph{Task-specific datasets for perception-limited monitoring.}
                  Datasets focused on hard-to-measure but high-impact variables
                  (e.g., fish stocks or iceberg dynamics) can provide important
                  societal value.  Combining diverse data sources (e.g., video,
                  acoustics, eDNA) may provide a pathway to training and evaluating
                  perception algorithms for such tasks.
      \end{itemize}
\end{boxMarginLeft}

\titledsubsection{Field Robotics}
\label{sec:earthsciences:field-robotics-design}
Field robotics, focused on designing and deploying robots in unstructured
environments,  is central to robotics' role in the Earth systems. Developing
effective field systems requires integrating expertise in mechanical design,
autonomy, sensing, and environmental physics. As the Earth sciences become
increasingly data-driven, the ability to deploy autonomous systems in diverse,
extreme environments  becomes a significant enabler for scientific progress.

Energy efficiency is a particularly important consideration, directly tied to
choices in propulsion systems, aerodynamic or hydrodynamic design (for aerial or
marine vehicles), fuel and power sources, onboard computation, and operational
behaviors. In remote deployments, reducing energy consumption extends mission
duration and lowers operational costs, often determining whether a mission is
viable or not. This applies not only to mobility but also to sensing and inference
workloads: perception and autonomy stacks can impose non-trivial energy demands,
especially when deployed on edge compute hardware under duty-cycled or low-power
constraints. Addressing these challenges will require innovative designs that
are robust to harsh conditions, compatible with novel sensors and sampling
mechanisms, and tailored to the environmental constraints of specific domains
(e.g., high-pressure ocean depths, polar terrain, or turbulent air masses).
There is no one-size-fits-all solution; success will depend on matching platform
design to both environmental context and scientific need.

\titledsubsubsection{Scalable, Long-Duration, and Energy-Efficient Platforms}

Owing to the vastness of our natural world, autonomous observation technologies
provide the greatest value when they can be deployed for long durations and at
large spatial scales. Achieving this requires balancing three tightly coupled
factors: energy efficiency (to support persistent deployments), cost (to enable
widespread scaling), and range (to cover geophysically relevant domains). These
trade-offs were central to the success of the Argo program, which used simple,
drifting platforms that leveraged natural ocean circulation to minimize
propulsion needs while maintaining scientific utility.

The need for scalable, long-duration platforms is especially acute in underwater
and atmospheric environments. Underwater, energy demands for propulsion and
localization are high, and accurate navigation often depends on expensive
equipment. While buoyancy-driven gliders
\cite{jones2014slocum,sherman2001autonomous,eriksen2001seaglider} have enabled
longer deployments with low-power payloads, their speed and coverage are
limited. Hybrid systems such as long-range AUVs aim to merge endurance with
capability, allowing for larger sensor payloads and multi-day missions
\cite{kukulya2016development}. On the ocean surface, wind- and solar-powered
vehicles (e.g., Saildrone) have demonstrated promising results in persistent
sensing campaigns \cite{gentemann2020saildrone}, yet similar endurance has been
harder to achieve in dynamic subsurface environments or for high-power sensors
(e.g., for biogeochemistry or acoustic mapping).

In the atmosphere, high-altitude balloons and long-endurance drones face
constraints from payload capacity, power generation, and exposure to weather.
While superpressure balloons and solar-powered aerial vehicles offer extended
flight duration, there are still open challenges in improving cost efficiency,
onboard autonomy, and robustness in the face of high winds, turbulence, and
icing conditions. These platforms would benefit from lightweight sensor
integration, adaptive control strategies, and low-cost manufacturing to support
large-scale deployments.

Similar needs exist in the cryosphere, where vast, sparsely instrumented regions
are difficult to monitor using traditional infrastructure. Long-duration
autonomous platforms, whether aerial, surface, or subsurface, must contend with
low temperatures, unstable terrain, and minimal access to solar power. There is
significant opportunity for systems that can traverse ice sheets or operate
beneath sea ice with minimal energy expenditure, potentially by using local
environmental cues (e.g., gravity, slope, or currents) to aid mobility and
guidance.

Ultimately, addressing these challenges will require innovations across
mechanical design, energy systems, sensor integration, and autonomy. Promising
directions include energy harvesting (e.g., wave, thermal, or solar) \cite{hine2009wave},
intelligent duty-cycling of sensors and compute \cite{smith2021abyssal, mccammon2022adadaptive}, bio-inspired
locomotion \cite{zhong2021tunable}, and use of environmental dynamics to passively assist
navigation and coverage \cite{breier2020revealing}.  Designing platforms with these strategies in
mind may provide keys to enabling scalable, persistent, and cost-effective
observation of remote and climate-critical environments. Furthermore, the development of
reliable, modular docking systems for recharging and data transfer will greatly aid in
extending mission durations and reducing the need for costly retrieval
operations \cite{curtin1993autonomous,singh1996integrated,singh2001docking}.

\titledsubsubsection{Sensors, Instrumentation, and Sampling Mechanisms}
\label{sec:earthsciences:field-robotics:sensors-instrumentation-sampling}

To advance Earth systems research, we not only need improved autonomous platforms,
but also improved sensors and sampling mechanisms that expand what those
platforms can measure. This includes both \emph{in situ} sensors for onboard
observation and physical sampling systems that enable \emph{ex situ}
laboratory analysis. Together, these capabilities determine the scope and
utility of robotic observations across ocean, atmospheric, and cryospheric
domains.

In oceanography, the most widely used sensors measure conductivity (a proxy for
salinity), temperature, and depth (CTD). These are often paired with water
sampling bottles mounted on deployment frames (Rosette), used to collect water at
specified depths for laboratory analysis \cite{edgcomb2016comparison}. However,
such techniques offer limited temporal resolution and spatial coverage, are
labor-intensive, and can introduce biases when environmental conditions change
during sample retrieval.

Programs like the World Ocean Circulation Experiment \cite{dickson2001world} and
the Ocean Observatories Initiative \cite{isern2003ocean,trowbridge2019ocean}
have enabled long-term physical monitoring across select regions, but fall short
in supporting large-scale biological and geochemical studies. Many important
variables, like nutrient concentrations, total alkalinity, dissolved inorganic
carbon, chlorophyll, or microbial community composition, are not routinely
measured, or cannot be captured in situ with sufficient frequency or resolution.
Furthermore, most current platforms lack the onboard processing or sample
preservation capabilities required to support real-time, high-throughput
analyses of transient or spatially patchy phenomena like plankton blooms,
hypoxic zones, or methane seeps \cite{brewin2021sensing, levin2016hydrothermal}.

Developing new sensors and samplers is only one part of the challenge. These
instruments must also be integrated into robotic systems, a process that
requires addressing calibration, drift, power consumption, sample storage
limitations, and environmental durability. Many high-value sensing techniques
(e.g., chemical analyzers, spectrometers) rely on consumables or have limited
sampling capacity, which can restrict deployment duration. Extreme environments
(e.g., high-pressure deep-sea settings, ice-covered surfaces, or turbulent
coastal zones) further compound these challenges, demanding tailored engineering
solutions.

To maximize advances in autonomy-enabled science, the robotics and Earth science
communities must collaborate to develop new instruments that meet
domain-specific scientific requirements while being compatible with autonomous
deployment. Priorities include miniaturizing lab-grade sensors for in situ use,
increasing sampling throughput, and ensuring robustness across harsh
environments. As instrumentation expands across modalities and domains, robotic
platforms will be able to answer increasingly complex and interdisciplinary
climate questions.

\paragraph{Atmospheric Sensing}

Accurate atmospheric sensing is critical for understanding climate processes,
particularly in guiding models of atmospheric physics and tracking composition
(e.g., aerosols, greenhouse gases, and particulates), which strongly influence
air quality and radiative forcing. Traditional monitoring relies on satellite
remote sensing, ground-based stations, and expendable weather balloons, each with
limitations. Satellites provide global coverage but often lack the spatial or
temporal resolution needed for boundary layer or cloud-scale processes
\cite{emery2017introduction}. Ground networks offer high-accuracy data but are
geographically sparse \cite{fuzzi2015particulate}. Weather balloons are
single-use and operate for only a few hours. Targeted airborne sampling (e.g.,
to quantify carbon fluxes over the Amazon \cite{gatti2021amazonia}) typically
requires piloted aircraft, which are costly and logistically intensive.

Autonomous systems offer a path toward more scalable, adaptive, and
cost-effective atmospheric sensing. New platforms such as steerable
high-altitude balloons \cite{windborne} can achieve days-long operation with
altitude control, while lightweight drones have been developed to profile the
atmospheric boundary layer in both academic \cite{segales2020coptersonde} and
commercial \cite{meteodrone} contexts.

Autonomous robotic platforms offer new opportunities to close persistent
observational gaps in climate-relevant atmospheric science. Some of the
highest-impact sensing priorities include:
(i) high-resolution profiling of atmospheric boundary layer
dynamics to improve understanding of heat, moisture, and momentum exchange
between the surface and atmosphere;
(ii) in situ measurements of cloud properties and their interactions with
airborne particles (aerosols), to better understand how clouds reflect sunlight
and trap heat; and
(iii) scalable platforms for regional methane and
hydroxyl radical (OH) monitoring, to better understand greenhouse gas lifetimes.
Achieving these goals will require lightweight, purpose-built sensors, reliable
autonomous capabilities, and new strategies for coordinated sensing across
multiple robotic platforms.
\alan{citations needed for the above three points}

\paragraph{Coastal, Freshwater, and Shallow Environments.}
Coastal and freshwater zones are both densely populated and ecologically
dynamic, yet remain difficult to monitor with current robotic tools. Complex
hydrodynamics, shallow depths, strong tidal flows, and high turbidity introduce
distinct sensing and mobility challenges. These differ markedly from the
constraints faced in open-ocean deployments. Meanwhile, these regions are at the
forefront of climate impacts, facing sea-level rise, erosion, habitat loss, and
ecosystem shifts, while also supporting critical infrastructure and livelihoods
for billions of people worldwide
\cite{coleman2022quantifying,halpern2008global,hoegh2010impact}.

Remote sensing provides valuable large-scale monitoring, but remains limited in
nearshore zones by cloud cover, coastal geometry, water clarity, and coarse
resolution. in situ robotic observations could fill these gaps, but face
significant constraints: wave-driven turbulence and sharp topographic gradients
can complicate operations; dense human activity (shipping,
fishing, recreation) imposes safety, regulatory, and access constraints on
autonomous deployments; and shallow environments complicate sensing (e.g.,
multipath in acoustics, distortion in vision) \cite{ferguson2018sound,solonenko2015inherent}.

Although small AUVs may be viable in such settings, they currently remain expensive to
scale. UAVs are often prohibited in sensitive coastal areas due to
disturbance risks to wildlife, airspace users, or local communities
\cite{llewellyn2015getting,gohari2024systematic}. To unlock robotic monitoring
in these regions, we need new platforms that are (i) low-cost and easily
redeployable, (ii) minimally invasive to human and ecological systems, and (iii)
can operate reliably in these complex settings.
Advancing robotics for coastal and freshwater environments potentially offers
outsized impact, supporting climate observation, ecosystem monitoring, and the
resilience of communities that depend on these vulnerable regions.

\paragraph{Ice-Ocean Interactions}
Understanding ice-ocean interactions is crucial for climate science,
particularly in polar regions where glaciers, sea ice, and ice shelves influence
global sea level and ocean circulation. There are many important scientific
questions that autonomous sensing and monitoring can help address. For example,
in the case of sea ice, it is of great value and difficulty to get beneath-the-ice
measurements of quantities such as light transmission,
turbulence, and biogeochemical parameters
\cite{smith2023thin,roach2024physics,vancoppenolle2013role}. Such
sensing requires non-standard sensors that are not typically integrated into
underwater robotic vehicles. Developing robotic platforms capable of carrying
and operating these specialized instruments in sub-ice and ice-adjacent
environments has the potential to greatly advance our understanding of these
important polar processes.

\paragraph{Core Sampling}
Core sampling (obtaining cylindrical samples of soil, sediment, ice, or rock
layers) is a valuable way to obtain direct insights into past climate
conditions, geological processes, and biological activity. Such insights, in
turn, help us better understand what the future may hold under different
scenarios. Robotic coring has great value, as it offers the potential for
sampling in difficult or dangerous environments (e.g., the deep ocean). By
incorporating autonomy, we also gain the
potential for more frequent and consistent sampling than systems that
require substantial human operation. Notably, many current methods rely on
expensive ship time, motivating more autonomous and persistent solutions.

Core sampling has been successfully demonstrated for
terrestrial robots obtaining soil samples
and in the marine domain for solid rock \cite{stakes1997diamond}
and sediment coring \cite{fundis2022nautilus,phung2023enhancing} using
underwater vehicles.  Additionally, motivated by studying extraplanetary
environments, a number of robotic coring mechanisms for ice and rock have been
developed for mobile robots
\cite{paulsen2006robotic,quan2017drilling,zhang2021robotic,trautner2024prospect}.

While these systems have been successful in their respective domains,
substantial limitations on coring capabilities remain. For example, it is difficult
to obtain cores in many underwater settings with mobile platforms, due to the
need for a stable platform capable of balancing the forces generated during
coring. There is a need for coring mechanisms that can provide such reaction
forces and be adapted to a wide range of platforms (e.g.,
\cite{backus2020design}). Similarly, issues of platform stability, power
consumption, cost, and sample management are open challenges for coring in
cryospheric environments (e.g., for ice or permafrost samples), and represent
promising opportunities for innovation in autonomous systems design.

\paragraph{Ice and Permafrost}

Beyond core sampling, there is great potential for robotic systems to expand
scientific capabilities in ice and permafrost environments.
Many valuable cryospheric instruments (e.g., non-destructive tools such as
ground penetrating radar), pose challenges in integrating into robotic platforms
due to requirements such as platform stability, power, and payload
capacity.  However, these instruments allow valuable studies on topics such as
ice thickness, subsurface
layering, melt dynamics, and permafrost thaw
\cite{lapazaran2016errors,vaughan1999distortion,campbell2018ground}. These
parameters are key to understanding the trends and stability of the
cryosphere. There has been exciting work towards robotic deployment of such
instruments, notably custom drone-mounted ice penetrating radar
\cite{teisberg22uav} and ground-based vehicles designed for Arctic
terrain which can tow GPR systems \cite{lever2013autonomous,elliott2018new}.

There is also room for further innovation in the development of autonomous
systems that can safely operate in these snowy, low-temperature environments and
for non-destructive sensors (e.g., radio based) that integrate into these
systems to provide high-resolution, low-cost, persistent monitoring of ice and
permafrost. Additionally, there is substantial value in autonomous systems that
are capable of borehole drilling and sampling for measurements such as
ice and permafrost temperature profiles.

\paragraph{Toward Robotic Field Laboratories}
Oftentimes when autonomous systems are deployed, they are limited to data
collection, while sample processing and detailed analysis still require post-hoc
laboratory work
\cite{manjanna2018heterogeneous,das2015data,liu2024progress}. Progress has been
made on this front. For example, environmental DNA (eDNA) sampling, a technique
that can be used for tasks such as biodiversity monitoring, algal bloom detection, and
species tracking, has been successfully integrated into autonomous systems
for in situ sampling and analysis
\cite{zhang2024two_auv_edna_sampling,hendricks2023compact,scholin2017quest}.
Similarly exciting progress has been made in the development of in situ
chemical sensors (e.g., dissolved oxygen and nutrients) \cite{wang2019advancing}.

Future robotic systems should explore the feasibility of in situ robotic field
laboratories that take advantage of continued advances in miniaturized sensors,
edge processing, and lab-on-a-chip technologies \cite{wang2019advancing} to
drive the onboard scientific capabilities of autonomous platforms.

\titledsubsubsection{Traversability in Remote, Harsh, or Ice-Covered Environments}

There are many terrestrial field situations (e.g., surveying permafrost or atop
glaciers) in which notions of traversability become key to ensuring that robots
can safely and effectively navigate the environment \cite{baril2022subarctic,
      pomerleau2022robotics, pereira2013risk, kunz2009toward, bares1999dante}. These
environments are often highly unstructured and present surface conditions like snow, ice, scree, or mud that differ significantly from those
encountered in urban or temperate regions. Common terrain features like
crevasses, snow drifts, or soft ground can make locomotion hazardous, while
environmental conditions such as low contrast, glare, or whiteout reduce the
reliability of vision-based sensing.

Climate-relevant applications include autonomous sampling of permafrost, glacier
surface surveys, ice shelf monitoring, and robotic support for field campaigns
in polar regions. In these settings, robots must contend with cold temperatures,
limited power availability, GPS-denied navigation, and unreliable
communications.

Future research should focus on developing robust multimodal sensing approaches
(e.g., thermal, radar, proprioceptive) and traversability models tailored to
low-texture or otherwise ambiguous terrain. Improving robotic mobility in these
environments will substantially aid expanding autonomous operations into remote
regions of high scientific and climate relevance.

\begin{boxMarginLeft}
      \vspace{1em}
      \subsubsection*{Future Directions: Field Robotics in the Earth Systems}
      \begin{itemize}
            \item \emph{Endurance, scalability, and adaptive energy use.}
                  Extend deployment duration and scale through energy-efficient
                  design (e.g., gliders, solar- and wave-powered platforms),
                  autonomous docking, and adaptive duty-cycling of sensing and
                  computation. Hybrid locomotion may unlock broader spatial
                  coverage across diverse terrain (e.g., coastal zones).

            \item \emph{Embedded sensing and sample processing.}
                  Develop systems that not only collect data, but also analyze
                  it onboard.  This includes integration of novel instruments
                  (e.g., eDNA, nutrient sensors, spectrometers), lab-on-chip
                  technologies, and autonomous sample preservation, enabling
                  robots to act as mobile field laboratories.

            \item \emph{Robotics for under-observed and difficult environments.}
                  Push capability boundaries in challenging domains, from
                  turbid, dynamic coastal zones to cryospheric regions with low
                  sunlight and harsh terrain.  This includes robust sub-ice and
                  permafrost sensing, visually sparse localization, and
                  terrain-aware locomotion across snow, ice, and soft ground.

            \item \emph{Subsurface studies for ice and permafrost science.}
                  Advance autonomous systems for stable, adaptive subsurface
                  sampling in cryospheric environments. Priorities include
                  borehole drilling in snow, glacial ice, and permafrost;
                  recovery and preservation of climate-relevant cores; and
                  onboard analysis or autonomous return of samples for
                  stratigraphic, geochemical, and microbiological analysis.
      \end{itemize}
\end{boxMarginLeft}

\titledsubsection{Manipulation}
\label{sec:earthsciences:manipulation}

Robotic manipulation can aid in a range of climate-relevant marine
applications, including in situ instrumentation, ecological sampling,
infrastructure maintenance, and aquaculture automation. These tasks often
require precise, adaptive interaction in unstructured and dynamic environments
where traditional human or remotely operated interventions are limited by cost,
access, and safety.

In scientific settings, manipulators can enable sampling of fragile benthic
organisms that act as bioindicators of ocean acidification and warming
\cite{galloway2016soft}.  Autonomous sediment coring can support paleoclimate
reconstruction by accessing previously unsampled regions and improving temporal
resolution (see
\cref{sec:earthsciences:field-robotics:sensors-instrumentation-sampling}). In the context of
infrastructure, platforms capable of manipulation can support inspection and
repair of offshore energy systems \cite{mitchell2022review}, reducing reliance
on support vessels and expanding operational windows into otherwise dangerous
settings.  In aquaculture, automation of repair and sensor deployment can
similarly reduce dangerous labor demands and improve feasibility of long-term,
large-scale operations \cite{wang2021intelligent}.

\begin{boxMarginLeft}
      \vspace{1em}
      \subsubsection*{Future Directions: Manipulation in the Earth Systems}
      \begin{itemize}
            \item \emph{Precision scientific sampling in unstructured environments.}
                  Design manipulators that enable minimally invasive collection of
                  fragile biological and sediment samples, supporting studies of
                  past climate and ecosystem monitoring in regions otherwise difficult,
                  dangerous, or expensive to access.

            \item \emph{Autonomous intervention for offshore infrastructure.}
                  Enable robotic inspection, repair, and sensor deployment for offshore wind,
                  carbon capture, and aquaculture systems — reducing emissions from crewed
                  vessels, minimizing downtime, and supporting the resilience of low-carbon
                  energy and food systems in harsh or remote environments.

      \end{itemize}
\end{boxMarginLeft}

\titledsubsection{Conclusion}

Robotics offers a powerful means to advance climate science and sustainability
by improving Earth system monitoring, enabling adaptive sampling, and expanding
our ability to operate in extreme and remote environments. In applications such as autonomous ocean
profiling, atmospheric sensing, and ice sheet monitoring, robotic systems can
aid in filling critical observational gaps and enhance our understanding of
climate processes. These technologies also support sustainable human activities,
such as aquaculture, marine energy, and climate intervention efforts.

However, we would also like to emphasize that technical advances are not the
only area for impact. Regulatory and policy action can support ongoing efforts
and help steer society toward effective mitigation and adaptation using current
and emerging capabilities. Additionally, administrative and logistical efforts
to break down data silos and improve data sharing between institutions and
organizations can make a major difference by unlocking massive amounts of
valuable data that has already been collected but is inaccessible to the broader
community \cite{brett2020ocean}.

Altogether, through advancing our technical capabilities, fostering
collaboration between scientists and roboticists, reducing data silos,
and making science-driven policy decisions, there are many opportunities for
climate impact in the Earth systems.


\section{Conclusion}
\label{sec:conclusion}

Climate change is a generation-defining problem, and many in the robotics
community are looking for ways to contribute their skills. In this paper, we
have laid out a roadmap for roboticists to contribute to climate change
mitigation, adaptation, and science, highlighting opportunities for high-impact
research in six climate domains. In each of these domains --- energy, buildings,
transportation, industry, land use, and earth systems --- we believe that
interdisciplinary research can bring not only physical robots but also the core
computational tools of robotics (controls, planning, perception, etc.) to bear
on important climate-relevant problems. Our goal is to spark new lines of
inquiry and foster interdisciplinary collaborations between roboticists and
climate experts.

Before concluding, it is important to discuss two final points that apply to all
research themes identified in this paper. The first is the sustainability of
robotics itself, and the second is the need for interdisciplinary research to
connect roboticists with climate domain experts and policymakers.

\titledsubsection{Sustainability of robotics}

Throughout this paper, we have focused on the potential benefits of robotics
research to climate change mitigation and adaptation efforts. It is also
important to discuss the potential drawbacks; namely, whether robotics research
poses any potential risks to sustainability. In this section, we briefly discuss
the climate impacts from building, training, and operating robots. While we do
not believe that these risks outweigh the potential benefits, they are important
to consider.

First, we can ask whether the emissions from physical robots pose a
sustainability risk; that is, what is the climate cost of building and running
more robots? While prior works have conducted lifecycle assessments for robots
in specific applications (e.g. drone
delivery~\cite{koiwanitAnalysisEnvironmentalImpacts2018,figliozziLifecycleModelingAssessment2017,stolaroffNeedLifeCycle2014}
and agricultural robotics~\cite{pradelComparativeLifeCycle2022}), it is
difficult to generalize these results to different applications and robotic
platforms. For example, a steel robot arm will have a very different impact than
a drone with a carbon fiber frame. However, we note that a typical drone battery
is almost 1000 times smaller than a typical EV battery. As a result, we believe
that concerns about the embodied emissions of robots are minimal at the
current scale of robot deployment. However, the robotics community should
monitor these impacts as adoption increases.

We find more reason for concern when it comes to the emissions involved in
training machine learning models used in robotics. Increased adoption of large
language models in recent years has led to well-founded concerns about the
amount energy used to train and deploy such models. As large ``foundation
models'' become more common in robotics, it is important to be aware of the
energy required to train these models. For example, the vision-language-action
model trained in~\cite{kimOpenVLAOpenSourceVisionLanguageAction2024} required
$\approx 6.4$ MWh to train (not counting the energy used to train the Llama 2
model used as a base), more than half the annual electricity consumption of a
single household~\cite{u.s.energyinformationadministrationElectricityUseHomes}.
While this energy consumption is currently small compared to non-robotics use of
large language models, it is worth considering as robotics adoption continues to
scale.

As robots become more capable and more widely deployed, it is likely that their
cost and efficiency may decline, mirroring learning curves observed in other
technologies. This factor, combined with the phenomenon known as Jevons paradox
where improved efficiency leads to increased usage that more than offsets the
savings from efficiency, makes it difficult to predict the future climate impact
of robotics. While it is not possible to completely avoid the environmental
impact of robotics, we urge the robotics community to be mindful of this
potential tradeoff.

\titledsubsection{A call to action for interdisciplinary research}

Many roboticists, ourselves included, naturally focus on technical solutions to
any problem. This focus is reflected in this paper, which has highlighted
primarily \textit{technical} solutions to climate problems. It is important to
emphasize that technical solutions alone are not sufficient in most domains. Effective climate
interventions will require technology to work in tandem with policy and
socio-economic factors. We urge roboticists to understand the broader systems in
which their technologies operate and recognize when to engage with diverse sets
of experts and stakeholders.

The success of climate-relevant robotics will depend not only on technical
innovation but also on systems-level thinking and sustained
collaboration across disciplines. We hope this roadmap empowers roboticists to
engage deeply with climate challenges, identify impactful research directions,
and partner with experts across domains. We call on the robotics community to
take up the challenge to build collaborative, impact-driven research programs to
address pressing climate needs.

\section{Acknowledgements}


This work was supported in part by the Northeastern University Institute for
Experiential Robotics, the National Science Foundation Graduate Research
Fellowship Program under Grant No. 2141064, the U.S. Department of Energy
through the Energy Innovator Fellowship, the MIT MathWorks Fellowship, the
MIT-Accenture Fellowship, and the U.S. National Science Foundation Division of
Biological Infrastructure under Grant No. 2152671.
Any views expressed in this paper are solely those of the authors and do not
necessarily reflect those of MA DOER, US DOE, or any other funding agency.

\printbibliography

\end{document}